\documentclass[11pt]{article}

\usepackage{lineno}

\usepackage{geometry}
\usepackage{setspace}
\newgeometry{vmargin={0.8in}, hmargin={0.75in}}
\usepackage{pdflscape}
\usepackage{ragged2e}

\usepackage[skip=1pt, indent=20pt]{parskip}

\usepackage{amsmath}
\usepackage{amsfonts}
\usepackage{bbm}
\usepackage{mathtools}

\usepackage{url}
\usepackage[hidelinks]{hyperref}
\hypersetup{
    colorlinks = black,
    linkcolor = black,
    anchorcolor = black,
    citecolor = black,
    filecolor = black,
    urlcolor = black
}
\usepackage[nameinlink]{cleveref}
\crefname{appendix}{section}{sections}
\creflabelformat{equation}{#2\textup{#1}#3}
 
\usepackage[labelfont=bf,labelsep=period,font=scriptsize,sf,textfont=sf]{caption} 
\usepackage{subcaption}
\usepackage{rotating}
\usepackage{makecell,cellspace}
\usepackage{graphicx}
\usepackage{tabularx,booktabs}
\usepackage{multirow}
\usepackage{array}
\usepackage{longtable}
\usepackage{placeins}
\usepackage{float}

\usepackage{minitoc}

\usepackage{tocloft}

\usepackage{jabbrv}
\usepackage[round,numbers,sort&compress,sectionbib]{natbib}

\usepackage[resetlabels]{multibib}
\newcites{SI}{SI References}

\usepackage{times}
\usepackage{hyphenat}
\newcommand{\ie}{\emph{i.e.}}
\newcommand{\eg}{\emph{e.g.}}
\newcommand{\cf}{\emph{cf.}}
\newcommand{\versus}{\emph{vs.}}
\newcommand{\etal}{\emph{et~al.}} 
\newcommand{\etcetera}{\emph{etc.}}

\usepackage{listings} 

\Crefname{lstlisting}{Code Block}{Code Blocks}
\usepackage[dvipsnames]{xcolor}

\definecolor{codegreen}{rgb}{0,0.6,0}
\definecolor{codegray}{rgb}{0.5,0.5,0.5}
\definecolor{codepurple}{rgb}{0.58,0,0.82}
\definecolor{backcolour}{rgb}{0.95,0.95,0.92}

\lstdefinestyle{mystyle}{
    backgroundcolor=\color{backcolour},   
    commentstyle=\color{codegreen},
    keywordstyle=\color{magenta},
    numberstyle=\ttfamily\scriptsize\scriptsize\color{codegray},
    stringstyle=\color{codepurple},
    basicstyle=\ttfamily\footnotesize,
    breakatwhitespace=false,         
    breaklines=true,                 
    captionpos=b,                     
    keepspaces=true,                 
    numbers=left,                    
    numbersep=5pt,                  
    showspaces=false,                
    showstringspaces=false,
    showtabs=false,                  
    tabsize=2
}

\usepackage[page,toc,title]{appendix}
\newcommand{\siref}[1]{\hyperref[appendixStart]{\emph{SI Appendix} \Cref{#1}}}
\newcommand{\subfigref}[2]{\hyperref[#1]{\Cref*{#1}#2}}
\newcommand{\mmref}[0]{\emph{\nameref{sec:materials}}}
\newcommand{\overviewref}[0]{\emph{\nameref{sec:overview}}}

\newcommand{\papertitle}[0]{A large-scale evaluation of commonsense knowledge in humans and large language models}
\title{\vspace*{-1.5em}\fontsize{18}{13.5} \sf \textbf{\papertitle}}
\date{}
\author{
  \sf
  Tuan Dung Nguyen\textsuperscript{\cis,\correspondingA}, 
  Duncan J. Watts\textsuperscript{\cis,\oid,\asc}, and
  Mark E. Whiting\textsuperscript{\cis,\oid,\pareto}
}

\makeatletter
\newcommand{\myfnsymbol}[1]{%
  \expandafter\@myfnsymbol\csname c@#1\endcsname
}
\newcommand{\@myfnsymbol}[1]{%
  \ifcase #1
  \or a
  \or b
  \or c
  \or d
  \or \TextOrMath{1}{\dagger}
  \fi
}

\newcommand{\cis}{\@myfnsymbol{1}}
\newcommand{\oid}{\@myfnsymbol{2}}
\newcommand{\asc}{\@myfnsymbol{3}}
\newcommand{\pareto}{\@myfnsymbol{4}}
\newcommand{\correspondingA}{\@myfnsymbol{5}}
\makeatother

\begin{document}
\doparttoc 
\faketableofcontents 
\maketitle

\noindent {\sf \small \textit{\textbf{Abstract}}---Commonsense knowledge, a major constituent of artificial intelligence (AI), is primarily evaluated in practice by human-prescribed ground-truth labels. An important, albeit implicit, assumption of these labels is that they accurately capture what any human would think, effectively treating human common sense as homogeneous. However, recent empirical work has shown that humans vary enormously in what they consider commonsensical; thus what appears self-evident to one benchmark designer may not be so to another. Here, we propose a method for assessing commonsense knowledge in AI, specifically in large language models (LLMs), that incorporates empirically observed heterogeneity among humans by measuring the correspondence between a model's judgment and that of a human population. We first find that, when treated as independent survey respondents, most LLMs remain below the human median in their individual commonsense competence. Second, when used as simulators of a hypothetical population, LLMs correlate with real humans only modestly in the extent to which they agree on the same set of statements. In both cases, smaller, open-weight models are surprisingly more competitive than larger, proprietary frontier models. Our evaluation framework, which ties commonsense knowledge to its cultural basis, contributes to the growing call for adapting AI models to human collectivities that possess different, often incompatible, social stocks of knowledge.}

\vspace{1em}

\renewcommand{\thefootnote}{\sf\myfnsymbol{footnote}}
\maketitle
\footnotetext[1]{\sf \footnotesize Department of Computer and Information Science, University of Pennsylvania, Philadelphia, PA 19104, USA.}
\footnotetext[2]{\sf \footnotesize Operations, Information and Decisions Department, The Wharton School, University of Pennsylvania, Philadelphia, PA 19104, USA.}
\footnotetext[3]{\sf \footnotesize Annenberg School for Communication, University of Pennsylvania, Philadelphia, PA 19104, USA.}
\footnotetext[4]{\sf \footnotesize Pareto, San Francisco, CA 94111, USA.}
\footnotetext[5]{\sf \footnotesize Corresponding author. E-mail: joshtn@seas.upenn.edu.}

\setcounter{footnote}{0}
\renewcommand{\thefootnote}{\arabic{footnote}}

\renewcommand{\thefootnote}{\fnsymbol{footnote}}
\let\oldfootnote\footnote
\renewcommand{\footnote}[1]{\oldfootnote{\scriptsize\sffamily#1}}
\long\def\@makefnmark{%
    \hbox {\@textsuperscript {\sffamily\@thefnmark}}
}

\maketitle
 
\section*{Introduction}
\label{sec:intro}

\noindent The physical and social worlds are tremendously complex and unpredictable, yet humans are able to navigate these environments almost effortlessly thanks to a special aptitude called common sense~\citep{gordonFormalTheoryCommonsense2017}. Endowing machines with this ability has remained a grand challenge throughout the history of artificial intelligence (AI) research~\citep{mccarthyProgramsCommonSense1959,minskyEmotionMachineCommensense2007,davisCommonsenseReasoningCommonsense2015,brachmanMachinesUsAI2023,kejriwalCanAIHave2024}. Recently, significant progress was made by large language models (LLMs), a class of machine learning systems that can synthesize commonsense knowledge extensively from their training data~\citep{petroniLanguageModelsKnowledge2019,davisonCommonsenseKnowledgeMining2019,jiangHowCanWe2020,jiangHowCanWe2021,liSystematicInvestigationCommonsense2022}, make highly flexible generalizations~\citep{choiCuriousCaseCommonsense2022}, and thus exhibit increasingly human-like tendencies~\citep{bubeckSparksArtificialGeneral2023,binzUsingCognitivePsychology2023,meiTuringTestWhether2024,strachanTestingTheoryMind2024,kosinskiEvaluatingLargeLanguage2024}.

\begin{figure*}[t]
    \centering
    \includegraphics[width=1\linewidth]{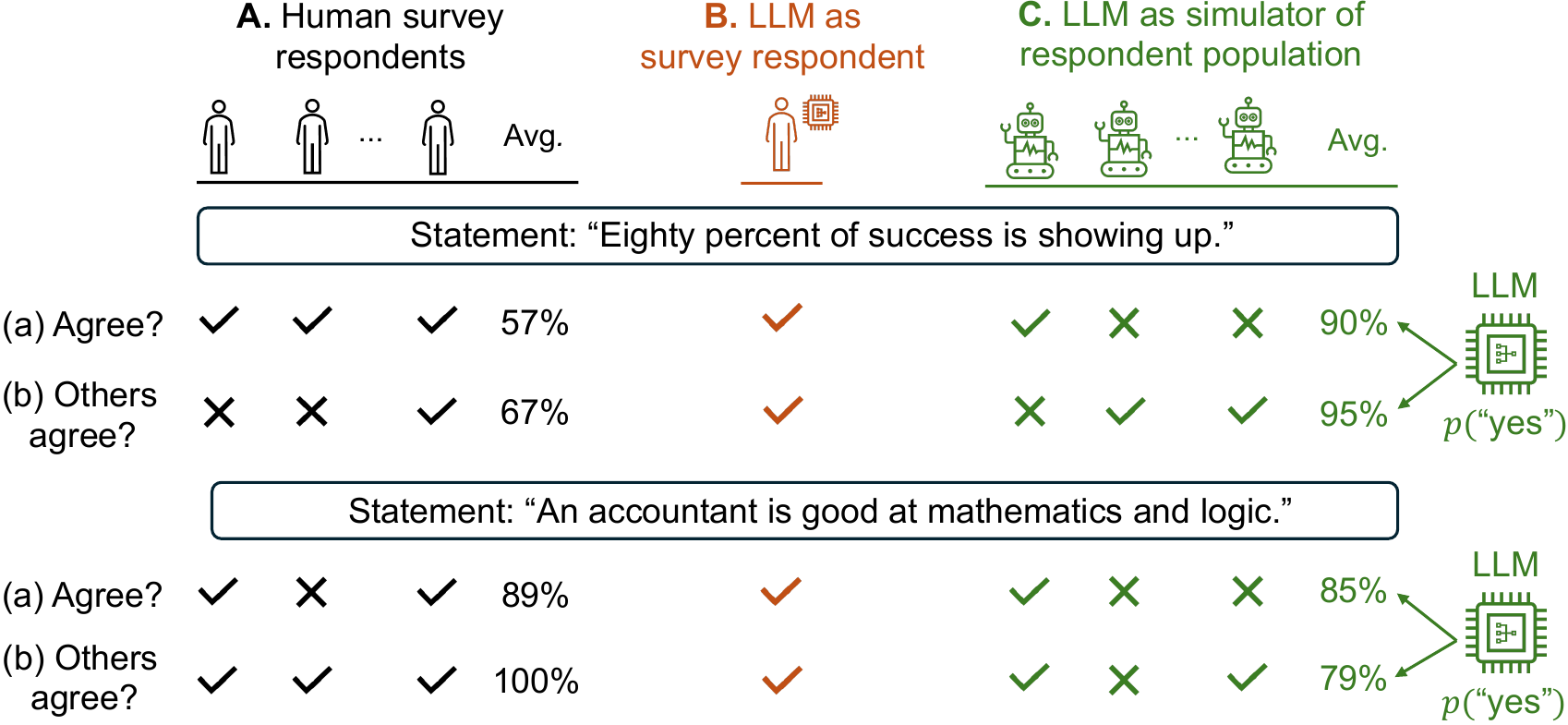}
    \caption{ 
    \textbf{Evaluation settings to measure the common sense of humans and large language models (LLMs).}
    For every statement, humans and LLMs are asked to indicate (a) whether they agree with it and (b) whether they think most other people would agree with it. 
    In panel (\textbf{A}), a total of \textit{N} = 2,046 human participants were recruited to perform this task. The ``Avg.'' column denotes the percentage of people who answered ``yes'' to the corresponding question. 
    In panel (\textbf{B}), we treat each LLM (in a total of \textit{N} = 35 models) as an independent survey respondent, just like every human in panel (A). This gives rise to the \emph{individual}-level view of common sense, in which this model is measured based on its agreement with the majority of other people on every statement.
    In panel (\textbf{C}), we treat every LLM's probability in its output answer as the average response of a hypothetical population of ``silicon samples'' (depicted as robots). For instance, if the LLM agrees with the statement ``Eighty percent of success is showing up'' with 90\% probability, we interpret this as 90\% of the silicon samples would agree with this statement. This gives rise to the \emph{statement}-level metric of common sense, which is used to measure the correlation between the human (panel A) and silicon sample (panel C) populations.}
    \label{fig:overview}
\end{figure*}

Progress in LLM common sense is frequently evaluated by standardized benchmarks~\citep{davisBenchmarksAutomatedCommonsense2024}. While their details vary, most existing benchmarks are conceptualized around the notion of \emph{correctness}: they assume humans apprehend matters of everyday reality in a uniform manner, and hence assess models by how accurately they recognize this ``ground truth.'' However, what different individuals may hold as trivial, commonsensical truths necessarily vary because their experience of the world is often highly subjective~\citep{schutzCommonSenseScientificInterpretation1953,bergerSocialConstructionReality1966,mertonSocialTheorySocial1968,fletcherPsychologyCommonSense1984,rosenfeldCommonSensePolitical2011}. Empirical research corroborates this view, showing that humans are extremely heterogeneous in their judgment even of simple, seemingly obvious propositions~\citep{whitingFrameworkQuantifyingIndividual2024}. For instance, when asked to evaluate the aphorism ``Eighty percent of success is showing up,'' illustrated in \subfigref{fig:overview}{A}, nearly half of the human respondents actually disagreed with it.  

Recent audits of AI commonsense benchmarks have found that ground-truth labels often achieve low agreement among independent data annotators, and up to a quarter of them are even contradicted after relabeling~\citep{yingBenchmarkingHumanLikeIntelligence2025}.
Thus, when an LLM achieves a high accuracy score, it appears \emph{only} similar to its benchmark designer. To another person who prescribes a vastly different set of ground truths---according to their own common sense---the same model could appear just as dubious. The challenge to evaluation, then, is in reconciling the aim for an informative metric with the pluralism in human judgment that permits almost no correct labels. 

In the human sciences, common sense is treated not as a set of irrefutable truths, but rather as a system of beliefs that are mutually upheld by members of a social community~\citep{geertzCommonSenseCultural1975}. In other words, a person has common sense only if their beliefs coincide with whatever their community holds collectively. This conception allows for an empirical investigation of commonsense knowledge, and thus motivates our contribution of a new LLM evaluation framework. We start with a population---of humans and of LLMs alike---and collect their judgments toward a number of statements, from which consensus arises. Common sense is then measured by the degree to which members of this population agree with one another about these statements. We elaborate this notion of \emph{collective agreement} via two major but logically independent uses of an LLM, illustrated in \Cref{fig:overview}.

First, an LLM can be viewed as an independent survey respondent that is evaluated on an individual basis (\subfigref{fig:overview}{B}). Under this role-playing paradigm~\citep{andreasLanguageModelsAgent2022,shanahanRolePlayLarge2023}, the respondent must both subjectively agree with the majority opinion of other people (\subfigref{fig:overview}{A}), and accurately predict this majority opinion regardless of what it subjectively holds. These two signals are combined to measure the commonsense competence of the LLM respondent, just like how a human would be scored. The results show that both humans (\textit{N} = 2,046) and models (\textit{N} = 35) vary significantly in this respect. While the highest-ranked model is rated as commonsensical as 64.5\% of recruited human participants, over two-thirds of LLMs are placed below the human median. Surprisingly, we find that smaller, open-weight models like Mistral-7B and Flan-T5-XXL are in fact comparable or even more competitive than larger, proprietary models like GPT-4/5 or Claude 3 Opus.

Second, an LLM can be viewed as a summarizer of social and cultural knowledge~\citep{farrellLargeAIModels2025,jordanCollectivistEconomicPerspective2025} and thus be evaluated by how well it reproduces the distribution of human beliefs via simulating a hypothetical collective~\citep{argyleOutOneMany2023,santurkarWhoseOpinionsLanguage2023,dillionCanAILanguage2023,hamalainenEvaluatingLargeLanguage2023,jansenEmployingLargeLanguage2023,grossmannAITransformationSocial2023,durmusMeasuringRepresentationSubjective2024}. In such a population, ``silicon samples'' (depicted as robots in \subfigref{fig:overview}{C}) provide ratings just like humans, and common sense is measured for each statement based on how widely agreed upon it is. Between the two populations in \subfigref{fig:overview}{A} and \subfigref{fig:overview}{C}, we find that commonsense scores for the same set of statements are modestly correlated (Pearson's $r$ up to $.43$), but this relationship remains significantly below the human internal reliability ($r = .60$). Some groups of silicon samples, moreover, exhibit traceable qualitative differences from humans; for instance, the population constructed by Gemini Pro 1.0 overwhelmingly associates common sense with figures of speech, while humans tend to see eye to eye much more often on simpler, literal expressions.

Current practices in AI commonsense evaluation have been subjected to several criticisms. For example, as a product of large-scale crowdsourcing efforts, benchmarks often contain noisy human annotations~\citep{kejriwalNoiseAuditHumanlabeled2024} or semantically incoherent stimuli~\citep{paltaPlausiblyProblematicQuestions2024}, thereby casting doubt on the reliability of their performance metrics. Our present argument, however, is that not only can humans be noisy in their judgments, they may also hold genuinely conflicting beliefs about what is self-evident. Developing human-like AI requires explicitly acknowledging this pluralism, incorporating it in benchmarking domains where ground-truth labels likely do not exist, and appropriately defining what human performance is~\citep{kejriwalCanAIHave2024,yingBenchmarkingHumanLikeIntelligence2025}. We address these concerns by contributing a framework to evaluate commonsense knowledge in LLMs that is grounded in its social basis, \ie, in correspondence with the judgments of a human group surveyed on a large scale. From an AI alignment perspective, our bottom-up analysis allows for a fine-grained, empirical assessment of the content of commonsense knowledge represented in intelligent machines, especially when they are situated in highly diverse social contexts where questions of cultural awareness often arise~\citep{pawarSurveyCulturalAwareness2024}.

\section*{Overview}
\label{sec:overview}

We use a dataset introduced in ref.~\citenum{whitingFrameworkQuantifyingIndividual2024} which contains \textit{N} = 4,407 statements in English taken from seven sources, including two AI corpora and two types of direct elicitation by online participants. Humans (\textit{N} = 2,046) were independently recruited on Amazon Mechanical Turk to rate these statements. The \mmref~and \siref{si:tab:human_descriptives} describe this human sample in more detail. Participants were shown one statement at a time and asked to indicate (a) whether they agreed with it and (b) whether they thought most other people would agree with it. For example, in \subfigref{fig:overview}{A}, 89\% of humans agreed with the statement ``An accountant is good at mathematics and logic,'' while 100\% of the same people believed that most others would agree with it. Each participant was assigned 50 statements selected at random, and on average every statement was labeled by 23 people. We determine a statement's human majority rating---agree or disagree with it, in response to question (a)---as the judgment that was held by at least half of those who were assigned to rate it (see the \mmref~and \siref{si:sec:commonsense_recap}).

We examine a total of 35 autoregressive LLMs including 10 proprietary, frontier models such as GPT-3.5/4/5, Claude 3, Gemini Pro 1.0 and Mistral-Large, as well as 25 open-weight models such as LLaMA-2/3 and Falcon.  Among the open-weight LLMs, the smallest model has 80M parameters, while the largest has 180B. The full list can be found in \siref{si:tab:model_list}. For each statement, we prompt a model with the same questions (a) and (b) above and record the probabilities with which it generates the tokens ``yes'' and ``no'' (depicted in the ``Avg.'' column in \subfigref{fig:overview}{C}). When binary answers are called for---such as in \subfigref{fig:overview}{B} where the LLM is treated as an individual agent---we choose the answer that is associated with the higher probability. Otherwise, these probabilities are interpreted as the frequencies with which the hypothetical silicon samples (\subfigref{fig:overview}{C}) would answer the same questions with a ``yes'' or ``no.'' The \mmref~and \siref{si:sec:lm_prompt} provide more information and precise definitions. \siref{si:fig:calibration_q1,si:fig:calibration_q2} further evaluate the calibration between a model's probability and the frequency of the same response among humans for several models.

\section*{LLMs as Independent Survey Respondents}
\label{sec:individual}

Upon its release, GPT-4 was demonstrated to score above 90\% of the Uniform Bar Examination takers~\citep{openaiGPT4TechnicalReport2023}. Notably, this evaluation is unambiguous for exam questions are designed to have correct answers---and humans and machines can be objectively measured by how accurately they select such answers. Unfortunately, as we have argued, this does not apply to common sense: what appears correct to someone may not be so to another.

\begin{figure*}[t!]
    \centering
    \makebox[\textwidth][c]{
    \includegraphics[width=1\linewidth]{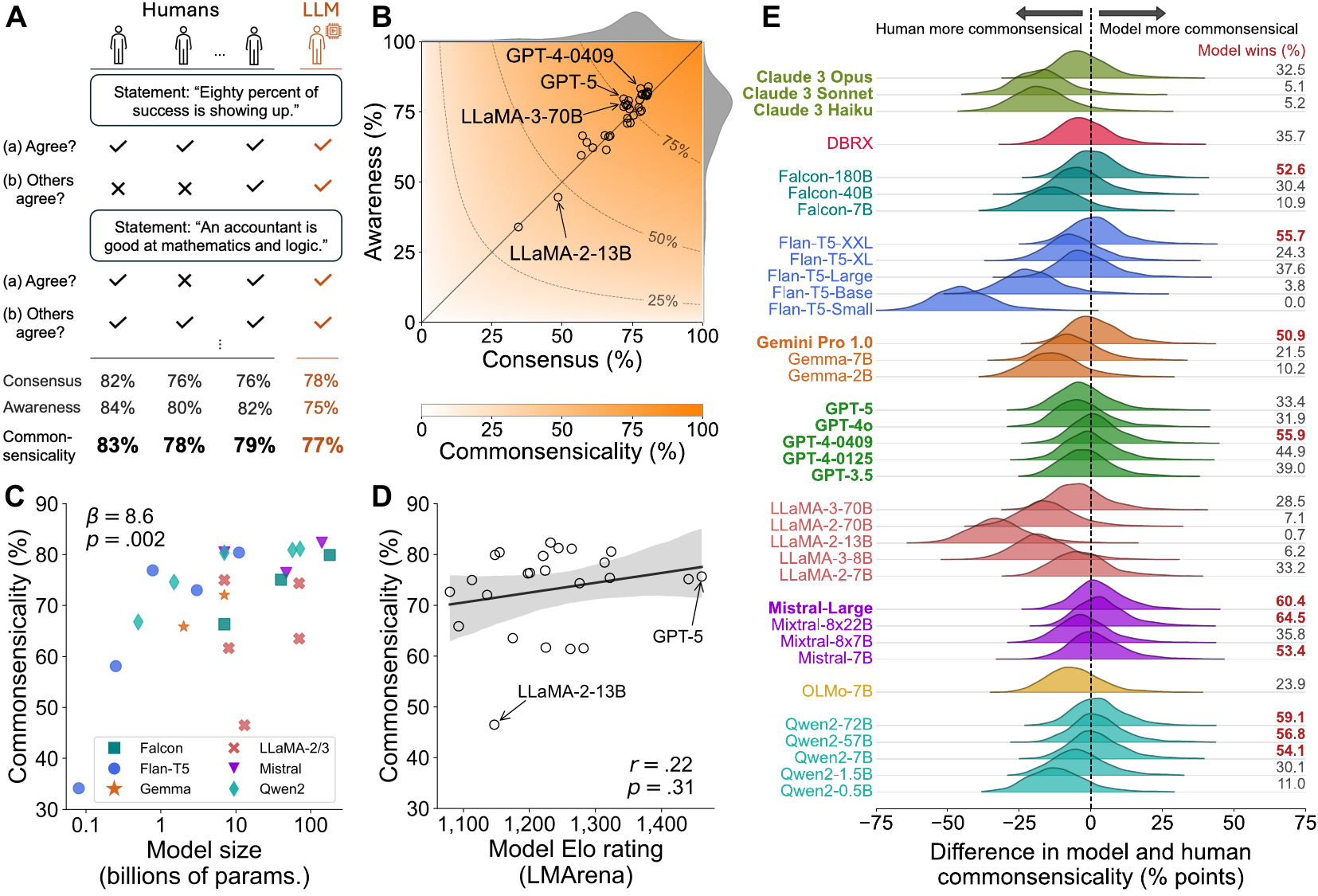}
    }
    \caption{\textbf{Individual-level commonsensicality of large language models.}
    Panel (\textbf{A}) shows conceptually how individual commonsensicality, defined for every human and model, is calculated based on their judgments of each candidate statement.
    Panel (\textbf{B}) shows each model's consensus and awareness scores. The level curves depict combinations of consensus and awareness that produce three different values of commonsensicality scores: 25\%, 50\% and 75\%.
    Panel (\textbf{C}) shows the relationship between a model's commonsensicality and its size, measured by the number of trainable parameters. Here, we only select six model families that each have at least two models of which we know the sizes. Also illustrated are the regression coefficient ($\beta$) and its two-sided \textit{p}-value, estimated using a linear mixed-effect model predicting commonsensicality using an LLM's (log-)size, grouped by model family, such as Flan-T5.
    Panel (\textbf{D}) shows the relationship between a model's commonsensicality and its Elo rating on the LMArena benchmark. Only 24 models with an Elo rating are shown in this figure. Pearson correlation $r$ and its two-sided \textit{p}-value are displayed. Correlation is also displayed by the best-fit line and a 95\% CI for the regression estimate (using 1,000 bootstrapped samples). 
    Panel (\textbf{E}) compares commonsensicality between humans and LLMs. The \textit{x}-axis represents the percentage-point difference in commonsensicality between a model and a person, where a positive difference indicates the model is more commonsensical. The \textit{y}-axis represents the kernel density of this difference. The ``model wins'' column to the right is the frequency with which a model is judged more commonsensical than a person, which equals the area under the density curve to the right of the vertical dashed line at 0. Closed-source models' names are in bold.}
    \label{fig:commonsensicality_comparison}
\end{figure*}

Here we describe an alternative LLM evaluation strategy, previously illustrated in \subfigref{fig:overview}{B} and now in more detail in \subfigref{fig:commonsensicality_comparison}{A}, that can overcome this ground-truth deficiency. Since there is no guarantee of a correct answer, the common sense of humans and models is simply measured by how much they agree with one another. In particular, we view each LLM as an independent survey respondent who, like every other human participant, provides binary ratings for every statement. The model's common sense is determined via two signals. First, does its judgment of a statement, via question (a), coincide with the human majority rating? Second, is its prediction of the human majority, via question (b), correct? Averaged over all \textit{N} = 4,407 statements, these signals respectively give us the model's \emph{consensus} and \emph{awareness} scores. We take the geometric average of these two scores and call the result the model's \emph{commonsensicality}. \subfigref{fig:commonsensicality_comparison}{B} presents these three scores for all 35 models. See the \mmref~and \siref{si:sec:commonsense_recap,si:sec:model_commonsensicality} for definitions and \siref{si:tab:model_individual_commonsensicality} for precise figures.

The notions of consensus and awareness are rooted in the functional analysis of common sense in sociology~\citep{lindenbergCommonSenseSocial1987}. More specifically, in order for members of a social group to meaningfully interact with one another, a baseline in the form of a shared body of knowledge---what they each know about the ordinary life they live in and share with others---must be achieved among them. Consensus is intended to measure this ``shared'' aspect. Whenever necessary in the course of social interaction, moreover, the same people must also be able to appeal to this shared knowledge and receive appropriate response from others. This requires not only knowing something, but also knowing that other people share that knowledge. This is what awareness aims to capture.

A model with high commonsensicality must both agree with the human majority (leading to high consensus) and accurately predict what most people think, regardless of its own judgment (high awareness). For example, in \subfigref{fig:commonsensicality_comparison}{A}, the statement ``Eighty percent of success is showing up'' received ratings from 21 human participants, 12 of whom (57 percent) agreed with it. To have high commonsensicality, the model is expected both to agree with this statement and to predict that most people would agree with it. At one extreme we have a maximum score of 100\%, where this always happens. A lower commonsensicality score could be driven by a model's lower consensus, awareness, or both. For instance, Claude 3 Opus disagrees with this statement, thereby lowering its consensus score, but it correctly predicts that most people would agree with it, thereby raising its awareness score. On average, this kind of divergence in a model's answers to questions (a) and (b) happens about 10 percent of the time, mostly in response to aphorisms; further example statements can be found in \siref{si:tab:nonconformity_statements}.

The commonsensicality score achieves two important goals. First, it requires no prior ground truth for any candidate statement: what is ``true'' is entirely determined by what people articulate as true.  Second, the commonsensicality score subjects humans and LLMs to the same calculation, for they perform the same rating task; this allows us to make empirically commensurable comparisons between them, as will be shown shortly. Our metric thus bears resemblance to the concept of cultural competence~\citep{romneyCultureConsensusTheory1986} or cultural consonance~\citep{dresslerCultureIndividualTheory2018}---the propensity of an individual to know what their society collectively deems appropriate---which often exhibits significant heterogeneity among that society's members. Note, however, that the majority vote as used here is based solely on human judgments, and thus humans may have a ``head start'' when compared to LLMs under this metric. \siref{si:sec:model_commonsensicality:alternative_calculation} introduces an alternative formulation addressing this concern and shows that the results reported here remain robust under this different setting.

We observe in \subfigref{fig:commonsensicality_comparison}{B} that most models lie close to the diagonal line, implying their consensus and awareness are roughly equal. This is largely because these models almost always give the same answer to questions (a) and (b). All three scores are left-skewed in \subfigref{fig:commonsensicality_comparison}{B}; for instance, the average commonsensicality score among models is 71.9\% (\textit{SD} = 10.6 percentage points). The most commonsensical LLM, with a score of 82.3\%, is Mixtral-8x22B, Mistral AI's open-weight model based on the mixture-of-expert architecture with 141B parameters. Other top-performing models include both open- and closed-source LLMs such as Mistral-Large (81.3\%, closed), Qwen2 (72B: 81.1\%, 57B: 80.9\%, open), GPT-4 (80.6\%, 0409 version, closed), and Flan-T5-XXL (80.4\%, open). The least commonsensical models also include open-weight and closed-source models such as Flan-T5 (Small: 34.2\%, Base: 58.1\%, open), LLaMA-2-13B (46.5\%, open), and Claude 3 (Haiku: 61.4\%, Sonnet: 61.6\%, closed).

It is generally observed that within the same LLM family (such as LLaMA), a model's benchmarking performance tends to increase with its size, or number of trainable parameters---a phenomenon dubbed the ``scaling law''~\citep{kaplanScalingLawsNeural2020}. We also find the same pattern on this commonsensicality test. \subfigref{fig:commonsensicality_comparison}{C} depicts this relationship for 23 models belonging to 6 families in our collection, each of which contains at least 2 models whose sizes are known. As can be noticed, this correlation is the most pronounced within the Flan-T5 and the least within the Mistral families. It should be noted, however, that the scaling law does not only apply to model size; another significant factor is the quantity and quality of their training data, which might explain the large variance in commonsensicality among models of roughly 7 billion parameters in the LLaMA-2/3, Qwen2 and Gemma families. To quantify this relationship while accounting for heterogeneity across model families, we fit a mixed-effect regression model with commonsensicality as the outcome, the logarithm of a model's size as the fixed effect, and the model family as the random effect. The result reveals that, on average, a ten-fold increase in an LLM's size is associated with an 8.6 percentage-point increase in its commonsensicality score ($p = .002$, 95\% CI: [3.4, 13.4]; see the \mmref~and \siref{si:sec:model_commonsensicality:effect_of_model_size}).

Interestingly, an LLM's commonsensicality does not seem to correlate with its general appeal to humans. \subfigref{fig:commonsensicality_comparison}{D} depicts the relationship between a model's commonsensicality and its Elo rating on the LMArena benchmark~\citep{chiangChatbotArenaOpen2024}. This rating is calculated based on human preference signals, where online users interacted with a pair of randomly chosen models and decided which one they preferred. We choose this benchmark because it is based on realistic, on-the-fly interactions between LLMs and humans, which likely involve a lot of everyday commonsense reasoning unlike other static benchmarks. If a model scores 100 points higher than another, then the former is preferred to the latter about 64\% of the time. See the \mmref~and \siref{si:sec:model_commonsensicality:effect_of_model_elo} for more detail. Surprisingly, we do not find any significant correlational evidence between the commonsensicality and Elo scores; Pearson's $r(22) = .22$, $p = .31$, 95\% CI: [-.21, .57]. With a large gap in Elo rating of 90, for example, GPT-4-0125 is preferred to GPT-3.5 about 63\% of the time by humans when pitched side by side. Yet their commonsensicality scores are very similar, at 78.4\% and 76.8\%, respectively. In \siref{si:sec:model_commonsensicality:openbookqa}, we perform this analysis with respect to an alternative commonsense reasoning benchmarking result~\citep{mihaylovCanSuitArmor2018}, and make the same observation.

Like LLMs, humans also vary considerably in their individual commonsensicality; \siref{si:fig:commonsensicality} presents the same version of \subfigref{fig:commonsensicality_comparison}{B} but for humans. Thus, it would be unreasonable to accept a singular level of ``human performance'' to which LLMs can be compared---an assumption made by most existing AI benchmarks. In light of this fact, we instead examine where exactly in this distribution of human commonsensicality scores each LLM is positioned. Note that during data collection, every human respondent was only tasked with labeling a random subset of 50 statements, whereas each model manages to label all 4,407 of them. For fairness, when comparing each LLM with every human participant, we restrict the calculation of commonsensicality to the subset of 50 statements that the participant was asked to label. The \mmref~and \siref{si:sec:commonsense_recap,si:sec:model_commonsensicality} provide more detail on this calculation.

\subfigref{fig:commonsensicality_comparison}{E} presents the result of this comparison. The \textit{x}-axis is the percentage-point difference between a model's and a human's commonsensicality scores; a positive difference indicates that the model is more commonsensical than the human. The \textit{y}-axis depicts the estimated density of this difference across all 2,046 humans. Therefore, the area under this curve, from 0 onward, represents the model's percentile in the distribution of human commonsensicality scores, which is also reported in the ``model wins'' column.

Relative to humans, most LLMs are modest in their individual-level common sense. Over two-thirds of models (25 out of 35) are placed below the human median, meaning they would be judged less competent than a participant chosen uniformly at random. For instance, Claude 3 Opus, a frontier model, is judged as competent as only about a third of humans. The rest of these LLMs, which are indeed above the human median, include both closed-source models like GPT-4 (0409 version), Gemini Pro 1.0, and Mistral-Large, and open-weight models such as Falcon-180B, Mixtral-8x22B, and Flan-T5-XXL. The highest-ranked model according to this metric is Mixtral-8x22B, which is rated above 64.5\% of human participants. Most surprisingly, Flan-T5-XXL is an LLM that was probably trained on orders of magnitude \emph{less} data than today's models, yet it is comparable to GPT-4-0409 and even ranked higher than Falcon-180B, a model that is about 16 times its size.

The LLaMA family is also a notable case. No LLM within this family manages to be more commonsensical than a third of humans. Moreover, even though the scaling law is observed in other model families, this is not the case for LLaMA-2. The highest-scoring model turns out to be the smallest variant, LLaMA-2-7B, with a winning rate of 33.2\%. This rate drops to almost zero for the 13B version before coming back to 7.1\% for the 70B variant. This observation is not even consistent with the inverse scaling law~\citep{mckenzieInverseScalingWhen2023,linTruthfulQAMeasuringHow2022}, where models are expected to be \emph{worse} in some metric as they grow in size. Although without further evidence, we suspect that the creation of LLaMA-2 models may not have been uniform, especially in terms of the data used to train them.

In summary, the commonsensicality score presented here measures an LLM's commonsense competence with respect to how often it agrees with humans. Unlike existing AI benchmarks, this evaluation is bottom-up: it is derived from the consensus of real humans as opposed to a researcher-prescribed correct label. Our metric can therefore be flexibly adapted to any other population of interest to analyze an LLM \emph{relative} to that population. Moreover, the commonsensicality metric formalizes the notion of human performance which, as we discussed above, displays a non-trivial variation. This is also in contrast with current benchmarks, as human performance is often thought to be constant and (close to) perfect. As a result, we do not interpret commonsense intelligence as an LLM's ability to pass a rigid, predefined accuracy threshold, but rather analyze this model by locating it in the heterogeneity of human competencies. Based on this, we have shown that smaller, open-weight models can be as competitive as their larger, proprietary, frontier counterparts.

\section*{LLMs as Aggregators of Human Opinion}
\label{sec:collective}

Recently, a growing body of work in the social sciences has demonstrated that the distribution of human opinions and attitudes in many empirical surveys can be approximated surprisingly well by the same answers sampled from LLMs. This includes quantities such as public opinion~\citep{santurkarWhoseOpinionsLanguage2023,durmusMeasuringRepresentationSubjective2024,kimAIAugmentedSurveysLeveraging2023,kozlowskiSilicoSociologyForecasting2024}, voting preferences~\citep{argyleOutOneMany2023}, social conventions~\citep{asheryEmergentSocialConventions2025}, moral judgments~\citep{dillionCanAILanguage2023} and economic decisions~\citep{hortonLargeLanguageModels2023,brandUsingLLMsMarket2023}. In explaining this remarkable correspondence, several authors have argued that since so many human individuals have contributed their micro-level knowledge to the LLMs' training data, these models can be viewed as a ``distilled form of crowdsourcing''~\citep{burtonHowLargeLanguage2024} capable of reproducing the diversity of this knowledge with great fidelity~\citep{grossmannAITransformationSocial2023,heymanImpactChatGPTHuman2023,bailCanGenerativeAI2024}.

We apply this perspective to the evaluation of LLM common sense, making two running assumptions. First, similar to voting preferences and moral judgments, cultural artifacts such as commonsense knowledge are also widely captured in LLMs' training data. And second, throughout training LLMs can effectively synthesize these artifacts in a way that represents the average human contributor of that data. Simply put, the more frequently a statement is endorsed by humans in the data, the more likely LLMs are to accept it. The goal of this section is thus to examine how well LLMs, when used as a generator of a hypothetical population, can approximate human commonsense knowledge as articulated by the human participant pool.

\begin{figure*}[t!]
    \centering
    \makebox[\textwidth][c]{
        \includegraphics[width=1\linewidth]{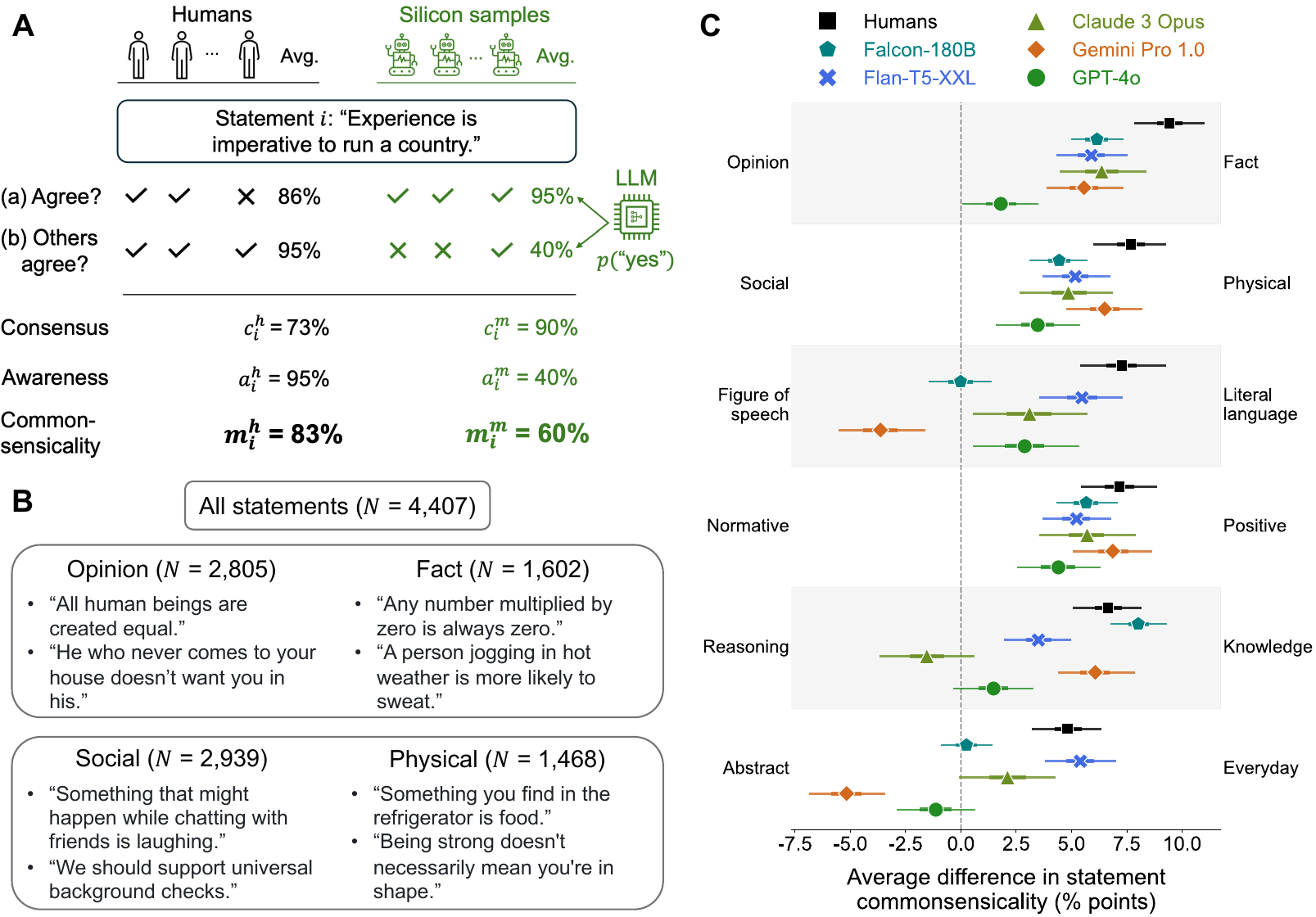}
    }
    \caption{
    \textbf{Commonsensicality of statements in different populations of raters.}
    In panel (\textbf{A}), we depict two populations: one consisting of real humans and one of silicon samples generated by repeatedly sampling the responses of GPT-3.5. For each population and every statement $i$, we measure two quantities: how close participants in that population are to a unanimous judgment of the statement (consensus, $c_i^h$ and $c_i^m$) and how well they can predict this majority opinion (awareness, $a_i^h$ and $a_i^m$). The statement's commonsensicality, $m_i^h$ and $m_i^m$, is the geometric average of its consensus and awareness scores. For example, the statement ``Experience is imperative to run a country'' is $m_i^h = 83\%$ commonsensical according to humans, but only $m_i^m = 60\%$ commonsensical according to GPT-3.5-simulated silicon samples. 
    Panel (\textbf{B}) illustrates some features of these statements in our corpus. Every statement is labeled as either an objective fact (\textit{N} = 1,602) or a subjective opinion (\textit{N} = 2,805); to describe either the physical world (\textit{N} = 1,468) or social reality (\textit{N} = 2,939). In total, there are six such dichotomies, which are described further in the \mmref~and \siref{si:sec:model_collective}.
    Panel (\textbf{C}) shows the difference in statement commonsensicality score within each dichotomy for several populations of raters. For instance, in the human population, each statement $i$ receives a commonsensicality score of $m_i^h$. The black square in the top row represents the average difference in this score between statements labeled as a fact and those labeled as an opinion. Therefore, to humans, facts are on average 9.38 points more commonsensical than statements. Thick and thin bars depict the 50\% and 95\% confidence intervals from 1,000 bootstraps.
    }
    \label{fig:statement_comm_features}
  \end{figure*}

Imagine a hypothetical society of ``silicon samples''~\citep{argyleOutOneMany2023}, previously illustrated in \subfigref{fig:overview}{C} and now in more detail in \subfigref{fig:statement_comm_features}{A}. For each statement, individuals in this society are tasked with answering the same questions (a) and (b) described in the \overviewref. These answers are generated by repeatedly sampling an LLM's response to the same prompt, taking advantage of its probabilistic nature. Assuming independence, the average response of a silicon sample in the limit is exactly the probability with which the LLM generates the ``yes'' token to a given question. Based on this signal, illustrated in the ``Avg.'' column of \subfigref{fig:statement_comm_features}{A}, we define a \emph{commonsensicality} score for the statement. Essentially, a higher score indicates that individuals in this population are close to unanimity in their judgment of this statement (thus increasing the statement's consensus) and that they can accurately predict this judgment in one another (thus increasing its awareness). Precise calculations can be found in the \mmref~and \siref{si:sec:model_collective}. Note that this \emph{statement}-level measure is separate from the \emph{individual}-level commonsensicality score (\cf~\subfigref{fig:commonsensicality_comparison}{A}) that was the subject of the previous section.

\subfigref{fig:statement_comm_features}{A} illustrates an example statement (indexed by $i$): ``Experience is imperative to run a country.'' Of the 22 human participants who were assigned statement $i$, 19 (or 86 percent) agreed with it, while 21 (or 95 percent) believed other people would agree with it. This statement accordingly receives a consensus score of $c_i^h = 73\%$, an awareness score of $a_i^h = 95\%$, and a commonsensicality score of $m_i^h = 83\%$; see the \mmref~for the calculation. Thus, within this human group, statement $i$ is 83\% commonsensical. On the other hand, in the population of silicon samples generated by GPT-3.5, depicted by robots, 95 percent of them agree with this statement, while 40 percent of them express that others would agree with it. By the same calculation, we arrive at statement $i$'s commonsensicality among the GPT 3.5-generated population, which is $m_i^m = 60\%$. 

\subsection*{Features of Common Sense in Statements between Humans and Silicon Samples}

What makes two populations of raters (\eg, humans and silicon samples) similar in their common sense? We propose to analyze the commonsensicality scores for the same set of statements between these groups. First, we look at what types of statements tend to attract high agreement in each population. Our corpus is accompanied by six epistemological features: Each statement was rated to depict either an objective fact or a subjective opinion; to use either literal language or a figure of speech; to be about either an abstract rule or a description of an everyday experience; and so on. \subfigref{fig:statement_comm_features}{B} gives some example statements with respect to two features; the \mmref~and \siref{si:sec:model_collective} provide a complete list as well as further examples.

For every population and each feature dichotomy, we compare the two groups of statements separated by that dichotomy with respect to their average commonsensicality scores. The results for humans and five silicon sample populations are shown in \subfigref{fig:statement_comm_features}{C}. For example, according to humans---represented by a black square at the top of the figure---statements rated as facts (\eg, ``The Pope is the leader of the Catholic Church'') are on average 9.38 points more commonsensical than statements rated as opinions (\eg, ``Never go on trips with anyone you do not love''); mean difference (MD) = 9.38, 95\% CI: [7.72, 10.93]. Among silicon samples generated by Falcon-180B (teal pentagon), this difference is 6.18 points (95\% CI: [4.86, 7.45]).

Thus, \subfigref{fig:statement_comm_features}{C} shows that human common sense exhibits very clear tendencies: it strongly favors facts over opinions, descriptions of physical over social realities, literal expressions over figures of speech, \etcetera When examining LLMs, we find that most of these tendencies are preserved by silicon samples. For instance, those generated by Flan-T5-XXL (blue crosses) are almost indistinguishable from humans across all six dichotomies. 

The silicon samples simulated by Falcon-180B are also similar to humans except along two dimensions. First, this population displays no significant preference between statements that use literal language (\eg, ``A cat doesn't want to get wet'') and those employing a figure of speech (\eg, ``Rudeness is the weak man's imitation of strength'') (MD = 0.03, 95\% CI: [-1.51, 1.46]). Second, it also does not favor statements that depict ordinary, everyday experiences (\eg, ``A grain of sand is very small'') over abstract rules of thumb or aphorisms (\eg, ``Morality is just a concept that can change depending on the situation'') (MD = 0.26, 95\% CI: [-0.87, 1.51]).

The three most popular and capable closed-source LLMs---Claude 3 Opus, Gemini Pro 1.0 and GPT-4o---all exhibit the same tendencies with humans in four out of six dichotomies. For instance, all three models generate a community that rates statements describing the physical world (``You are likely to find a shirt in closet'') as significantly more commonsensical than those about social experience (``Justice without force is powerless, force without justice is tyrannical''). GPT-4o's simulated population (green circle), however, does not differentiate between claims about everyday reality with abstract statements (MD = -1.11, 95\% CI: [-2.81, 0.65]). Whereas Gemini Pro 1.0 (orange diamond), like humans, favors statements that contain simple declarative knowledge about the world (\eg, ``Plants cannot survive without light'') over those that involve logical reasoning (\eg, ``If we ask for someone else to explain things they will think about it against their own justification''), Claude 3 Opus (green triangle) and GPT-4o do not display such distinction.

More starkly, we observe that the population generated by Gemini Pro 1.0 significantly diverges from humans in two dichotomies of knowledge. According to this simulated group, statements that use a figure of speech (\eg, ``The clash of ideas brings forth the spark of truth'') are on average 3.67 points (95\% CI: [1.76, 5.57]) more commonsensical than those that use more literal expressions (\eg, ``A person doesn't want a low paying job''). Similarly, abstract statements (\eg, ``Our greatest misfortunes come to us from ourselves'') are 5.17 points (95\% CI: [3.36, 6.78]) more commonsensical than statements describing everyday reality (\eg, ``The last thing you do when you take a shower is dry off''). Although the reason for these surprising findings is unclear, they may suggest a significant difference in Gemini Pro 1.0's training strategy, such as involving an over-representation of abstract figures of speech in its training set.

\subsection*{Correlation in Statement Commonsensicality between Humans and Silicon Samples}

We have seen in \Cref{fig:statement_comm_features} that LLM-simulated populations display largely similar qualitative tendencies to humans. Here, we provide a quantitative analysis of this correspondence. More specifically, if the populations of silicon samples and humans are similar with respect to their common sense, then the commonsensicality score of each statement should be similar in both populations. The ``fidelity'' of an LLM, therefore, can be measured by the Pearson correlation between $m_i^m$ and $m_i^h$ for all statements $i$.

First, to establish a baseline for the internal consistency of human judgments---given their subjectivity---we perform a resampling analysis. We repeatedly split the full human sample (\textit{N} = 2,046) into two random halves, calculate statement commonsensicality scores for each half, and compute the correlation in these scores between the two. Even among humans, the agreement on statement commonsensicality scores only reaches a moderate level; mean correlation $r = .60$ (95\% CI from 1,000 repetitions: $[.58, .61]$).

\begin{figure*}[t!]
    \centering
    \makebox[\textwidth][c]{
        \includegraphics[width=1\linewidth]{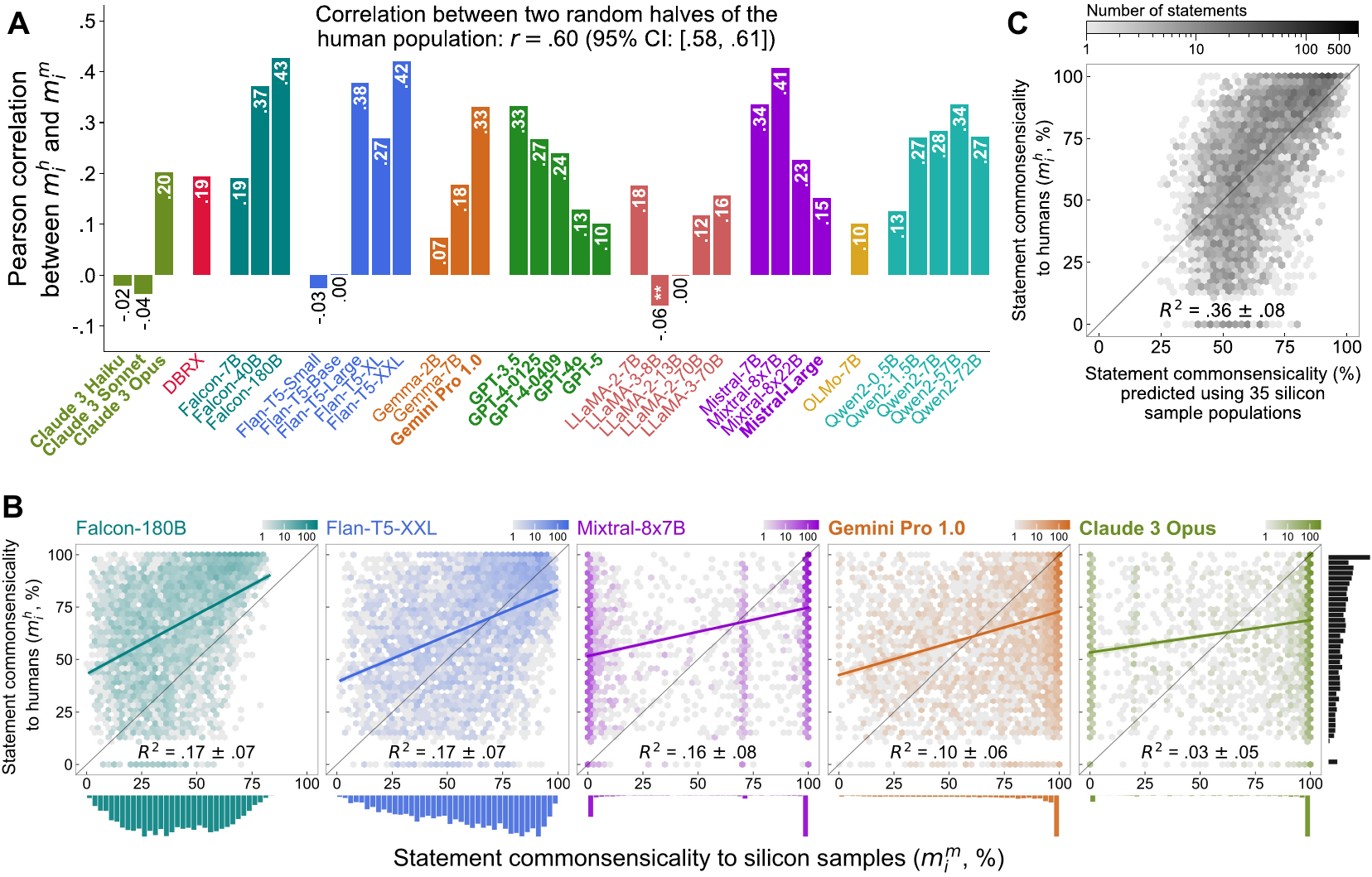} 
    }
    \caption{\textbf{Correspondence between human and silicon sample populations with respect to statement scores.}
    In each population, every statement receives a commonsensicality score. 
    Panel (\textbf{A}) shows the Pearson correlation between $m_i^h$, the score in humans, and $m_i^m$, the score in silicon samples, for all models. All positive correlations are significant at the $p < .001$ level. All negative correlations are insignificant with $p > .05$, except for LLaMA-3-8B ($p = .003$, depicted with ``**'' in the figure). \textit{p}-values are two-sided and Bonferroni corrected. As a baseline, we also display the same correlation between two randomly split subpopulations of humans, which is $r = .60$. The 95\% CI is derived from 1,000 repetitions of such splits.
    Panel (\textbf{B}) expands this correlation for some models. The shade of each hexagon represents its density, \ie, the number of statements within that hexagon. We also illustrate a best-fit line in each plot, predicting $m_i^h$ with $m_i^m$ using a linear regression model. The out-of-sample $R^2$ for this model (mean and standard deviation) is calculated using 50-fold cross-validation.
    In panel (\textbf{C}), we combine the statement commonsensicality scores $m_i^m$ in \emph{all} 35 silicon sample populations to predict the same score in humans, $m_i^h$, using a multiple regression model. We also report the out-of-sample $R^2$ for this model (mean and standard deviation), calculated using 50-fold cross-validation.
    }
    \label{fig:statement_comm_correlation}
\end{figure*}

\subfigref{fig:statement_comm_correlation}{A} shows the correlation in statement commonsensicality scores between humans and silicon samples generated by every model in our collection. The most faithfully representative models---Falcon-180B, Flan-T5-XXL and Mixtral-8x7B---correlate with people at the $r = .41$ to $.43$ level, well below the baseline of $r = .60$. Unsurprisingly, for some model families like Falcon and Gemini/Gemma, larger or newer models tend to be more representative. For instance, the correlation coefficient rises from .19 to .37 and .43 for Falcon-7B, 40B, and 180B, respectively. On the other hand, for some families like GPT-3.5/4 and Mistral, the opposite is observed. For example, the correlation decreases from .33 for the earliest model, GPT-3.5, down to .13 for the latest variant, GPT-4o. One model, LLaMA-3-8B, has a negative correlation with humans ($r = -.06$, $p = .003$), while five models (Claude 3 Haiku, Claude 3 Sonnet, Flan-T5-Small, Flan-T5-Base, and LLaMA-2-13B) demonstrate no significant correlation at all.

Five of these correlations are magnified in \subfigref{fig:statement_comm_correlation}{B}, from which we report two important observations. First, a considerable amount of variation in statement commonsensicality in humans ($m_i^h$) is not accounted for by the same score emerging from silicon samples ($m_i^m$). Using a linear model to regress $m_i^h$ via $m_i^m$ as the predictor, we depict the best-fit lines as well as their out-of-sample coefficients of determination ($R^2$) here. Only up to 17\% of the variation in statement commonsensicality score according to humans can be explained, which is the case for Falcon-180B and Flan-T5-XXL. This figure is 10\%, 3\% and 1\% for Gemini Pro 1.0, Claude 3 Opus and GPT-4o, respectively (the last of which is not shown in \subfigref{fig:statement_comm_features}{B}).\footnote{In \siref{si:fig:statement_comm_errors_with_humans}, we illustrate this point further by showing in greater detail the difference between $m_i^m$ and $m_i^h$. In particular, we calculate the mean absolute error (MAE) between these two scores, which represents the average difference in statement commonsensicality between humans and silicon samples. The highest-ranked model is Flan-T5-XXL, with the smallest MAE of 21.41. This means that on average, the commonsensicality score of a statement, according to Flan-T5-XXL's silicon samples, is 21.41 points away from the score of the same statement according to humans. This error is the highest for Falcon-40B, with an MAE of 52.69.}

Second, the best-fit lines, as well as the marginal distributions in \subfigref{fig:statement_comm_correlation}{B}, suggest some unique patterns among these silicon sample populations. For example, within the population simulated by Falcon-180B, statements are on average less commonsensical than they are to humans (most scores lie above the diagonal perfect-calibration line). In addition, while they have roughly the same correlation with humans, for Mixtral-8x7B statement commonsensicality scores tend to be collapsed to polar values of 0\% and 100\%, whereas for Falcon-180B and Flan-T5-XXL they tend to spread more evenly.

Similar to Mixtral-8x7B, the tendency of statement commonsensicality to be close to extreme values is common among many models like Gemini 1.0 and Claude 3 Opus, also shown in \subfigref{fig:statement_comm_correlation}{B}. In other words, statements are either very commonsensical or the opposite according to the ratings by the simulated populations from these models. This is partly a result of LLMs being extremely confident in their answers to our prompts by outputting probabilities close to 0 or 1. (Contrast this with what we observe in humans, depicted in the black histogram to the right end of \subfigref{fig:statement_comm_correlation}{B}.)  While the cause of this ``uniformity''~\citep{kozlowskiSimulatingSubjectsPromise2025} in their answers is uncertain, it is likely because these models have gone through extensive instruction fine-tuning and human alignment, two phases of training that tend to collapse their outputs toward singular choices associated with high rewards~\citep{baiTrainingHelpfulHarmless2022,kadavathLanguageModelsMostly2022,openaiGPT4TechnicalReport2023}---thereby sacrificing their distributional representativeness.

Finally, we explore the informativeness of \emph{all} models at once by performing a regression analysis predicting a statement's score in humans using the same 35 scores in all silicon sample populations. As can be seen in \subfigref{fig:statement_comm_correlation}{C}, taking an ensemble of all models does indeed improve the prediction of statement scores; average out-of-sample $R^2 = .36$. 

In summary, this section views every LLM as an aggregator of human knowledge on a large scale, and tests whether the population simulated by this model is an adequate representation of real humans. Once again, the heterogeneity in human commonsense beliefs is evidenced by a moderate split-half reliability of $r = .60$, but the best model-generated silicon samples can only correlate with humans at $r = .43$, well below this baseline. In addition, while a statement's score in a silicon-sample population is predictive of its score in the human population, the majority of the variation in the latter is left unaccounted for.

\section*{Discussion}
\label{sec:discussion}

Designing machines that exhibit human-like tendencies in both individual and social environments is a central objective of AI research~\citep{rahwanMachineBehaviour2019,collinsBuildingMachinesThat2024,tsvetkovaNewSociologyHumans2024}. This often requires AI models to attain some form of commonsense intelligence~\citep{gweonSociallyIntelligentMachines2023}, of which commonsense knowledge is an important constituent~\citep{davisCommonsenseReasoningCommonsense2015}. While standardized benchmarks are a straightforward method to evaluate progress in this area, we argue that the notion of invariant ground truth---widely assumed but rarely articulated in benchmarking practices---has likely misguided AI researchers, creating an illusion of success as LLMs continue to pass a growing number of tests (\eg, refs.~\citenum{levesqueWinogradSchemaChallenge2012,mihaylovCanSuitArmor2018,talmorCommonsenseQAQuestionAnswering2019,zellersHellaSwagCanMachine2019}).
 
Our argument is motivated by the heterogeneity in the content of human commonsense knowledge. Since a correct label is unlikely to exist in this domain, what a model ``gets right'' may be meaningful to some people but not so much to others~\citep{nguyenExtractingCulturalCommonsense2023,huangCulturallyAwareNatural2023}. This ``one-truth myth''~\citep{aroyoTruthLieCrowd2015} manifests itself in many domains involving subjective interpretation~\citep{almSubjectiveNaturalLanguage2011} such as toxicity detection~\citep{goyalYourToxicityMy2022}, image classification~\citep{miceliSubjectivityImpositionPower2020}, moral judgment~\citep{fraserDoesMoralCode2022,sorensenValueKaleidoscopeEngaging2024,jiangInvestigatingMachineMoral2025}, rhetoric decoding~\citep{mendelsohnDogwhistlesBullhornsUnveiling2023}, and even medical relations understanding~\citep{aroyoTruthLieCrowd2015}. Several works, in response, have called for treating empirically observed human judgments as a normative standard for human-like AI benchmarking~\citep{pelletierWhatShouldDefault1997,davaniDealingDisagreementsLooking2022,plankProblemHumanLabel2022,cabitzaPerspectivistTurnGround2023,yingBenchmarkingHumanLikeIntelligence2025}. Following this direction, we propose two interpretations of LLMs---as independent survey participants and as generators a silicon population---and evaluate the degree to which their elicited ratings correspond to the distribution of human opinions observed on a large scale.

The proposed evaluation framework rests on several assumptions and design choices. For example, to obtain an LLM's opinion distribution, we rely on the probabilities of its generated tokens. Other approaches, such as repeated sampling or verbalization~\citep{meisterBenchmarkingDistributionalAlignment2025}, have shown some comparative success but are not employed in this paper. Regardless, there is an underlying frequentist assumption that the model's internal confidence in a statement is to some extent calibrated to the empirical frequency with which humans---especially those who have contributed to its training dataset---would endorse the same statement. This is not theoretically substantiated but is rather an empirical demonstration among recent work that studies how LLMs can be used to approximate public opinion~\citep{santurkarWhoseOpinionsLanguage2023,jansenEmployingLargeLanguage2023,durmusMeasuringRepresentationSubjective2024,kozlowskiSilicoSociologyForecasting2024}. In addition, a model's calibration to the actual frequency of a belief can be controlled by its sampling temperature. In our case, this parameter is set to 1.0 (see the \mmref) because we are primarily concerned with evaluating the model in its default setting rather than fine-tuning it for out-of-sample prediction.

We also posit that the most significant factor that predisposes an LLM toward a commonsense belief is its training data. However, another driver that may explain the model's response is its inherent sycophantic tendency: rather than being faithful to the data, it could simply be attempting to look for whatever answer it thinks the user wants. Although it is difficult to isolate either of these two drivers, we can at least inspect what happens if we further clarify to the model the role it is supposed to play. The results presented in \siref{si:sec:other_prompts}, which include the judgments and reasoning by some GPT models in response to several prompting variants, suggest that these models are likely role-playing independently rather than making an effort to impute what their user may believe. While not attempted in this work, an in-depth analysis of LLM reasoning can also help unpack certain behaviors of these models, such as when they subjectively endorse a statement but predict that most others would not.

Human and model common sense, in addition, depends on what kinds of statements they are asked to rate. In \siref{si:sec:statement_subsets_by_feature}, we show that the calculation of commonsensicality is sensitive to different features of a statement. For instance, almost all models appear significantly more commonsensical if they only rate statements that depict a fact or describe the physical world. The opposite is observed if these statements use a figure of speech or describe the social world instead. Our corpus of choice, moreover, is likely not representative of the total stock of human commonsense knowledge. Some areas posited to be in this domain---\eg, theory of mind inferences~\citep{kosinskiEvaluatingLargeLanguage2024,strachanTestingTheoryMind2024}---have received increasing attention from the AI community but are largely missing in our context. Thus, the performance metrics reported here may only reflect certain parts of a much bigger picture of human and machine common sense. Curating a larger statement collection, guided by an established taxonomy (\eg, refs.~\citenum{ilievskiDimensionsCommonsenseKnowledge2021,davisBenchmarksAutomatedCommonsense2024}), spanning different languages, and involving more concrete tasks, will address this potential incomprehensiveness. As we have remarked earlier, however, our model assessment methodology is agnostic to the very content of such statements, and therefore can be readily applied to any new corpus used in future work.

Beyond these considerations, it is worth reiterating that the term ``humans'' used in this work refers more precisely to the group of English-speaking participants recruited on Amazon Mechanical Turk. In focusing only on their aggregated ratings, our framework does not take into account the potential importance of any cognitive and behavioral diversity among individuals in this sample. Past research, however, has found that social perceptiveness---the ability of a person to ``read'' other people's emotions---can reliably predict commonsensicality~\citep{whitingFrameworkQuantifyingIndividual2024}. Thus, future work can consider enhancing the fidelity of LLM simulations by having this variable encoded in a model's persona when it engages in role-playing~\citep{shanahanRolePlayLarge2023,argyleOutOneMany2023,kozlowskiSilicoSociologyForecasting2024,parkGenerativeAgentSimulations2024,jiangPersonaLLMInvestigatingAbility2024,anthisLLMSocialSimulations2025}, although this method is not without caveats~\citep{bisbeeSyntheticReplacementsHuman2024,huQuantifyingPersonaEffect2024,wangLargeLanguageModels2025,gaoTakeCautionUsing2025,lymanBalancingLargeLanguage2025}. This work additionally leaves an open question of what systematic characteristics of a group and its social processes can determine the content of its shared knowledge. Thus, it is unclear, for example, how the commonsensicality of GPT-4o (which ranks below two-thirds of humans in our MTurk sample) would change if it were considered only alongside American college students. Significant depth can be added to future work by exploring elements of commonsense knowledge that are (near-)unanimously shared, or, when they are only locally shared, how exactly they vary from one group to another~\citep{sperberExplainingCultureNaturalistic1996,zerubavelSocialMindscapesInvitation1997}---and how the common sense of an AI model is expected to change accordingly.

Throughout history and across discussion spheres, the term ``common sense'' is notably overloaded~\citep{rosenfeldCommonSensePolitical2011,wattsEverythingObviousOnce2011,davisCommonsenseReasoningCommonsense2015}. Our exposition, in particular, views AI common sense \emph{only} as the ability to synthesize statistical regularities in an LLM's training data, then assesses if this synthesis can lead to a good approximation of the heterogeneity of human knowledge observed in reality. This approach is from an entirely pattern recognition point of view. By contrast, a primary focus of cognitive psychology is to devise, and replicate in AI, a formal mechanism for human commonsense capabilities~(\eg, ref.~\citenum{jara-ettingerNaiveUtilityCalculus2016}). This model-building approach, which addresses \emph{why} humans and LLMs respond the way they do to the statements~\citep{lakeBuildingMachinesThat2017}, is not attempted in this paper. Thus, the foregoing results are necessarily limited to the content of commonsense knowledge, as articulated by the human participants on MTurk, and are not intended to suggest any individual-level cognitive mechanism for commonsense psychology, either in humans or in LLMs.

\section*{Methods}
\label{sec:materials}

We use the dataset in ref.~\citenum{whitingFrameworkQuantifyingIndividual2024} which contains \textit{N} = 4,407 statements sourced from seven domains: the news media via Google News (\textit{N} = 290), political campaign emails during U.S. elections~\citep[\textit{N} = 668]{mathurManipulativeTacticsAre2023}, AI corpora like ConceptNet~\citep[\textit{N} = 581]{speerConceptNet55Open2017} and ATOMIC~\citep[\textit{N} = 697]{hwangCOMETATOMIC2020Symbolic2021}, aphorisms taken from books (\textit{N} = 709), and statements elicited by online participants---either via completing a short sentence (\textit{N} = 630) or in response to a prompted domain of knowledge (\textit{N} = 832). See \siref{si:tab:statement_sources} for some examples. Most statements are short with a median of 11 words. Human participants (\textit{N} = 2,046) were recruited from Amazon Mechanical Turk to rate these statements. The human sample contains English-speaking residents in the United States, is 44 years old on average, and has more female (51.2\%) than male (48.3\%) participants (\siref{si:tab:human_descriptives} presents more descriptive statistics). Given each statement, the human workers answered the following questions: (a) ``Do you agree with this statement?'' and (b) ``Do you think most other people would agree with this statement?'' The term ``other people'' in question (b) was not intended to refer to any specific population, so its interpretation was entirely up to the human raters (and LLMs by extension). Each participant was tasked with labeling 50 randomly chosen questions, and on average each statement received ratings from 23 participants. The human data collection protocol was reviewed by the University of Pennsylvania Institutional Review Board (IRB, protocol 849601) which determined that the proposal met eligibility criteria for IRB review exemption. All human subjects provided informed consent before participating.

\subsection*{Large Language Models and Their Ratings}
We make use of 35 widely used instruction-finetuned large language models. These include open-weight models (\ie, those whose weights are openly accessible) and closed-source models. The full list can be found in \siref{si:tab:model_list}. All open-weight models are loaded via the Hugging Face library in Python. For closed-source models, we directly use the APIs provided by their creators.

For every statement, we ask each LLM the same two questions (a) and (b). Each question is asked in a separate chat session, in order to eliminate any influence of the chat history on the answer. We design each prompt so that the model's answer is expected to start with a definitive ``yes'' or ``no.'' Exact prompts and conversation settings are the same for all models and can be found in \siref{si:sec:lm_prompt}. 

To generate an answer, the model uses a parametric probability distribution over its vocabulary to sample new tokens, one at a time. We extract the probabilities of the tokens ``yes'' and ``no'' and discard all other tokens, then rescale these two probabilities so they add up to one. For all open-weight models, this distribution can be accessed directly. For GPT-3.5/4, we can access the probabilities for up to the top 20 tokens. For Gemini, Claude 3, Mistral-Large and GPT-5, we perform repeated sampling, where we ask a model the same questions multiple times and report the empirical frequencies of their generated answers. See \siref{si:sec:lm_prompt}.

The probability that the model answers ``yes'' to question (a) is called its rating distribution; see \siref{si:eq:model_ratings_dist}. This can be considered as the model's inherent confidence in agreeing with a statement. \siref{si:fig:calibration_q1,si:fig:calibration_q2} examine the calibration between this confidence and the relative frequency with which humans would give the same response for some models. The output probability is also crucially dependent on the sampling temperature. It is possible that the model can be more calibrated using a different temperature, but our goal is to evaluate it in its default mode, corresponding to using the temperature of 1.0. When binary decisions are called for, we choose the answer---``yes'' or ``no''---that is associated with the higher probability, also called the ``argmax'' of the model's rating distribution. These prompting settings were preregistered prior to data collection on AsPredicted, project number 162475~\citep{nguyenTestingCommonSense2024}.

\subsection*{Measuring LLMs' Individual Commonsensicality}

We adopt the intuition in ref.~\citenum{whitingFrameworkQuantifyingIndividual2024} and calculate an LLM's degree of common sense as if it were a real human participant (see \subfigref{fig:commonsensicality_comparison}{A}). In particular, to have common sense, the model must agree with the human majority opinion, and correctly predict this consensus irrespective of its subjective opinion. These two criteria are reflected in the model's answer to questions (a) and (b) above, respectively.

Precisely, for every statement let the human majority rating (agree or disagree) be the opinion held by at least half of the participants. If the model's answer to question (a) (whether it agrees with the statement) coincides with this majority rating, this counts as a correct answer. The accuracy of the model, averaged over all 4,407 statements, is called its \emph{consensus} score, which depicts how often its subjective rating agrees with the human majority. Similarly, the same accuracy with respect to the model's answer to question (b) (whether it thinks most people would agree with this statement) is called its \emph{awareness} score, representing how often it predicts the position of the human majority, notwithstanding how it subjectively rates a statement previously via question (a). We combine a model's consensus and awareness scores by taking their geometric average into its \emph{commonsensicality} score. See \siref{si:eq:model_consensus,si:eq:model_awareness,si:eq:model_commonsensicality}. These three scores, each ranging between 0 and 100\%, are presented in \subfigref{fig:commonsensicality_comparison}{B} above.

To analyze the relationship between a model's size and its commonsensicality score, we choose 23 models from 6 model families in our collection: Falcon, Flan-T5, Gemma, LLaMA-2/3, Mistral and Qwen2. These are families that contain at least 2 models whose sizes are available to us. The models are presented in \subfigref{fig:commonsensicality_comparison}{C}. We then perform a linear mixed-effects regression analysis with commonsensicality as the outcome, model size as the fixed effect and model family as the random effect. A detailed setting can be found in \siref{si:eq:model_commonsensicality_vs_size_mixed_eff}. 

To investigate the relationship between an LLM's general performance on a popular benchmark and its commonsensicality, we extract the LMArena Elo scores~\citep{chiangChatbotArenaOpen2024}, last updated on August 16, 2025. Essentially, LMArena tests a total of 230 LLMs side by side and uses crowd-sourced human ratings to rank these models. A participating user interacts with two randomly chosen LLMs at the same time, asking them the same question and receiving answers from both. Then, the user indicates which model is better. From approximately 3.87 million such comparisons, models are assigned an Elo score that depicts their relative performance. More detail can be found in \siref{si:sec:model_commonsensicality}, especially \eqref{si:eq:elo_interp}. We are able to find Elo scores for 24 models, and in \Cref{fig:commonsensicality_comparison}{D}, we report the Pearson correlation between a model's Elo score and its commonsensicality. 

Since the calculation of model commonsensicality is exactly how human participants are measured on their common sense---see \siref{si:eq:human_consensus,si:eq:human_awareness,si:eq:human_commonsensicality}---this allows us to compare every LLM against every human directly. One major difference is that while every model can label all 4,407 statements, every person was only asked to label a random subset of 50 statements. To make a fair comparison, for each LLM-human pair, we restrict the calculation of their commonsensicality scores to the 50 questions that the participant previously rated. More detail can be found in \siref{si:sec:model_commonsensicality}. \subfigref{fig:commonsensicality_comparison}{E} depicts how often a model is judged to be more commonsensical than a human.

\subsection*{Analyzing LLMs as Simulators of Silicon Sample Populations}

Recall that when we ask an LLM questions (a) and (b) above, we record the probability with which it answers ``yes'' or ``no.'' In the limit, this probability can be interpreted as the frequency of answers by the silicon samples simulated by this model. For example, if an LLM answers ``yes'' with 70\% probability, then among 1,000 silicon samples that are generated by this model, 700 of them are expected to say ``yes.''

For each population (humans or silicon samples), we analyze common sense on the statement level. Here we describe this calculation with respect to the human population, more details of which can be found in \siref{si:sec:commonsense_recap}. Let $d_i^{h, a}$, called the human rating distribution, be the proportion of people who indicated they agreed with statement $i$, via question (a). The majority opinion for this statement is then $\texttt{majority}_i^h = \mathbbm{1}\left[ d_i^{h, a} \geq 0.5 \right]$, which is 1 if at least half of the people agreed with it, and 0 otherwise. The consensus score measures how far people are from a unanimous opinion: $c_i^h = 2 \times |d_{i}^{h, a} - 0.5|$. It is 0 when exactly half of the people agree with it, and 1 when everyone agrees or disagrees with it. The awareness score measures how accurately people predicted the majority rating via question (b). If the $\texttt{majority}_i^h = 1$, then the statement's awareness, $a_i^h$, is the proportion of people who predicted most others would agree; otherwise, it is the proportion who predicted most others would disagree. The commonsensicality score is defined as $m_i^h = \sqrt{c_i^h \times a_i^h}$. We denote the same scores for the model-simulated population with the superscript $m$ instead of $h$, such as $m_i^m$.

For example, in \subfigref{fig:statement_comm_features}{A}, we observe that 86\% of humans agree with the statement, ``Experience is imperative to run a country,'' while 95\% of the same people thought most others would agree with it. Hence, $d_i^{h, a} = 0.86$, which means the majority rating is 1, or ``agree.'' The consensus score is thus $c_i^h = 2 \times |0.86 - 0.5| = 0.73$, and the awareness score is $a_i^h = 0.95$, because 95\% of the people correctly predicted that it would receive an ``agree.'' The commonsensicality score is, therefore, $m_i^h = \sqrt{0.73 \times 0.95} = 0.83$. The same scores ($c_i^m$, $a_i^m$ and $m_i^m$) are identically defined for the population of LLM-based silicon samples.

Each statement in our corpus was labeled using six epistemological dichotomies. Online participants were asked to rate whether a statement depicts a \emph{fact} (something that can be objectively demonstrated to be true) \versus~an \emph{opinion} (something that might be held true by some but not by others); describes \emph{physical} reality (objective features of the world according to, say, physics and biology) \versus~\emph{social} beliefs, perceptions and rules that govern human experience; uses \emph{literal language} that means exactly what it says \versus~a \emph{figure of speech} such as an aphorism or metaphor; conveys \emph{positive} empirical regularities in the world (\eg, ``Hot things will burn you.'') \versus~a \emph{normative} judgment, belief, value or convention (\eg, ``Treat others how you want them to treat you.''); conveys declarative \emph{knowledge} about the world \versus~a \emph{reasoning} that involves both knowledge and logic; and depicts \emph{everyday} experiences of humans \versus~\emph{abstract} rules and regularities that must be synthesized from experience. See \siref{si:sec:model_collective} for more detail.

In \subfigref{fig:statement_comm_features}{C}, we calculate the reported differences as follows. First, we separate the entire corpus of \textit{N} = 4,407 into two categories: those that depict a fact and those that portray a personal opinion. Each statement, again, has a commonsensicality score $m_i^h$ (assuming we are considering the human population). We calculate the average difference in this score between the fact-like statements and the opinion-like statements. To quantify uncertainty, we perform this calculation via 1,000 random bootstraps. The mean difference and the 95\% bootstrapped CI are depicted in the figure. This applies to other dichotomies as well as the populations of silicon samples by LLMs.

In \subfigref{fig:statement_comm_correlation}{A}, we compute the Pearson correlation between $m_i^h$ and $m_i^m$---the same commonsensicality scores for every statement within two populations: humans \versus~silicon samples. All two-sided \textit{p}-values are Bonferroni-corrected. In \subfigref{fig:statement_comm_correlation}{B}, we show this result in more detail via scatter plots for five models: Falcon-180B, Flan-T5-XXL, Mixtral-8x7B, Gemini Pro 1.0 and Claude 3 Opus. We also fit a linear model, predicting $m_i^h$ using $m_i^m$, and plot the best-fit line within these plots. The out-of-sample coefficient of determination, $R^2$, for this linear model is also depicted. Finally, in \subfigref{fig:statement_comm_correlation}{C}, we combine all 35 statement scores $m_i^m$ in 35 silicon sample populations to predict the same score in humans, $m_i^h$.

\section*{Data Availability}

All data and code to reproduce the results in this article have been deposited to Zenodo~\citep{nguyenLargescaleEvaluationCommonsense2026} and can also be found on GitHub at \url{https://github.com/Watts-Lab/commonsense-llm-eval}.

\footnotesize
\sf
\bibliographystyle{pnas-new-custom}
\bibliography{common-sense.bib}

\begin{thebibliography}{10}

\bibitem{whitingFrameworkQuantifyingIndividual2024SI}
M.~E. Whiting, D.~J. Watts,  A Framework for Quantifying Individual and Collective Common Sense.
\newblock {\em Proceedings of the National Academy of Sciences} \textbf{121}(4): e2309535121 (2024).

\bibitem{baron-cohenReadingMindEyes2001SI}
S. Baron-Cohen, S. Wheelwright, J. Hill, Y. Raste, I. Plumb,  The ``{{Reading}} the {{Mind}} in the {{Eyes}}'' {{Test Revised Version}}: {{A Study}} with {{Normal Adults}}, and {{Adults}} with {{Asperger Syndrome}} or {{High}}-functioning {{Autism}}.
\newblock {\em Journal of Child Psychology and Psychiatry} \textbf{42}(2): 241--251 (2001).

\bibitem{bialekCognitiveReflectionTest2018SI}
M. Bialek, G. Pennycook,  The Cognitive Reflection Test Is Robust to Multiple Exposures.
\newblock {\em Behavior Research Methods} \textbf{50}(5): 1953--1959 (2018).

\bibitem{sapAtomicAtlasMachine2019SI}
M. Sap, \emph{et~al.},  Atomic: {{An Atlas}} of {{Machine Commonsense}} for {{If-Then Reasoning}}.
\newblock {\em Proceedings of the AAAI Conference on Artificial Intelligence} \textbf{33}(1): 3027--3035 (2019).

\bibitem{anthropicClaude3Model2024SI}
Anthropic,  The {{Claude}} 3 {{Model Family}}: {{Opus}}, {{Sonnet}}, {{Haiku}}. (2024).
\newblock \url{https://www-cdn.anthropic.com/de8ba9b01c9ab7cbabf5c33b80b7bbc618857627/Model_Card_Claude_3.pdf} (Accessed 2024-05-28).

\bibitem{mosaicIntroducingDBRXNew2024SI}
Mosaic,  Introducing {{DBRX}}: {{A New State-of-the-Art Open LLM}}. (2024).
\newblock \url{https://www.databricks.com/blog/introducing-dbrx-new-state-art-open-llm} (Accessed 2024-06-15).

\bibitem{almazroueiFalconSeriesOpen2023SI}
E. Almazrouei, \emph{et~al.},  The {{Falcon Series}} of {{Open Language Models}}.
\newblock arXiv [Preprint] (2023).
\newblock \url{https://doi.org/10.48550/arXiv.2311.16867} (Accessed 2024-01-01).

\bibitem{chungScalingInstructionfinetunedLanguage2022SI}
H.~W. Chung, \emph{et~al.},  Scaling Instruction-Finetuned Language Models.
\newblock arXiv [Preprint] (2022).
\newblock \url{https://doi.org/10.48550/arXiv.2210.11416} (Accessed 2023-04-05).

\bibitem{googleGemmaOpenModels2024SI}
Google,  Gemma: {{Open Models Based}} on {{Gemini Research}} and {{Technology}}. (2024).
\newblock \url{https://storage.googleapis.com/deepmind-media/gemma/gemma-report.pdf} (Accessed 2024-02-03).

\bibitem{googleGemini15Unlocking2024SI}
Google,  Gemini 1.5: {{Unlocking}} Multimodal Understanding across Millions of Tokens of Context. (2024).
\newblock \url{https://storage.googleapis.com/deepmind-media/gemini/gemini_v1_5_report.pdf} (Accessed 2024-03-01).

\bibitem{brownLanguageModelsAre2020SI}
T. Brown, \emph{et~al.},  Language {{Models}} Are {{Few-Shot Learners}}.
\newblock {\em Advances in Neural Information Processing Systems} \textbf{33}: 1877--1901 (2020).

\bibitem{openaiGPT4TechnicalReport2023SI}
OpenAI, \emph{et~al.},  {{GPT-4}} Technical Report.
\newblock arXiv [Preprint] (2023).
\newblock \url{https://arxiv.org/abs/2303.08774} (Accessed 2024-03-20).

\bibitem{openaiGPT4oSystemCard2024SI}
OpenAI,  {{GPT-4o System Card}}. (2024).
\newblock \url{https://cdn.openai.com/gpt-4o-system-card.pdf} (Accessed 2024-08-01).

\bibitem{openaiGPT5SystemCard2025SI}
OpenAI,  {{GPT-5 System Card}}. (2025).
\newblock \url{https://cdn.openai.com/pdf/8124a3ce-ab78-4f06-96eb-49ea29ffb52f/gpt5-system-card-aug7.pdf} (Accessed 2025-08-19).

\bibitem{touvronLlama2Open2023SI}
H. Touvron, \emph{et~al.},  Llama 2: {{Open}} Foundation and Fine-Tuned Chat Models.
\newblock arXiv [Preprint] (2023).
\newblock \url{https://doi.org/10.48550/arXiv.2307.09288} (Accessed 2023-09-12).

\bibitem{grattafioriLlama3Herd2024SI}
A. Grattafiori, \emph{et~al.},  The {{Llama}} 3 {{Herd}} of {{Models}}.
\newblock arXiv [Preprint] (2024).
\newblock \url{https://arxiv.org/abs/2407.21783} (Accessed 2025-01-06).

\bibitem{jiangMistral7B2023SI}
A.~Q. Jiang, \emph{et~al.},  Mistral {{7B}}.
\newblock arXiv [Preprint] (2023).
\newblock \url{https://arxiv.org/abs/2310.06825} (Accessed 2023-11-13).

\bibitem{jiangMixtralExperts2024SI}
A.~Q. Jiang, \emph{et~al.},  Mixtral of Experts.
\newblock arXiv [Preprint] (2024).
\newblock \url{https://doi.org/10.48550/arXiv.2401.04088} (Accessed 2024-06-07).

\bibitem{mistralaiCheaperBetterFaster2024SI}
MistralAI,  Cheaper, {{Better}}, {{Faster}}, {{Stronger}}. (2024).
\newblock \url{https://mistral.ai/news/mixtral-8x22b/} (Accessed 2024-06-15).

\bibitem{mistralaiAuLarge2024SI}
MistralAI,  Au {{Large}}. (2024).
\newblock \url{https://mistral.ai/news/mistral-large/} (Accessed 2024-05-28).

\bibitem{groeneveldOLMoAcceleratingScience2024SI}
D. Groeneveld, \emph{et~al.},  ``{{OLMo}}: {{Accelerating}} the {{Science}} of {{Language Models}}'' in {\em Proceedings of the 62nd {{Annual Meeting}} of the {{Association}} for {{Computational Linguistics}} ({{Volume}} 1: {{Long Papers}})} (2024), pp. 15789--15809.

\bibitem{yangQwen2TechnicalReport2024SI}
A. Yang, \emph{et~al.},  Qwen2 Technical Report.
\newblock arXiv [Preprint] (2024).
\newblock \url{https://doi.org/10.48550/arXiv.2407.10671} (Accessed 2024-12-11).

\bibitem{chiangChatbotArenaOpen2024SI}
W.-L. Chiang, \emph{et~al.},  ``Chatbot {{Arena}}: {{An Open Platform}} for {{Evaluating LLMs}} by {{Human Preference}}'' in {\em Proceedings of the 41st {{International Conference}} on {{Machine Learning}}} (2024).

\bibitem{mihaylovCanSuitArmor2018SI}
T. Mihaylov, P. Clark, T. Khot, A. Sabharwal,  ``Can a {{Suit}} of {{Armor Conduct Electricity}}? {{A New Dataset}} for {{Open Book Question Answering}}'' in {\em Proceedings of the 2018 {{Conference}} on {{Empirical Methods}} in {{Natural Language Processing}}} (2018), pp. 2381--2391.

\bibitem{liangHolisticEvaluationLanguage2023SI}
P. Liang, \emph{et~al.},  Holistic Evaluation of Language Models.
\newblock {\em Transactions on Machine Learning Research} (2023).

\end{thebibliography}


\begin{thebibliography}{100}

\bibitem{gordonFormalTheoryCommonsense2017}
A.~S. Gordon, J.~R. Hobbs,  {\em A Formal Theory of Commonsense Psychology: {{How}} People Think People Think} (Cambridge University Press, 2017).

\bibitem{mccarthyProgramsCommonSense1959}
J. McCarthy,  ``Programs with Common Sense'' in {\em Mechanisation of {{Thought Processes}}} (1959), vol.{}~1, pp. 75--91.

\bibitem{minskyEmotionMachineCommensense2007}
M. Minsky,  {\em The Emotion Machine: Commensense Thinking, Artificial Intelligence, and the Future of the Human Mind} (Simon \& Schuster, 2007).

\bibitem{davisCommonsenseReasoningCommonsense2015}
E. Davis, G. Marcus,  Commonsense Reasoning and Commonsense Knowledge in Artificial Intelligence.
\newblock {\em Communications of the ACM} \textbf{58}(9): 92--103 (2015).

\bibitem{brachmanMachinesUsAI2023}
R.~J. Brachman, H.~J. Levesque,  {\em Machines like Us: Toward {{AI}} with Common Sense} (MIT Press, 2023).

\bibitem{kejriwalCanAIHave2024}
M. Kejriwal, \emph{et~al.},  Can {{AI}} Have Common Sense? {{Finding}} out Will Be Key to Achieving Machine Intelligence.
\newblock {\em Nature} \textbf{634}(8033): 291--294 (2024).

\bibitem{petroniLanguageModelsKnowledge2019}
F. Petroni, \emph{et~al.},  ``Language {{Models}} as {{Knowledge Bases}}?'' in {\em Proceedings of the 2019 {{Conference}} on {{Empirical Methods}} in {{Natural Language Processing}} and the 9th {{International Joint Conference}} on {{Natural Language Processing}} ({{EMNLP-IJCNLP}})} (2019), pp. 2463--2473.

\bibitem{davisonCommonsenseKnowledgeMining2019}
J. Davison, J. Feldman, A. Rush,  ``Commonsense {{Knowledge Mining}} from {{Pretrained Models}}'' in {\em Proceedings of the 2019 {{Conference}} on {{Empirical Methods}} in {{Natural Language Processing}} and the 9th {{International Joint Conference}} on {{Natural Language Processing}} ({{EMNLP-IJCNLP}})} (2019), pp. 1173--1178.

\bibitem{jiangHowCanWe2020}
Z. Jiang, F.~F. Xu, J. Araki, G. Neubig,  How {{Can We Know What Language Models Know}}?
\newblock {\em Transactions of the Association for Computational Linguistics} \textbf{8}: 423--438 (2020).

\bibitem{jiangHowCanWe2021}
Z. Jiang, J. Araki, H. Ding, G. Neubig,  How {{Can We Know}} {{{\emph{When}}}} {{Language Models Know}}? {{On}} the {{Calibration}} of {{Language Models}} for {{Question Answering}}.
\newblock {\em Transactions of the Association for Computational Linguistics} \textbf{9}: 962--977 (2021).

\bibitem{liSystematicInvestigationCommonsense2022}
X.~L. Li, \emph{et~al.},  ``A {{Systematic Investigation}} of {{Commonsense Knowledge}} in {{Large Language Models}}'' in {\em Proceedings of the 2022 {{Conference}} on {{Empirical Methods}} in {{Natural Language Processing}}} (2022), pp. 11838--11855.

\bibitem{choiCuriousCaseCommonsense2022}
Y. Choi,  The {{Curious Case}} of {{Commonsense Intelligence}}.
\newblock {\em Daedalus} \textbf{151}(2): 139--155 (2022).

\bibitem{bubeckSparksArtificialGeneral2023}
S. Bubeck, \emph{et~al.},  Sparks of {{Artificial General Intelligence}}: {{Early}} Experiments with {{GPT-4}}.
\newblock arXiv [Preprint] (2023).
\newblock \url{https://arxiv.org/abs/2303.12712} (Accessed 2024-10-10).

\bibitem{binzUsingCognitivePsychology2023}
M. Binz, E. Schulz,  Using Cognitive Psychology to Understand {{GPT-3}}.
\newblock {\em Proceedings of the National Academy of Sciences} \textbf{120}(6): e2218523120 (2023).

\bibitem{meiTuringTestWhether2024}
Q. Mei, Y. Xie, W. Yuan, M.~O. Jackson,  A {{Turing}} Test of Whether {{AI}} Chatbots Are Behaviorally Similar to Humans.
\newblock {\em Proceedings of the National Academy of Sciences} \textbf{121}(9): e2313925121 (2024).

\bibitem{strachanTestingTheoryMind2024}
J.~W.~A. Strachan, \emph{et~al.},  Testing Theory of Mind in Large Language Models and Humans.
\newblock {\em Nature Human Behaviour} \textbf{8}(7): 1285--1295 (2024).

\bibitem{kosinskiEvaluatingLargeLanguage2024}
M. Kosinski,  Evaluating Large Language Models in Theory of Mind Tasks.
\newblock {\em Proceedings of the National Academy of Sciences} \textbf{121}(45): e2405460121 (2024).

\bibitem{davisBenchmarksAutomatedCommonsense2024}
E. Davis,  Benchmarks for {{Automated Commonsense Reasoning}}: {{A Survey}}.
\newblock {\em ACM Computing Surveys} \textbf{56}(4): 1--41 (2024).

\bibitem{schutzCommonSenseScientificInterpretation1953}
A. Schutz,  Common-{{Sense}} and {{Scientific Interpretation}} of {{Human Action}}.
\newblock {\em Philosophy and Phenomenological Research} \textbf{14}(1): 1--37 (1953).

\bibitem{bergerSocialConstructionReality1966}
P.~L. Berger, T. Luckmann,  {\em The {{Social Construction}} of {{Reality}}: {{A Treatise}} in the {{Sociology}} of {{Knowledge}}} (Anchor Books, 1966).

\bibitem{mertonSocialTheorySocial1968}
R.~K. Merton,  {\em Social {{Theory}} and {{Social Structure}}} (Free Press, 1968).

\bibitem{fletcherPsychologyCommonSense1984}
G.~J. Fletcher,  Psychology and Common Sense.
\newblock {\em American Psychologist} \textbf{39}(3): 203--213 (1984).

\bibitem{rosenfeldCommonSensePolitical2011}
S.~A. Rosenfeld,  {\em Common Sense: A Political History} (Harvard University Press, 2011).

\bibitem{whitingFrameworkQuantifyingIndividual2024}
M.~E. Whiting, D.~J. Watts,  A Framework for Quantifying Individual and Collective Common Sense.
\newblock {\em Proceedings of the National Academy of Sciences} \textbf{121}(4): e2309535121 (2024).

\bibitem{yingBenchmarkingHumanLikeIntelligence2025}
L. Ying, \emph{et~al.},  On {{Benchmarking Human-Like Intelligence}} in {{Machines}}.
\newblock arXiv [Preprint] (2025).
\newblock \url{https://arxiv.org/abs/2502.20502} (Accessed 2025-03-12).

\bibitem{geertzCommonSenseCultural1975}
C. Geertz,  Common Sense as a Cultural System.
\newblock {\em The Antioch Review} \textbf{33}(1): 5--26 (1975).

\bibitem{andreasLanguageModelsAgent2022}
J. Andreas,  ``Language {{Models}} as {{Agent Models}}'' in {\em Findings of the {{Association}} for {{Computational Linguistics}}: {{EMNLP}}} (2022), pp. 5769--5779.

\bibitem{shanahanRolePlayLarge2023}
M. Shanahan, K. McDonell, L. Reynolds,  Role Play with Large Language Models.
\newblock {\em Nature} \textbf{623}(7987): 493--498 (2023).

\bibitem{farrellLargeAIModels2025}
H. Farrell, A. Gopnik, C. Shalizi, J. Evans,  Large {{AI}} Models Are Cultural and Social Technologies.
\newblock {\em Science} \textbf{387}(6739): 1153--1156 (2025).

\bibitem{jordanCollectivistEconomicPerspective2025}
M.~I. Jordan,  A {{Collectivist}}, {{Economic Perspective}} on {{AI}}.
\newblock arXiv [Preprint] (2025).
\newblock \url{http://arxiv.org/abs/2507.06268} (Accessed 2025-07-14).

\bibitem{argyleOutOneMany2023}
L.~P. Argyle, \emph{et~al.},  Out of {{One}}, {{Many}}: {{Using Language Models}} to {{Simulate Human Samples}}.
\newblock {\em Political Analysis} \textbf{31}(3): 337--351 (2023).

\bibitem{santurkarWhoseOpinionsLanguage2023}
S. Santurkar, \emph{et~al.},  ``Whose Opinions Do Language Models Reflect?'' in {\em Proceedings of the 40th {{International Conference}} on {{Machine Learning}}} (2023), vol.{} 202, pp. 29971--30004.

\bibitem{dillionCanAILanguage2023}
D. Dillion, N. Tandon, Y. Gu, K. Gray,  Can {{AI}} Language Models Replace Human Participants?
\newblock {\em Trends in Cognitive Sciences} \textbf{27}(7): 597--600 (2023).

\bibitem{hamalainenEvaluatingLargeLanguage2023}
P. H{\"a}m{\"a}l{\"a}inen, M. Tavast, A. Kunnari,  ``Evaluating {{Large Language Models}} in {{Generating Synthetic HCI Research Data}}: A {{Case Study}}'' in {\em Proceedings of the 2023 {{CHI Conference}} on {{Human Factors}} in {{Computing Systems}}} (2023), pp. 1--19.

\bibitem{jansenEmployingLargeLanguage2023}
B.~J. Jansen, S.-g. Jung, J. Salminen,  Employing Large Language Models in Survey Research.
\newblock {\em Natural Language Processing Journal} \textbf{4}: 100020 (2023).

\bibitem{grossmannAITransformationSocial2023}
I. Grossmann, \emph{et~al.},  {{AI}} and the Transformation of Social Science Research.
\newblock {\em Science} \textbf{380}(6650): 1108--1109 (2023).

\bibitem{durmusMeasuringRepresentationSubjective2024}
E. Durmus, \emph{et~al.},  ``Towards Measuring the Representation of Subjective Global Opinions in Language Models'' in {\em Conference on {{Language Modeling}}} (2024).

\bibitem{kejriwalNoiseAuditHumanlabeled2024}
M. Kejriwal, H. Santos, K. Shen, A.~M. Mulvehill, D.~L. McGuinness,  A Noise Audit of Human-Labeled Benchmarks for Machine Commonsense Reasoning.
\newblock {\em Scientific Reports} \textbf{14}(1): 8609 (2024).

\bibitem{paltaPlausiblyProblematicQuestions2024}
S. Palta, \emph{et~al.},  ``Plausibly {{Problematic Questions}} in {{Multiple-Choice Benchmarks}} for {{Commonsense Reasoning}}'' in {\em Findings of the {{Association}} for {{Computational Linguistics}}: {{EMNLP}} 2024} (2024), pp. 3451--3473.

\bibitem{pawarSurveyCulturalAwareness2024}
S. Pawar, \emph{et~al.},  Survey of {{Cultural Awareness}} in {{Language Models}}: {{Text}} and {{Beyond}}.
\newblock arXiv [Preprint] (2024).
\newblock \url{http://arxiv.org/abs/2411.00860} (Accessed 2025-04-23).

\bibitem{openaiGPT4TechnicalReport2023}
OpenAI, \emph{et~al.},  {{GPT-4}} Technical Report.
\newblock arXiv [Preprint] (2023).
\newblock \url{https://arxiv.org/abs/2303.08774} (Accessed 2024-03-20).

\bibitem{lindenbergCommonSenseSocial1987}
S. Lindenberg,  Common {{Sense}} and {{Social Structure}}: {{A Sociological View}} in {\em Common {{Sense}}: {{The Foundations}} for {{Social Science}}}, Sources in Semiotics, eds.{} F. van Holthoon, D.~R. Olson.
\newblock vol.{}~6, pp. 199--215 (University Press of America, 1987).

\bibitem{romneyCultureConsensusTheory1986}
A.~K. Romney, S.~C. Weller, W.~H. Batchelder,  Culture as {{Consensus}}: {{A Theory}} of {{Culture}} and {{Informant Accuracy}}.
\newblock {\em American Anthropologist} \textbf{88}(2): 313--338 (1986).

\bibitem{dresslerCultureIndividualTheory2018}
W.~W. Dressler,  {\em Culture and the {{Individual}}: {{Theory}} and {{Method}} of {{Cultural Consonance}}} (Routledge, 2018).

\bibitem{kaplanScalingLawsNeural2020}
J. Kaplan, \emph{et~al.},  Scaling {{Laws}} for {{Neural Language Models}}.
\newblock arXiv [Preprint] (2020).
\newblock \url{https://arxiv.org/abs/2001.08361} (Accessed 2024-10-23).

\bibitem{chiangChatbotArenaOpen2024}
W.-L. Chiang, \emph{et~al.},  ``Chatbot {{Arena}}: {{An Open Platform}} for {{Evaluating LLMs}} by {{Human Preference}}'' in {\em Proceedings of the 41st {{International Conference}} on {{Machine Learning}}} (2024).

\bibitem{mihaylovCanSuitArmor2018}
T. Mihaylov, P. Clark, T. Khot, A. Sabharwal,  ``Can a {{Suit}} of {{Armor Conduct Electricity}}? {{A New Dataset}} for {{Open Book Question Answering}}'' in {\em Proceedings of the 2018 {{Conference}} on {{Empirical Methods}} in {{Natural Language Processing}}} (2018), pp. 2381--2391.

\bibitem{mckenzieInverseScalingWhen2023}
I.~R. McKenzie, \emph{et~al.},  Inverse Scaling: {{When}} Bigger Isn't Better.
\newblock {\em Transactions on Machine Learning Research} (2023).

\bibitem{linTruthfulQAMeasuringHow2022}
S. Lin, J. Hilton, O. Evans,  ``{{TruthfulQA}}: {{Measuring How Models Mimic Human Falsehoods}}'' in {\em Proceedings of the 60th {{Annual Meeting}} of the {{Association}} for {{Computational Linguistics}} ({{Volume}} 1: {{Long Papers}})} (2022), pp. 3214--3252.

\bibitem{kimAIAugmentedSurveysLeveraging2023}
J. Kim, B. Lee,  {{AI-Augmented Surveys}}: {{Leveraging Large Language Models}} and {{Surveys}} for {{Opinion Prediction}}.
\newblock arXiv [Preprint] (2023).
\newblock \url{https://arxiv.org/abs/2305.09620} (Accessed 2024-10-10).

\bibitem{kozlowskiSilicoSociologyForecasting2024}
A.~C. Kozlowski, H. Kwon, J.~A. Evans,  In {{Silico Sociology}}: {{Forecasting COVID-19 Polarization}} with {{Large Language Models}}.
\newblock arXiv [Preprint] (2024).
\newblock \url{https://arxiv.org/abs/2407.11190} (Accessed 2025-04-06).

\bibitem{asheryEmergentSocialConventions2025}
A.~F. Ashery, L.~M. Aiello, A. Baronchelli,  Emergent Social Conventions and Collective Bias in {{LLM}} Populations.
\newblock {\em Science Advances} \textbf{11}(20): eadu9368 (2025).

\bibitem{hortonLargeLanguageModels2023}
J. Horton,  \emph{Large {{Language Models}} as {{Simulated Economic Agents}}: {{What Can We Learn}} from {{Homo Silicus}}?}, Working {{Paper}} 31122, National Bureau of Economic Research (2023).

\bibitem{brandUsingLLMsMarket2023}
J. Brand, A. Israeli, D. Ngwe,  \emph{Using {{LLMs}} for {{Market Research}}}, Working {{Paper}} 23-062, Harvard Business School (2023).

\bibitem{burtonHowLargeLanguage2024}
J.~W. Burton, \emph{et~al.},  How Large Language Models Can Reshape Collective Intelligence.
\newblock {\em Nature Human Behaviour} \textbf{8}(9): 1643--1655 (2024).

\bibitem{heymanImpactChatGPTHuman2023}
T. Heyman, G. Heyman,  The Impact of {{ChatGPT}} on Human Data Collection: {{A}} Case Study Involving Typicality Norming Data.
\newblock {\em Behavior Research Methods} \textbf{56}(5): 4974--4981 (2023).

\bibitem{bailCanGenerativeAI2024}
C.~A. Bail,  Can {{Generative AI}} Improve Social Science?
\newblock {\em Proceedings of the National Academy of Sciences} \textbf{121}(21): e2314021121 (2024).

\bibitem{kozlowskiSimulatingSubjectsPromise2025}
A.~C. Kozlowski, J. Evans,  Simulating {{Subjects}}: {{The Promise}} and {{Peril}} of {{Artificial Intelligence Stand-Ins}} for {{Social Agents}} and {{Interactions}}.
\newblock {\em Sociological Methods \& Research} \textbf{54}(3): 1017--1073 (2025).

\bibitem{baiTrainingHelpfulHarmless2022}
Y. Bai, \emph{et~al.},  Training a {{Helpful}} and {{Harmless Assistant}} with {{Reinforcement Learning}} from {{Human Feedback}}.
\newblock arXiv [Preprint] (2022).
\newblock \url{https://arxiv.org/abs/2204.05862} (Accessed 2024-03-01).

\bibitem{kadavathLanguageModelsMostly2022}
S. Kadavath, \emph{et~al.},  Language Models (Mostly) Know What They Know.
\newblock arXiv [Preprint] (2022).
\newblock \url{https://doi.org/10.48550/arXiv.2207.05221} (Accessed 2023-08-01).

\bibitem{rahwanMachineBehaviour2019}
I. Rahwan, \emph{et~al.},  Machine Behaviour.
\newblock {\em Nature} \textbf{568}(7753): 477--486 (2019).

\bibitem{collinsBuildingMachinesThat2024}
K.~M. Collins, \emph{et~al.},  Building Machines That Learn and Think with People.
\newblock {\em Nature Human Behaviour} \textbf{8}(10): 1851--1863 (2024).

\bibitem{tsvetkovaNewSociologyHumans2024}
M. Tsvetkova, T. Yasseri, N. Pescetelli, T. Werner,  A New Sociology of Humans and Machines.
\newblock {\em Nature Human Behaviour} \textbf{8}(10): 1864--1876 (2024).

\bibitem{gweonSociallyIntelligentMachines2023}
H. Gweon, J. Fan, B. Kim,  Socially Intelligent Machines That Learn from Humans and Help Humans Learn.
\newblock {\em Philosophical Transactions of the Royal Society A: Mathematical, Physical and Engineering Sciences} \textbf{381}(2251): 20220048 (2023).

\bibitem{levesqueWinogradSchemaChallenge2012}
H. Levesque, E. Davis, L. Morgenstern,  ``The {{Winograd Schema Challenge}}'' in {\em Proceedings of the {{Thirteenth International Conference}} on {{Principles}} of {{Knowledge Representation}} and {{Reasoning}}} (2012), pp. 552--561.

\bibitem{talmorCommonsenseQAQuestionAnswering2019}
A. Talmor, J. Herzig, N. Lourie, J. Berant,  ``{{CommonsenseQA}}: A Question Answering Challenge Targeting Commonsense Knowledge'' in {\em Proceedings of the 2019 Conference of the North {{American}} Chapter of the Association for Computational Linguistics: {{Human}} Language Technologies, Volume 1 (Long and Short Papers)} (2019), pp. 4149--4158.

\bibitem{zellersHellaSwagCanMachine2019}
R. Zellers, A. Holtzman, Y. Bisk, A. Farhadi, Y. Choi,  ``{{HellaSwag}}: {{Can}} a Machine Really Finish Your Sentence?'' in {\em Proceedings of the 57th Annual Meeting of the Association for Computational Linguistics} (2019), pp. 4791--4800.

\bibitem{nguyenExtractingCulturalCommonsense2023}
T.-P. Nguyen, S. Razniewski, A. Varde, G. Weikum,  ``Extracting {{Cultural Commonsense Knowledge}} at {{Scale}}'' in {\em Proceedings of the {{ACM Web Conference}} 2023} (2023), pp. 1907--1917.

\bibitem{huangCulturallyAwareNatural2023}
J. Huang, D. Yang,  ``Culturally {{Aware Natural Language Inference}}'' in {\em Findings of the {{Association}} for {{Computational Linguistics}}: {{EMNLP}} 2023} (2023), pp. 7591--7609.

\bibitem{aroyoTruthLieCrowd2015}
L. Aroyo, C. Welty,  Truth {{Is}} a {{Lie}}: {{Crowd Truth}} and the {{Seven Myths}} of {{Human Annotation}}.
\newblock {\em AI Magazine} \textbf{36}(1): 15--24 (2015).

\bibitem{almSubjectiveNaturalLanguage2011}
C.~O. Alm,  ``Subjective Natural Language Problems: {{Motivations}}, Applications, Characterizations, and Implications'' in {\em Proceedings of the 49th Annual Meeting of the Association for Computational Linguistics: {{Human}} Language Technologies} (2011), pp. 107--112.

\bibitem{goyalYourToxicityMy2022}
N. Goyal, I.~D. Kivlichan, R. Rosen, L. Vasserman,  Is {{Your Toxicity My Toxicity}}? {{Exploring}} the {{Impact}} of {{Rater Identity}} on {{Toxicity Annotation}}.
\newblock {\em Proceedings of the ACM on Human-Computer Interaction} \textbf{6}(CSCW2): 1--28 (2022).

\bibitem{miceliSubjectivityImpositionPower2020}
M. Miceli, M. Schuessler, T. Yang,  Between {{Subjectivity}} and {{Imposition}}: {{Power Dynamics}} in {{Data Annotation}} for {{Computer Vision}}.
\newblock {\em Proceedings of the ACM on Human-Computer Interaction} \textbf{4}(CSCW2): 1--25 (2020).

\bibitem{fraserDoesMoralCode2022}
K.~C. Fraser, S. Kiritchenko, E. Balkir,  ``Does {{Moral Code}} Have a {{Moral Code}}? {{Probing Delphi}}'s {{Moral Philosophy}}'' in {\em Proceedings of the 2nd {{Workshop}} on {{Trustworthy Natural Language Processing}} ({{TrustNLP}} 2022)} (2022), pp. 26--42.

\bibitem{sorensenValueKaleidoscopeEngaging2024}
T. Sorensen, \emph{et~al.},  Value {{Kaleidoscope}}: {{Engaging AI}} with {{Pluralistic Human Values}}, {{Rights}}, and {{Duties}}.
\newblock {\em Proceedings of the AAAI Conference on Artificial Intelligence} \textbf{38}(18): 19937--19947 (2024).

\bibitem{jiangInvestigatingMachineMoral2025}
L. Jiang, \emph{et~al.},  Investigating Machine Moral Judgement through the {{Delphi}} Experiment.
\newblock {\em Nature Machine Intelligence} \textbf{7}(1): 145--160 (2025).

\bibitem{mendelsohnDogwhistlesBullhornsUnveiling2023}
J. Mendelsohn, R. Le~Bras, Y. Choi, M. Sap,  ``From {{Dogwhistles}} to {{Bullhorns}}: {{Unveiling Coded Rhetoric}} with {{Language Models}}'' in {\em Proceedings of the 61st {{Annual Meeting}} of the {{Association}} for {{Computational Linguistics}} ({{Volume}} 1: {{Long Papers}})} (2023), pp. 15162--15180.

\bibitem{pelletierWhatShouldDefault1997}
F.~J. Pelletier, R. Elio,  What {{Should Default Reasoning}} Be, by {{Default}}?
\newblock {\em Computational Intelligence} \textbf{13}(2): 165--187 (1997).

\bibitem{davaniDealingDisagreementsLooking2022}
A.~M. Davani, M. D{\'\i}az, V. Prabhakaran,  Dealing with Disagreements: {{Looking}} beyond the Majority Vote in Subjective Annotations.
\newblock {\em Transactions of the Association for Computational Linguistics} \textbf{10}: 92--110 (2022).

\bibitem{plankProblemHumanLabel2022}
B. Plank,  ``The ``{{Problem}}'' of {{Human Label Variation}}: {{On Ground Truth}} in {{Data}}, {{Modeling}} and {{Evaluation}}'' in {\em Proceedings of the 2022 {{Conference}} on {{Empirical Methods}} in {{Natural Language Processing}}} (2022), pp. 10671--10682.

\bibitem{cabitzaPerspectivistTurnGround2023}
F. Cabitza, A. Campagner, V. Basile,  Toward a {{Perspectivist Turn}} in {{Ground Truthing}} for {{Predictive Computing}}.
\newblock {\em Proceedings of the AAAI Conference on Artificial Intelligence} \textbf{37}(6): 6860--6868 (2023).

\bibitem{meisterBenchmarkingDistributionalAlignment2025}
N. Meister, C. Guestrin, T. Hashimoto,  ``Benchmarking {{Distributional Alignment}} of {{Large Language Models}}'' in {\em Proceedings of the 2025 {{Conference}} of the {{Nations}} of the {{Americas Chapter}} of the {{Association}} for {{Computational Linguistics}}: {{Human Language Technologies}} ({{Volume}} 1: {{Long Papers}})} (2025), pp. 24--49.

\bibitem{ilievskiDimensionsCommonsenseKnowledge2021}
F. Ilievski, \emph{et~al.},  Dimensions of Commonsense Knowledge.
\newblock {\em Knowledge-Based Systems} \textbf{229}: 107347 (2021).

\bibitem{parkGenerativeAgentSimulations2024}
J.~S. Park, \emph{et~al.},  Generative {{Agent Simulations}} of 1,000 {{People}}.
\newblock arXiv [Preprint] (2024).
\newblock \url{https://arxiv.org/abs/2411.10109} (Accessed 2024-12-10).

\bibitem{jiangPersonaLLMInvestigatingAbility2024}
H. Jiang, \emph{et~al.},  ``{{PersonaLLM}}: {{Investigating}} the {{Ability}} of {{Large Language Models}} to {{Express Personality Traits}}'' in {\em Findings of the {{Association}} for {{Computational Linguistics}}: {{NAACL}} 2024} (2024), pp. 3605--3627.

\bibitem{anthisLLMSocialSimulations2025}
J.~R. Anthis, \emph{et~al.},  ``{{LLM Social Simulations Are}} a {{Promising Research Method}}'' in {\em Proceedings of the 42nd {{International Conference}} on {{Machine Learning}}} (2025).

\bibitem{bisbeeSyntheticReplacementsHuman2024}
J. Bisbee, J.~D. Clinton, C. Dorff, B. Kenkel, J.~M. Larson,  Synthetic {{Replacements}} for {{Human Survey Data}}? {{The Perils}} of {{Large Language Models}}.
\newblock {\em Political Analysis} \textbf{32}(4): 401--416 (2024).

\bibitem{huQuantifyingPersonaEffect2024}
T. Hu, N. Collier,  ``Quantifying the {{Persona Effect}} in {{LLM Simulations}}'' in {\em Proceedings of the 62nd {{Annual Meeting}} of the {{Association}} for {{Computational Linguistics}} ({{Volume}} 1: {{Long Papers}})} (2024), pp. 10289--10307.

\bibitem{wangLargeLanguageModels2025}
A. Wang, J. Morgenstern, J.~P. Dickerson,  Large Language Models That Replace Human Participants Can Harmfully Misportray and Flatten Identity Groups.
\newblock {\em Nature Machine Intelligence} \textbf{7}: 400--411 (2025).

\bibitem{gaoTakeCautionUsing2025}
Y. Gao, D. Lee, G. Burtch, S. Fazelpour,  Take Caution in Using {{LLMs}} as Human Surrogates.
\newblock {\em Proceedings of the National Academy of Sciences} \textbf{122}(24): e2501660122 (2025).

\bibitem{lymanBalancingLargeLanguage2025}
A. Lyman, \emph{et~al.},  Balancing {{Large Language Model Alignment}} and {{Algorithmic Fidelity}} in {{Social Science Research}}.
\newblock {\em Sociological Methods \& Research} \textbf{54}(3): 1110--1155 (2025).

\bibitem{sperberExplainingCultureNaturalistic1996}
D. Sperber,  {\em Explaining {{Culture}}: A Naturalistic Approach} (Blackwell, 1996).

\bibitem{zerubavelSocialMindscapesInvitation1997}
E. Zerubavel,  {\em Social Mindscapes: An Invitation to Cognitive Sociology} (Harvard University Press, 1997).

\bibitem{wattsEverythingObviousOnce2011}
D.~J. Watts,  {\em Everything Is Obvious: Once You Know the Answer} (Crown Business, 2011).

\bibitem{jara-ettingerNaiveUtilityCalculus2016}
J. {Jara-Ettinger}, H. Gweon, L.~E. Schulz, J.~B. Tenenbaum,  The {{Na\"ive Utility Calculus}}: {{Computational Principles Underlying Commonsense Psychology}}.
\newblock {\em Trends in Cognitive Sciences} \textbf{20}(8): 589--604 (2016).

\bibitem{lakeBuildingMachinesThat2017}
B.~M. Lake, T.~D. Ullman, J.~B. Tenenbaum, S.~J. Gershman,  Building Machines That Learn and Think like People.
\newblock {\em Behavioral and Brain Sciences} \textbf{40}: e253 (2017).

\bibitem{mathurManipulativeTacticsAre2023}
A. Mathur, \emph{et~al.},  Manipulative Tactics Are the Norm in Political Emails: {{Evidence}} from {{300K}} Emails from the 2020 {{US}} Election Cycle.
\newblock {\em Big Data \& Society} \textbf{10}(1): 20539517221145371 (2023).

\bibitem{speerConceptNet55Open2017}
R. Speer, J. Chin, C. Havasi,  {{ConceptNet}} 5.5: {{An Open Multilingual Graph}} of {{General Knowledge}}.
\newblock {\em Proceedings of the AAAI Conference on Artificial Intelligence} \textbf{31}(1): 4444--4451 (2017).

\bibitem{hwangCOMETATOMIC2020Symbolic2021}
J.~D. Hwang, \emph{et~al.},  {{COMET-ATOMIC}} 2020: {{On}} Symbolic and Neural Commonsense Knowledge Graphs.
\newblock {\em Proceedings of the AAAI Conference on Artificial Intelligence} \textbf{35}(7): 6384--6392 (2021).

\bibitem{nguyenTestingCommonSense2024}
T.~D. Nguyen, D.~J. Watts, M.~E. Whiting,  Testing {{Common Sense}} in {{Large Language Models}}. (2024).
\newblock \url{https://aspredicted.org/XXL_ZPR} (Accessed 2024-02-16).

\bibitem{nguyenLargescaleEvaluationCommonsense2026}
T.~D. Nguyen, D.~J. Watts, M.~E. Whiting,  A Large-Scale Evaluation of Commonsense Knowledge in Humans and Large Language Models. (2026) Zenodo.
\newblock \url{https://doi.org/10.5281/zenodo.17281355} (Accessed 2026-01-20).

\end{thebibliography}

\normalsize
\normalfont

\newpage
\makeatletter
\setcounter{NAT@ctr}{0}
\makeatother
\renewcommand{\appendixpagename}{\LARGE Supplementary Information for \newline \papertitle}
\appendix
\phantomsection
\label{appendixStart}
\appendixpage
\justifying
\counterwithin{figure}{section}
\counterwithin{table}{section}
\counterwithin{lstlisting}{section}

\setcounter{figure}{0}
\setcounter{table}{0}
\setcounter{equation}{0}
\setcounter{lstlisting}{0}

\vspace{15pt}
{\noindent \Large Tuan Dung Nguyen\textsuperscript{\cis,\correspondingA}, Duncan J. Watts\textsuperscript{\cis,\oid,\asc}, and Mark E. Whiting\textsuperscript{\cis,\oid,\pareto}}

\vspace{20pt}

\noindent\textsuperscript{\cis} Department of Computer and Information Science, University of Pennsylvania, Philadelphia, PA 19104, USA.

\noindent\textsuperscript{\oid} Operations, Information and Decisions Department, The Wharton School, University of Pennsylvania, Philadelphia, PA 19104, USA.

\noindent\textsuperscript{\asc} Annenberg School for Communication, University of Pennsylvania, Philadelphia, PA 19104, USA.

\noindent\textsuperscript{\pareto} Pareto, San Francisco, CA 94111, USA.

\noindent \textsuperscript{\correspondingA} Corresponding Author: Tuan Dung Nguyen. Email: joshtn@seas.upenn.edu.

\vspace{-25pt}

\part{}
\noptcrule  
\parttoc

\newpage

\section{Measuring Common Sense in Humans and Statements}
\label{si:sec:commonsense_recap}

Here we describe the commonsensicality framework by Whiting and Watts~\citepSI{whitingFrameworkQuantifyingIndividual2024SI}, whose data we adopt to study common sense in LLMs. 

From September 2 to September 14, 2022, a total of 2,141 individuals were recruited on Amazon Mechanical Turk to perform statement rating. All but 10 participants were from the United States. The exceptions were 3 individuals from Canada, and one individual from each of the following countries: Bangladesh, Romania, France, Mexico, Italy, Germany and Australia.

Only individuals who self-declared---via explicit answers throughout the session---to give honest, accurate and non-random answers were included. In addition, only individuals who participated in both the Reading the Mind in the Eyes Test~\citepSI{baron-cohenReadingMindEyes2001SI} and the Cognitive Reflection Test \citepSI{bialekCognitiveReflectionTest2018SI} were included. This gives a final sample of 2,046 individuals. Some descriptive statistics of this sample are given in \Cref{si:tab:human_descriptives}. More information about the survey setup can also be found in the Supporting Information of ref.~\citenum{whitingFrameworkQuantifyingIndividual2024SI}.

Suppose we have a set of $n_s \geq 1$ statements (see \Cref{si:tab:statement_sources} for an example) and a set of $n_p \geq 1$ people. (In this case, $n_s = 4,407$ and $n_p = 2,046$.) Given a statement, such as ``If you want to play a \emph{[sic]} guitar then you should take lessons,'' a human participant was asked two questions:
\begin{enumerate}
    \item[(a)] ``Do you agree with this statement?''
    \item[(b)] ``Do you think most people would agree with this statement?''
\end{enumerate}

Common sense is measured for both humans and statements. The underlying rationale is that a person with common sense knows what is true (in the sense that their belief coincides with the general human consensus), and also knows what other people believe to be true (in the sense that they can accurately predict what the majority of other people believe). Similarly, a statement is common sense if it is (nearly) unanimously agreed upon by people in this population, who must also accurately predict what the human consensus on this statement is. All of this depends on participants' answers to the two questions above.

To make this precise, define $A$ be the $n_s \times n_p$ matrix such that 
\begin{align}
    A_{i, j} = \begin{cases}
      1 & \text{if the $j$-th participant answered ``yes'' to question (a) about the $i$-th claim} \\
      0 & \text{if the $j$-th participant answered ``no'' to question (a) about the $i$-th claim}.
    \end{cases}
    \label{si:eq:human_ratings_q1}
\end{align}
In other words, the matrix $A$ contains participants' answers to question (a) above. Similarly, let $B$ be the $n_s \times n_p$ matrix containing answers to question (b) such that
\begin{align}
    B_{i, j} = \begin{cases}
      1 & \text{if the $j$-th participant answered ``yes'' to question (b) about the $i$-th claim} \\
      0 & \text{if the $j$-th participant answered ``no'' to question (b) about the $i$-th claim}.
    \end{cases}
    \label{si:eq:human_ratings_q2}
\end{align}

One complication in this study is that not every claim was labeled by everyone, because every participant was required to label only a small number of claims, 50 to be exact. Every $(i, j)$ entry in matrices $A$ and $B$, therefore, is only valid if the $i$-th claim was indeed assigned to the $i$-th participant; otherwise, participant $j$ would have never seen and labeled claim $i$ in the first place, and hence the entry must not be taken into account. Hence, $A$ and $B$ are partially filled matrices.

Define $\Omega_i \subseteq \{ 1, 2, \ldots, n_p \}$ as the set of participants who were indeed assigned to rate the $i$-th statement, and $\Phi_j \subseteq \{ 1, 2, \ldots, n_s \}$ as the set of statements that were indeed assigned to the $j$-th participant. We hence have $| \Phi_j | = 50$, where $| \cdot |$ denotes the size of a finite set.

Before we go on to define commonsensicality, define the \emph{human rating distribution} for each statement $i$ as the proportion of people that indicated they agreed with statement $i$ (\ie, by answering ``yes'' to question (a)):
\begin{align}
    d_{i}^{h, a} = \frac{1}{|\Omega_i|} \sum_{j \in \Omega_i} A_{i, j}.
    \label{si:eq:human_ratings_dist}
\end{align}

\begin{table}[p!]
\small
\centering
\begin{tabular}{llr}
\toprule
\textbf{Type}                    & \textbf{Level}               & \textbf{Share of level (\%)}   \\ \midrule
\multirow{4}{*}{\shortstack[l]{Age\\\hspace{1em}$M$ = 44.25, $SD$ = 12.87\\\hspace{1em}Median = 42.00\\\hspace{1em}Min = 21, Max = 85}}             & 21--39                          & 42.52            \\
                                 & 40--59                          & 42.03            \\
                                 & 60--79                          & 15.34            \\ 
                                 & 80+                             & 0.10             \\\midrule
\multirow{3}{*}{Gender}          & Female                          & 51.22           \\
                                 & Male                            & 48.29           \\
                                 & Other                           & 0.49           \\ \midrule
\multirow{5}{*}{Race}            & White                           & 80.50           \\
                                 & Black                           & 7.87           \\
                                 & Asian                           & 8.21           \\
                                 & Indigenous                      & 1.52           \\
                                 & Other                           & 1.91           \\ \midrule
\multirow{8}{*}{Education}       & Less than high school           & 0.59           \\
                                 & High school diploma             & 9.58          \\
                                 & Some college or vocational degree  & 17.64           \\
                                 & 2-year college degree           & 10.22           \\
                                 & 4-year college degree           & 43.84           \\
                                 & Post-college degree             & 13.83           \\
                                 & Professional degree (e.g, J.D., M.D.)    & 2.44           \\
                                 & Doctoral degree                 & 1.86           \\ \midrule
\multirow{3}{*}{Political party} & Democrat                        & 50.24           \\
                                 & Republican                      & 26.00           \\
                                 & Independent                     & 23.75           \\ \midrule
\multirow{5}{*}{Politics: social views}    & Strongly liberal                & 25.51           \\
                                 & Somewhat liberal                & 32.01           \\
                                 & Moderate                        & 16.13           \\
                                 & Somewhat conservatie            & 16.28           \\
                                 & Strongly conservative           & 10.07           \\ \midrule
\multirow{5}{*}{Politics: fiscal views}    & Strongly liberal                & 17.89           \\
                                 & Somewhat liberal                & 28.40           \\
                                 & Moderate                        & 18.91           \\
                                 & Somewhat conservatie            & 23.07           \\
                                 & Strongly conservative           & 11.73           \\ \midrule
\multirow{5}{*}{Marital status}  & Single never married            & 36.61           \\
                                 & Married or domestic partnership & 51.76           \\
                                 & Separated                       & 0.83           \\
                                 & Divorced                        & 9.19           \\
                                 & Widowed                         & 1.61           \\ \midrule
\multirow{12}{*}{Income}   & Less than \$10,000              & 3.52           \\
                                 & \$10,000 to \$19,999            & 6.45           \\
                                 & \$20,000 to \$29,999            & 9.87           \\
                                 & \$30,000 to \$39,999            & 10.46           \\
                                 & \$40,000 to \$49,999            & 10.95          \\
                                 & \$50,000 to \$59,999            & 11.97           \\
                                 & \$60,000 to \$69,999            & 7.04           \\
                                 & \$70,000 to \$79,999            & 8.60           \\
                                 & \$80,000 to \$89,999            & 5.18           \\
                                 & \$90,000 to \$99,999            & 6.40           \\ 
                                 & \$100,000 to \$149,999          & 13.10           \\
                                 & \$150,000 or more               & 6.45           \\ \bottomrule
\end{tabular}
\caption{Descriptive statistics of the human sample (\textit{N} = 2,046).}
\label{si:tab:human_descriptives}
\end{table}

\begin{table}[p!]
    \small
    \begin{tabular}{p{0.1\linewidth}p{0.04\linewidth}p{0.77\linewidth}}
    \toprule
    \textbf{Data source} & \textbf{\textit{N}} & \textbf{Examples} \\
    \midrule 
    News \newline media       & 290               & 
    \begin{tabular}[c]{@{}p{1\linewidth}} 
        - A low covid positivity rate should allow for restrictions to be loosened. \\ 
        - Children should not be taught to be ashamed of their own skin color. \\
        - Providing access to loans for producers boosts agricultural economy during crisis. \\
        - We should aim to put our responsibilities before any personal or political agenda. \\
        - The proliferation of charter schools affects the ability of states to efficiently distribute tax dollars.
    \end{tabular} \\
    \midrule
    Campaign emails       & 668               & 
    \begin{tabular}[c]{@{}p{1\linewidth}} 
        - Abolish the police is an extreme stance taken by democrats. \\
        - Access to healthcare should be available for all without worry of debt. \\
        - Partisan politics and the democrats have held up real change. \\
        - The health care system can help a lot of americans that need jobs. \\
        - We should support freedom transparency and accountability in our elections.
    \end{tabular} \\
    \midrule
    ConceptNet       & 581               & 
    \begin{tabular}[c]{@{}p{1\linewidth}} 
        - A ball is round. \\
        - A desire to share knowledge would make you want to teach other people. \\
        - If you want to compete against someone then you should enter a competition. \\
        - Something that might happen as a consequence of having a conversation is exchange of information. \\
        - The first thing you do when you surf the web is turn on your computer.
    \end{tabular} \\
    \midrule
    ATOMIC       & 697               & 
    \begin{tabular}[c]{@{}p{1\linewidth}} 
        - If max accepts sam invitation then max intends to to gain person's friendship. \\
        - If max [\emph{sic}] about to get married then max wants to be happily married. \\
        - If max accepts the job then max intends to to be employed. \\
        - If max accompanies sam far then max intends to to keep sam company. \\
        - If max attracts sam's attention then sam feels curious about max.
    \end{tabular} \\
    \midrule
    Aphorisms       & 709               & 
    \begin{tabular}[c]{@{}p{1\linewidth}} 
        - A doctor and a farmer know more than a doctor alone. \\
        - It is always the best policy to speak the truth unless of course you are an exceptionally good liar. \\
        - Let us never negotiate out of fear but let us never fear to negotiate. \\
        - Live your life as though your every act were to become a universal law. \\
        - Without craftsmanship inspiration is a mere reed shaken in the wind.
    \end{tabular} \\
    \midrule
    Situational response       & 630               & 
    \begin{tabular}[c]{@{}p{1\linewidth}} 
        - If a person digs themselves into a hole they will be in a bad mood. \\
        - If alex calls sam's mother they would have a conversation about sam. \\
        - If alex lands a job they will arrive early the first day of work to make a good impression. \\
        - If alex sees a bat he will poke the bat. \\
        - If alex works the night shift they will sleep during the day so they are awake for work.
    \end{tabular} \\
    \midrule
    Categorical response       & 832               & 
    \begin{tabular}[c]{@{}p{1\linewidth}} 
        - 5 is a lot bigger than 1. \\
        - Advancements in technology can help save lives and improve the quality of life for everyone in the world. \\
        - Computers can never be an actual human. \\
        - If john wants to be a teacher he would apply for a teaching position. \\
        - Pausing before proceeding can lead to better results.
    \end{tabular} \\
    \bottomrule
    \end{tabular}
    \caption[]{Seven sources of statements in our corpus, taken from Whiting and Watts~\citepSI{whitingFrameworkQuantifyingIndividual2024SI}. For ATOMIC~\citepSI{sapAtomicAtlasMachine2019SI}, people's names such as ``max'' and ``sam'' were used to replace entities such as ``PersonX'' and ``PersonY.''}
    \label{si:tab:statement_sources}
\end{table}

Here, the subscript $i$ denotes the index of a statement; the superscripts $h$ and $a$ indicate that this is the distribution based on human ratings, and with respect to question (a), respectively. For example, if 70\% of people who were assigned to rate statement $i$ agreed with it, then $d_{i}^{h, a} = 0.7$. If we take the mode of this distribution, the \emph{human majority rating} of each statement $i$ is
\begin{align}
    \texttt{majority}_i^h = \mathbbm{1} \left[ d_{i}^{h, a} \geq 0.5 \right],
    \label{si:eq:human_majority_rating}
\end{align}
where the indicator function $\mathbbm{1}[\cdot]$ is equal to 1 if the given argument is true and 0 otherwise. Unlike the human rating distribution, the human majority rating is a binary variable and is equal to 1 if and only if at least half of the participants who rated $i$ agreed with it. Note that we resolve ties, where exactly half of the participants agreed with a statement, with a final label of 1, or ``agree.'' Note also that we only count ratings by participants in $\Omega_i$ because they were the only ones assigned to rate statement $i$.

Similarly, the proportion of human participants who think that most other people would agree with statement $i$ (\ie, by answering ``yes'' to question (b)) is
\begin{align}
    d_{i}^{h, b} = \frac{1}{|\Omega_i|} \sum_{j \in \Omega_i} B_{i, j}.
    \label{si:eq:human_ratings_dist_q2}
\end{align}

\begin{figure}[t!]
    \centering
    \begin{subfigure}[t]{1\textwidth}
        \centering
        \includegraphics[height=0.55\textwidth]{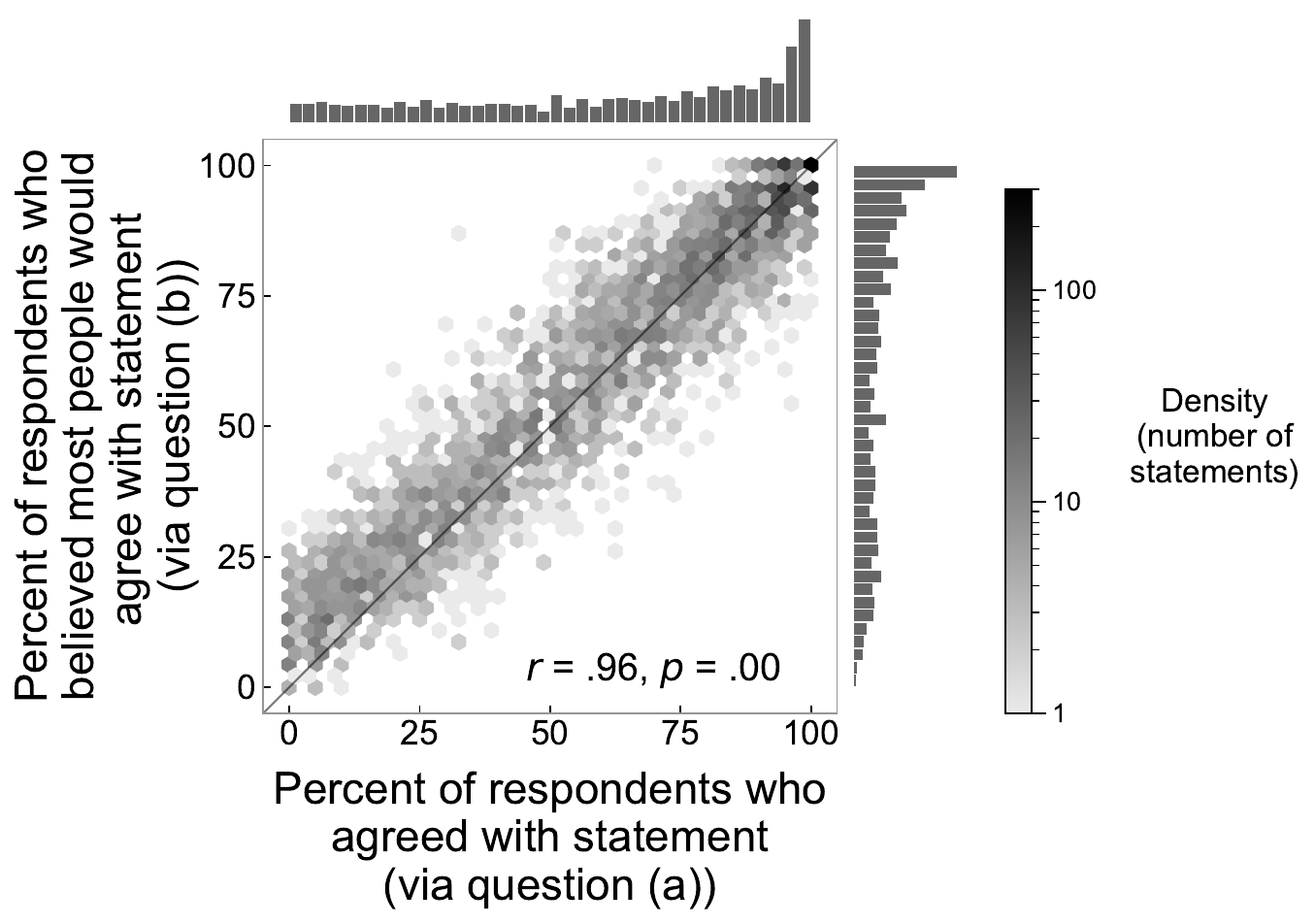}
        \label{si:fig:commonsensicality:human_rating_distribution}
    \end{subfigure}%
    \caption[]{Summary of the human ratings collected by Whiting and Watts~\citepSI{whitingFrameworkQuantifyingIndividual2024SI}. Participants (\textit{N} = 2,046) were given a statement (\textit{N} = 4,407) and asked to indicate (a) whether they agreed with the statement and (b) whether they thought most people would agree with the statement. Each participant was assigned only 50 randomly chosen statements, and on average, each statement received 23 unique ratings. Every statement is depicted in this figure, where the \textit{x}-axis represents the percentage of people who agreed with it, $d_i^{h, a}$ in \Cref{si:eq:human_ratings_dist}, and the \textit{y}-axis represents the percentage of people who believed most others would agree with it, $d_i^{h, b}$ in \Cref{si:eq:human_ratings_dist_q2}.}
    \label{si:fig:human_ratings_summary}
\end{figure}

\Cref{si:fig:human_ratings_summary} on page \pageref{si:fig:human_ratings_summary} shows the proportions of human participants who responded ``yes'' to questions (a) and (b) for all statements.

\subsection{Statement-Level Commonsensicality}
\label{si:sec:commonsense_recap:statement_commonsensicality}

The commonsensicality of a statement is a combination of its two scores, consensus and awareness.

Given a statement $i$, its \emph{consensus} score measures how close people's judgments of this statement are to absolute unanimity:
\begin{align}
    c_i^h = 2 \times \left| d_{i}^{h,a} - 0.5 \right|.
    \label{si:eq:statement_consensus}
\end{align}
High consensus requires that the human rating distribution $d_{i}^{h,a}$ be very close to 0 (when everyone disagrees with $i$) or 1 (when everyone agrees with $i$). Note that this definition is symmetric with respect to the polarity of opinions. For example, a statement that was agreed upon by 60\% of participants would have the same consensus score as a statement that was disagreed upon by 60\% of participants.

In addition, given statement $i$, its \emph{awareness} score measures how accurately people perceived its majority rating. Formally, we have
\begin{align}
    a_i^h = \frac{1}{|\Omega_i|} \sum_{j \in \Omega_i} \mathbbm{1} \left[ B_{i, j} = \texttt{majority}_i^h \right].
    \label{si:eq:statement_awareness}
\end{align}
High awareness requires that people know what most others believe, which is recorded in question (b) above.

For a statement to be common sense, it must score high in both consensus and awareness. The \emph{commonsensicality} score of statement $i$ is defined as the geometric mean of its consensus and awareness scores:
\begin{align}
    m_i^h = \sqrt{c_i^h \times a_i^h}.
    \label{si:eq:statement_commonsensicality}
\end{align}
The commonsensicality score $m_i^h$ is also between 0 and 1, and values closer to 1 indicate higher degrees of common sense.

\Cref{si:fig:commonsensicality} (bottom) on page \pageref{si:fig:commonsensicality} presents statements on two dimensions: consensus and awareness. Generally, statements received higher awareness scores than consensus scores. The figure also shows that most statements received a high commonsensicality score, with a median of 70.12\%.

\subsection{Human-Level Commonsensicality}
\label{si:sec:commonsense_recap:human_commonsensicality}

Commonsensicality is also defined for people, comprising consensus and awareness scores.

Given a person $j$, their \emph{consensus} score is defined as the fraction of times they agreed with the majority rating across all statements they rated:
\begin{align}
    C_j = \frac{1}{|\Phi_j|} \sum_{i \in \Phi_j} \mathbbm{1} \left[ A_{i, j} = \texttt{majority}_i^h \right].
    \label{si:eq:human_consensus}
\end{align}
To have high consensus, person $j$'s own ratings of the statements, via their answer to question (a) above, must coincide with the overall human majority ratings.

The \emph{awareness} score of person $j$ is the fraction of time they accurately predicted the majority rating across all statements:
\begin{align}
    A_j = \frac{1}{|\Phi_j|} \sum_{i \in \Phi_j} \mathbbm{1} \left[ B_{i, j} = \texttt{majority}_i^h \right].
    \label{si:eq:human_awareness}
\end{align}
To have a high awareness score, person $j$'s prediction of what most other people think---via question (b) above, and regardless of what they chose for question (a)---must also coincide with the overall human majority ratings.

\begin{figure*}[p!]
    \centering
    \begin{subfigure}[t]{1\textwidth}
        \centering
        \includegraphics[height=0.55\textwidth]{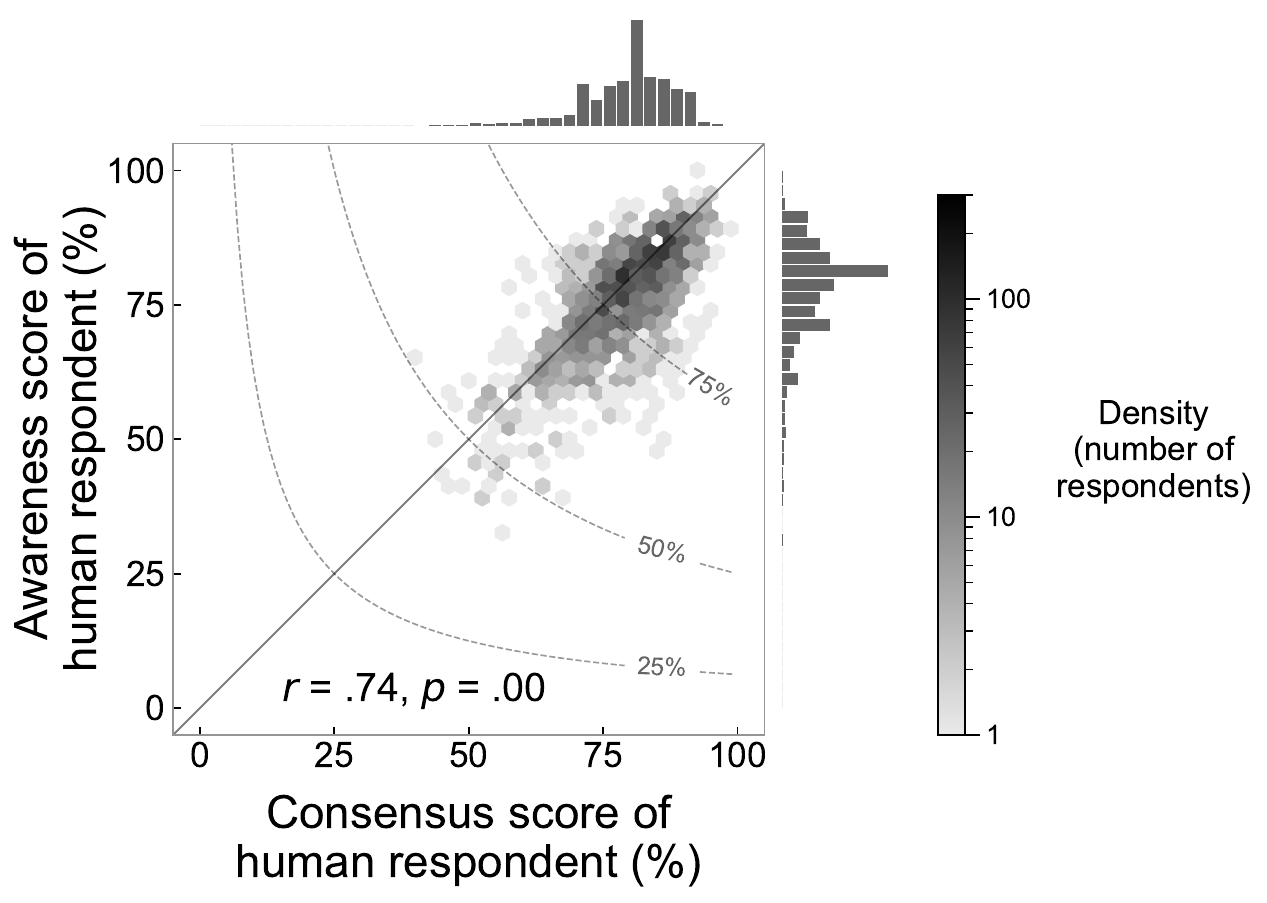}
    \end{subfigure} 
    \\
    \begin{subfigure}[t]{1\textwidth} 
        \centering
        \includegraphics[height=0.55\textwidth]{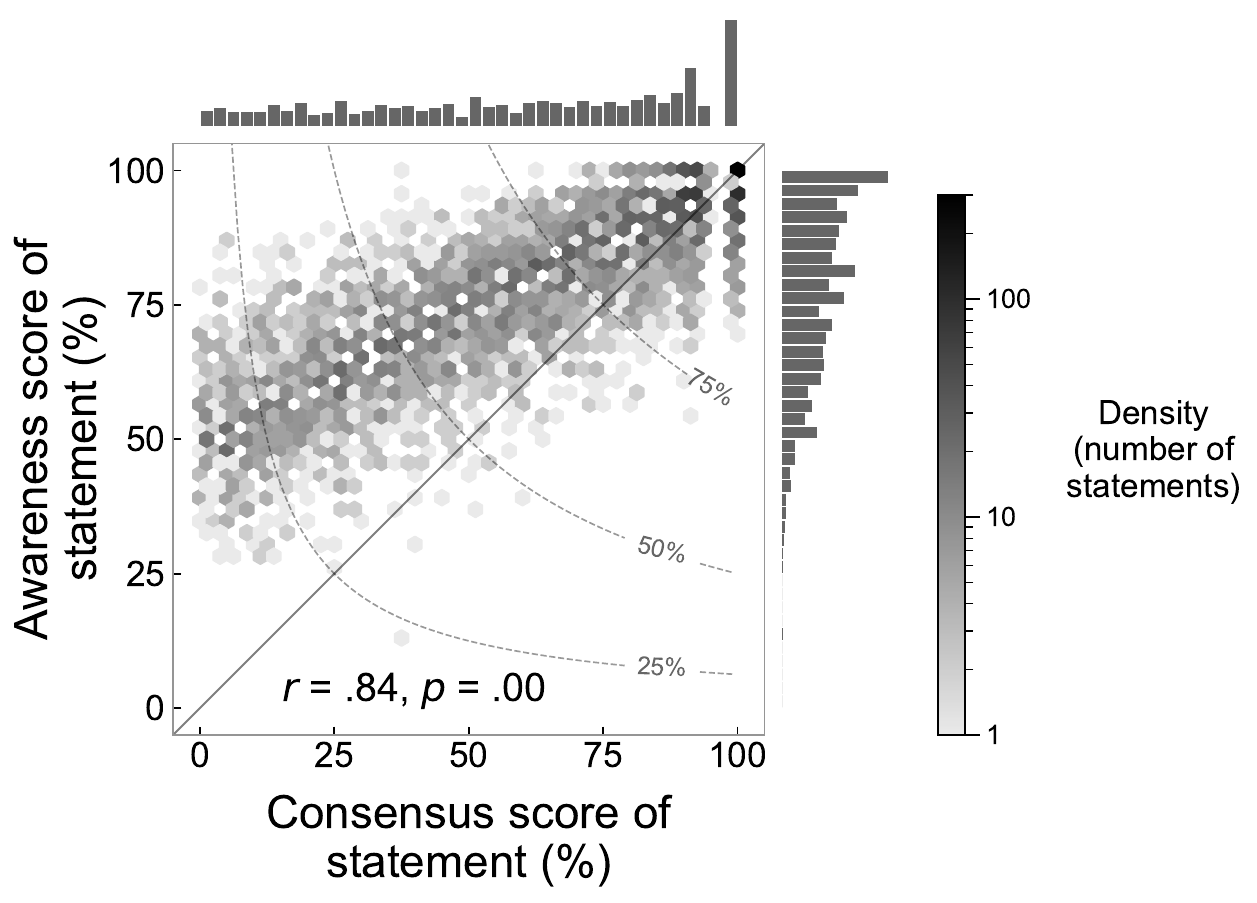}
    \end{subfigure}
    \caption{Top: Individual-level commonsensicality, $M_j$ in \Cref{si:eq:human_commonsensicality}, defined for every human participant. Bottom: collective-level commonsensicality, $m_i^h$ in \Cref{si:eq:statement_commonsensicality} defined for every statement. The commonsensicality score is the geometric average of the consensus and awareness scores, which are depicted in these figures. The three level curves depict combinations of consensus and awareness that produce three levels of commonsensicality: 25\%, 50\%, and 75\%.}
    \label{si:fig:commonsensicality}
\end{figure*}

Consensus $C_j$ and awareness $A_j$ scores are also between 0 and 1. Similarly, we take the geometric mean of these scores to give participant $j$'s \emph{commonsensicality} score as 
\begin{align}
    M_j = \sqrt{C_j \times A_j}.
    \label{si:eq:human_commonsensicality}
\end{align}
This score is also between 0 and 1, with higher values indicating higher degrees of common sense in a person.

\Cref{si:fig:commonsensicality} (top) also presents humans on two dimensions: consensus and awareness. Different from statements, humans tended to exhibit very similar scores on both dimensions (most of them lie close to the diagonal line). The median commonsensicality for humans is 79.43\%.

\newpage

\section{Collecting Responses from Large Language Models}
\label{si:sec:lm_prompt}

\begin{table}[p!]
    \small
    \centering
    \begin{tabular}{llccccl}
    \toprule
    \textbf{Model family}                 & \textbf{Model name}            & \textbf{Model size} & \textbf{Access} & \textbf{Source} & \textbf{Released} & \textbf{Comments} \\
    \midrule
    \multirow{3}{*}{Claude}     & Claude 3 Haiku       & --                   & Closed   &  \multirow{3}{*}{\citepSI{anthropicClaude3Model2024SI}} & \multirow{3}{*}{03/24} & Most capable      \\
                                 & Claude 3 Sonnet      & --                  & Closed   & & & Medium      \\
                                 & Claude 3 Opus      & --                  & Closed   & &  & Least capable     \\
    \midrule
    DBRX & dbrx-instruct & 132B & Open & \citepSI{mosaicIntroducingDBRXNew2024SI} & 03/24  \\
    \midrule
    \multirow{3}{*}{Falcon} & Falcon-7b-Instruct    & 7B                   & Open   &  \multirow{3}{*}{\citepSI{almazroueiFalconSeriesOpen2023SI}}  & \multirow{2}{*}{06/23}    \\
                                 & Falcon-40b-Instruct   & 40B                  & Open   &        \\
                                 & Falcon-180B-Chat      & 180B                 & Open   &  & 09/23     \\
    \midrule
    \multirow{5}{*}{Flan-T5}     & Flan-T5-Small         & 80M                  & Open   &  \multirow{5}{*}{\citepSI{chungScalingInstructionfinetunedLanguage2022SI}}  & \multirow{5}{*}{10/22} &    \\
                                 & Flan-T5-Base          & 250M                 & Open   &        \\
                                 & Flan-T5-Large         & 780M                 & Open   &        \\
                                 & Flan-T5-XL            & 3B                   & Open   &        \\
                                 & Flan-T5-XXL           & 11B                  & Open   &        \\
    \midrule
    \multirow{3}{*}{Gemini}      & Gemma-2b-it           & 2B                   & Open   &  \multirow{2}{*}{\citepSI{googleGemmaOpenModels2024SI}} & \multirow{3}{*}{02/24} &     \\
                                 & Gemma-7b-it           & 7B                   & Open   &        \\
                                 & Gemini Pro 1.0        & --                  & Closed & \citepSI{googleGemini15Unlocking2024SI} & & Original release      \\
    \midrule
    \multirow{5}{*}{GPT}         & GPT-3.5-Turbo-0125    & 175B                 & Closed &   \citepSI{brownLanguageModelsAre2020SI}  & 01/24 & Original model 2020  \\
                                 & GPT-4-Turbo-0125      & --                  & Closed &  \multirow{2}{*}{\citepSI{openaiGPT4TechnicalReport2023SI}} & 01/24 & \multirow{2}{*}{Original model 2023} \\
                                 & GPT-4-Turbo-0409      & --                  & Closed &  & 04/24    \\
                                 & GPT-4o                & --                  & Closed &  \citepSI{openaiGPT4oSystemCard2024SI} & 05/24     \\
                                 & GPT-5                 & --                  & Closed &  \citepSI{openaiGPT5SystemCard2025SI}   & 08/25 & Reasoning model \\
    \midrule
    \multirow{5}{*}{LLaMA}     & LLaMA-2-7B-Chat       & 7B                   & Open   &  \multirow{3}{*}{\citepSI{touvronLlama2Open2023SI}}  & \multirow{3}{*}{07/23} &    \\
                                 & LLaMA-2-13B-Chat      & 13B                  & Open   &        \\
                                 & LLaMA-2-70B-Chat      & 70B                  & Open   &        \\
         & LLaMA-3-8B-Instruct       & 7B                   & Open   &  \multirow{2}{*}{\citepSI{grattafioriLlama3Herd2024SI}}  & \multirow{2}{*}{04/24} &  \\
                                 & LLaMA-3-70B-Instruct      & 70B                  & Open   &        \\
    \midrule
    \multirow{4}{*}{Mistral}     & Mistral-7B-Instruct   & 7B                   & Open   &  \citepSI{jiangMistral7B2023SI} & 09/23 &     \\
                                 & Mixtral-8x7B-Instruct & 47B                  & Open   & \citepSI{jiangMixtralExperts2024SI}   & 11/23 &    \\
                                 & Mixtral-8x22B-Instruct & 141B                  & Open   & \citepSI{mistralaiCheaperBetterFaster2024SI}  & 04/24 &     \\
                                 & Mistral-Large & --                  & Closed   & \citepSI{mistralaiAuLarge2024SI} & 02/24 &      \\
    \midrule
    OLMo                         & OLMo-7B-Instruct      & 7B                   & Open   &  \citepSI{groeneveldOLMoAcceleratingScience2024SI} & 02/24 &     \\
    \midrule
    \multirow{5}{*}{Qwen}     & Qwen2-0.5B-Instruct   & 0.5B                   & Open   & \multirow{5}{*}{\citepSI{yangQwen2TechnicalReport2024SI}} & \multirow{5}{*}{06/24} &      \\
                                 & Qwen2-1.5B-Instruct & 1.5B                 & Open   &   &  &    \\
                                 & Qwen2-7B-Instruct & 7B                & Open   &  &  &     \\
                                 & Qwen2-57B-Instruct & 57B                  & Open   &  & &      \\
                                 & Qwen2-72B-Instruct & 72B                 & Open   & &  &      \\
    
    \bottomrule
    \end{tabular}
    \caption{ Language models used for our evaluation of commonsensicality. All closed-source models (Claude 3, GPT-3.5/4/5 and Mistral-Large) are accessed via their developers' APIs. All open-weight models are accessed via Hugging Face in Python. For closed-source models, the ``--'' sign denotes their unknown sizes (number of parameters).}
    \label{si:tab:model_list}
\end{table}

To assess common sense in LLMs, we select a number of models that have been trained to follow conversations (instruction-tuned models). The full list of models we experiment with is found in \Cref{si:tab:model_list} on page \pageref{si:tab:model_list}. A typical conversation with an LLM is as follows.
\begin{itemize}
    \item[] \texttt{[SYSTEM PROMPT]}
    \item[] \texttt{[USER PROMPT]}
    \item[] \texttt{[ASSISTANT RESPONSE]}
    \item[] \texttt{[USER PROMPT]}
    \item[] \texttt{[ASSISTANT RESPONSE]}
    \item[] \texttt{\ldots}
\end{itemize}
The conversation starts with an optional \emph{system prompt}, which is used to steer the general behavior of the LLM. For example, one can set this to ``You are a debate expert'' if the model is intended to be used to make arguments. Then, the system prompt is followed by a series of alternating \emph{user prompts} and \emph{assistant responses}. All messages in this conversation are concatenated into a \emph{chat history}, which is a string of the form \texttt{[SYSTEM PROMPT] [SEP] [USER PROMPT] [SEP] [ASSISTANT RESPONSE] [SEP] [USER PROMPT] [SEP] \newline [ASSISTANT RESPONSE]\ldots} The \texttt{[SEP]} symbol is a special token denoting the separation of messages, and is specific to each model.

To use an LLM to generate an assistant response, one prepares a chat history that ends with a user prompt, and hence expects an assistant response to follow from it. The model takes the chat history as input and autoregressively generates new tokens one by one, until a finishing condition is reached. These tokens are then appended to the chat history, thereby completing the assistant response. Instruction-tuned models are specifically trained in this manner, via many such conversations, and hence we choose these models for assessment.

\Cref{si:code:lm_probs} shows a concrete example of our procedure, which is described in more detail in the subsections below.

\subsection{Designing Prompts and Expected Responses}
\label{si:sec:lm_prompt:prompts}

We decide not to use any specific system prompt. For some models, this is set to their default system prompts, such as ``You are a helpful assistant.'' For some other models, this can be left blank, so the chat history actually starts with a user prompt.

For each of the $n_s = 4,407$ statements, we prompt a model to respond to the following two use prompts:
\begin{itemize}
    \item[(a)] \texttt{Consider the following statement, "[STATEMENT]" Do you agree with\\ this statement? Start your answer with a "yes" or "no".}
    \item[(b)] \texttt{Consider the following statement, "[STATEMENT]" Do you  think most\\ people would agree with this statement? Start your answer with a \\"yes" or "no".}
\end{itemize}

These are the same as questions (a) and (b) described in \Cref{si:sec:commonsense_recap}. We note three details of these prompts. First, questions (a) and (b) are used in two separate chat sessions. This is because any chat history can modify a model's answer in unexpected ways, which we do not like to experience. Second, we include the sentence ``Start your answer with a `yes' or `no'.'' at the end of each prompt to instruct the model to output a ``yes'' or ``no'' directly. Without this explicit instruction, the model may start its response with, for example, ``I agree with the answer,'' or give some reasoning before giving its final answer. Either way, extracting the binary answer would be more complicated and noisy. And third, in framing the questions in this way, we need the model to output only one token---``yes'' or ``no''---and we can extract the probability with which it generates either token, which represents the model's inherent uncertainty in its answer (more below).

\clearpage
\nolinenumbers
\begin{lstlisting}[
    language=Python,
    caption={\scriptsize Python code used to extract an open-source model's answer to an example prompt. Here, we demonstrate this process using the model DBRX. Note that there could be multiple tokens within this model's vocabulary that can be mapped to the answers ``yes'' and ``no,'' hence we need to enumerate them in the \texttt{answer2tokid} dictionary on lines 14--17. Based on the prompt (lines 20--23), the model is supposed to start its response with a definitive ``yes'' or ``no.'' Therefore, we perform one feedforward operation to get the probabilities of all tokens that can be generated next (lines 28--34). We then sum the probabilities of all tokens that map to ``yes'' and ``no'' to get the model's probability of answering with either option (lines 36--44).},
    captionpos=b,
    label={si:code:lm_probs}
]
# Load tokenizer and model
tokenizer = AutoTokenizer.from_pretrained(MODEL_NAME)
model = model = AutoModelForCausalLM.from_pretrained(
    "databricks/dbrx-instruct",
    device_map="auto",
    trust_remote_code=True,
    torch_dtype=torch.bfloat16,
)
model.eval()
model.tie_weights()

# These are all the IDs of the tokens in DBRX's vocabulary that map to 
# the answers "yes" and "no".
answer2tokid = {
    "yes": [9642, 60844, 14331, 58841, 20137, 60665, 7566, 85502, 77830, 95934, 10035, 86508, 98171, 41898, 14410, 9891], 
    "no": [72719, 6673, 42257, 86176, 43983, 67579, 2822, 79027, 18847, 40305, 34200, 58749, 5782, 38089, 39522, 73204, 9173, 99076, 2360, 38557, 31415, 12674, 85298, 29466, 2201, 912, 61559, 17184, 51899, 9278]
}

# Formulate the prompt
messages = [
    {"role": "user",
     "content": 'Consider the statement, "A desire to share knowledge would make you want to teach other people." Do you agree with this statement? Start your answer with a "yes" or "no".'}
]

# Encode the message
inputs = tokenizer.apply_chat_template(messages, add_generation_prompt=True)

# Feedforward the input
inputs = torch.tensor(inputs, device=model.device, dtype=torch.long).reshape(1, -1)
outputs = model(inputs, output_hidden_states=False)

# Extract the probability of immediate next token
probs = torch.softmax(outputs[0][0, -1], axis=0)
probs = probs.cpu().detach().float().numpy()

# Map token IDs to answers
answer_probs = {}
# The probability that the model answers "yes" is the sum of the
# probabilities of all tokens that map to "yes".
answer_probs["yes"] = float(probs[answer2tokid["yes"]].sum())
# Similar for "no".
answer_probs["no"] = float(probs[answer2tokid["no"]].sum())
# The probability that the model answers with something else.
answer_probs["other"] = 1 - answer_probs["yes"] - answer_probs["no"]
\end{lstlisting}

\clearpage

\subsection{Extracting Responses from LLMs}
\label{si:sec:lm_prompt:responses}

Questions (a) and (b) in \Cref{si:sec:lm_prompt:prompts} are formulated as user prompts for each model, which is then expected to generate a response starting with a ``yes'' or ``no.'' As explained above, the model does so by autoregression: it maintains a multinomial probability distribution over all possible tokens and randomly chooses a token to generate via sampling from this distribution. In other words, the tokens ``yes'' and ``no'' are each assigned a probability with which they are generated. This also means that every other token may be generated, as long as its probability is not zero, leading to ``illegal'' answers. We address this problem in the following.

First, a review of how we can obtain token probabilities for each model.
\begin{itemize}
    \item For GPT-3.5 and GPT-4, OpenAI's API allows us to have access to up to 20 tokens associated with the highest sampling probabilities via the \texttt{logprobs} parameter.
    \item For Gemini 1.0 Pro, Claude 3 and Mistral-Large, their APIs do not allow access to these probabilities. We repeatedly ask the same question multiple times and report the empirical frequencies of the generated tokens, which give us an unbiased estimate of the probabilities. We perform 23 repetitions for Claude 3 and Mistral-Large, which matches the average number of people that rated each question in the original dataset~\citepSI{whitingFrameworkQuantifyingIndividual2024SI}. For Gemini 1.0 Pro, we perform 50 repetitions.
    \item For GPT-5, we also perform repeated sampling. However, this is a reasoning model which tends to output its ``thinking'' before giving us its answer, and this tendency cannot be turned off. Thus, in the user prompt, we add the following message---and thus slightly deviate from our preregistration---to discourage this thinking behavior: ``Do not include anything else, such as an explanation or reasoning.'' Each API call allows for 8 repetitions (or ``choices''), so for each statement we make 6 API calls, resulting in 48 ratings per statement in total.
    \item For all other models, which are open-weight and loaded via the Hugging Face library in Python, we have access to the probabilities of all tokens.
\end{itemize}
In generating tokens, we set all sampling parameters to default; most notably, we keep the sampling temperature at 1.0. Once we have the probabilities of generated tokens, we perform the following processing steps.
\begin{itemize}
    \item We discard all tokens that do not represent ``yes'' or ``no'' answers. This usually happens when a model wants to avoid giving an answer, for example, ``As an AI model I cannot\ldots'' which leads to the first generated token being ``As,'' an invalid token.
    \item For some models, their tokenizers have different variants of the same answer, due to formatting reasons. For example, the answer ``yes'' can be generated from the tokens \texttt{Yes}, \texttt{yes}, \verb*| yes|, \texttt{"Yes}, \etcetera, which should all be valid. In this case, we compile a list of all such valid tokens and sum over their probabilities to get the final score.
    \item After this, we have probabilities for the ``yes'' and ``no'' answers. Since we have discarded all illegal tokens, these two probabilities do not necessarily add up to one. We hence rescale them by dividing each by their sum. See \Cref{si:code:lm_probs}.
\end{itemize}
The resulting binomial distribution of answers is called the \emph{model rating distribution} and denoted
\begin{align}
    d_{i}^{m, a} = p(\text{``yes''} \mid \text{prompt (a) for statement } i).
    \label{si:eq:model_ratings_dist}
\end{align}
where $i$ denotes a statement and $m$ denotes a model ($m \in \{1, \ldots, n_m\}$, $n_m$ being the number of models). Compare this with the human rating distribution in \Cref{si:eq:human_ratings_dist} on page \pageref{si:eq:human_ratings_dist}.

Similarly, the probability that the model thinks most people would agree with statement $i$ is 
\begin{align}
    d_{i}^{m, b} = p(\text{``yes''} \mid \text{prompt (b) for statement } i).
    \label{si:eq:model_ratings_dist_q2}
\end{align}

\begin{figure*}[p!]
    \centering
    \begin{subfigure}[t]{1\textwidth}
        \centering
        \includegraphics[width=0.9\textwidth]{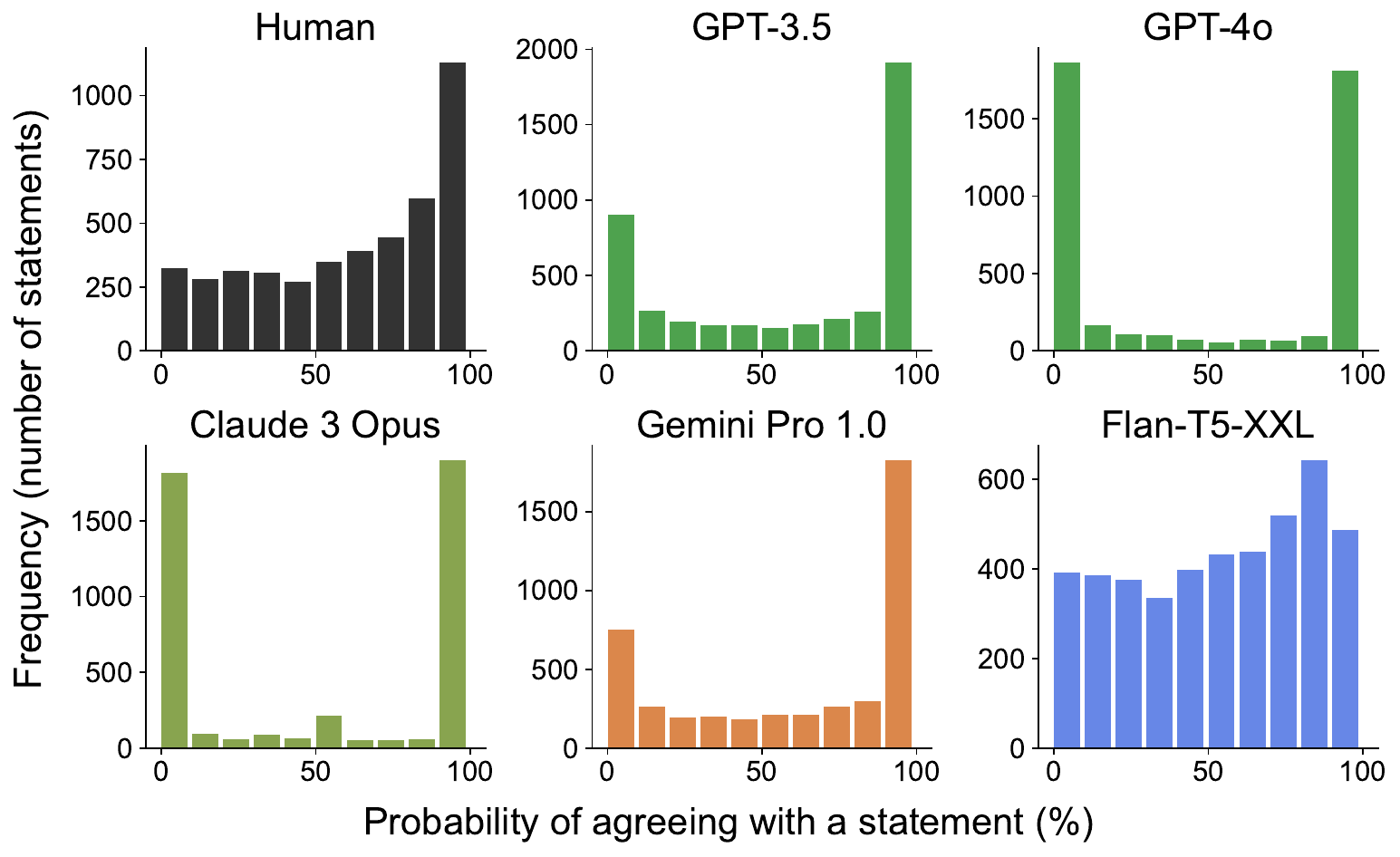}
    \end{subfigure} 
    \par\bigskip
    \vspace{2em}
    \begin{subfigure}[t]{1\textwidth} 
        \centering
        \includegraphics[width=0.9\textwidth]{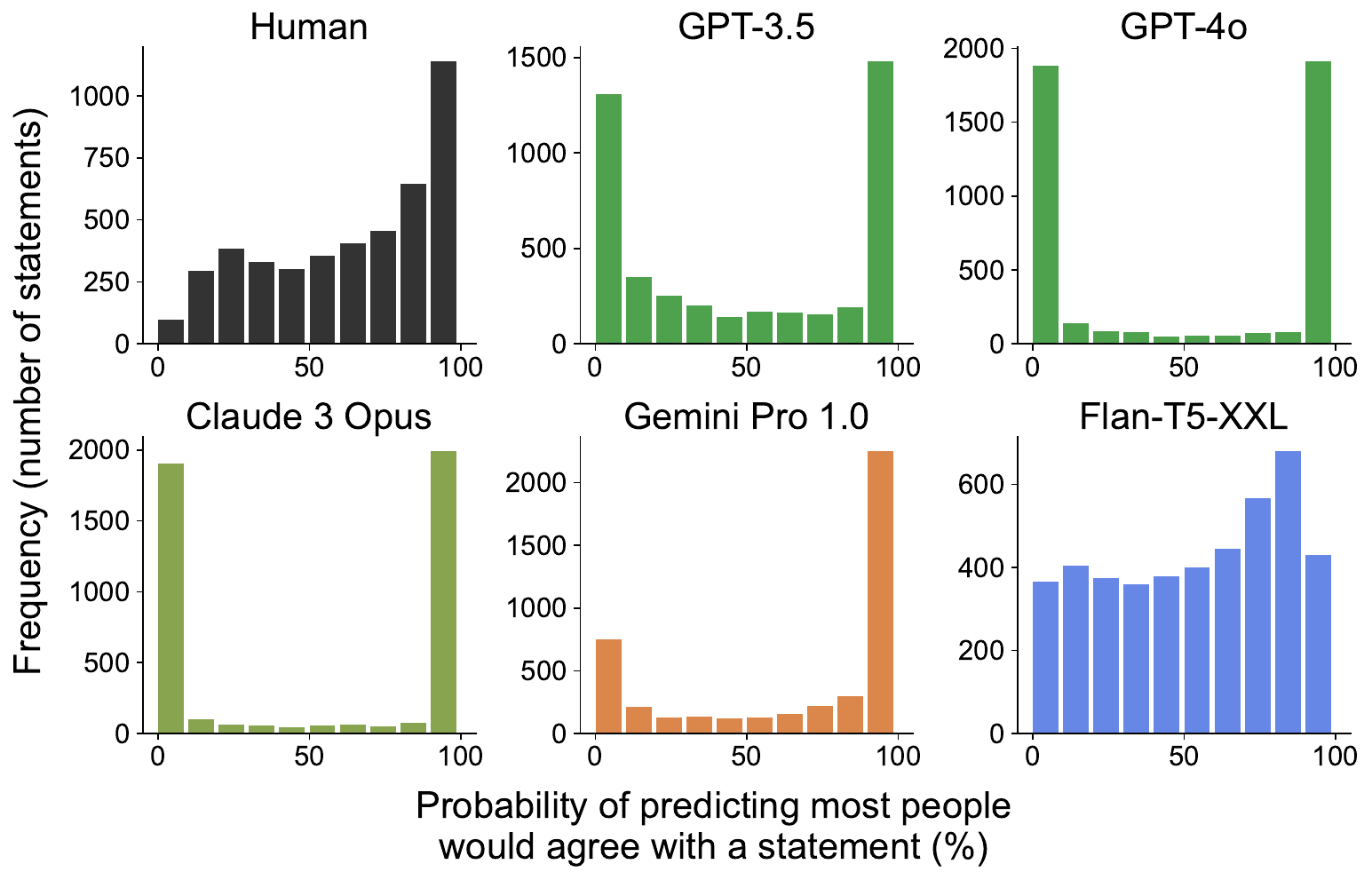}
    \end{subfigure} 
    \caption{ Top: frequencies of the human rating distribution, $d_{i}^{h, a}$ (\Cref{si:eq:human_ratings_dist}), and the model rating distribution $d_{i}^{m, a}$ (\Cref{si:eq:model_ratings_dist}) for some models. These are derived from responses to question (a). Bottom: distribution of the same quantities, but for question (b).}
    \label{si:fig:probs_q1_freqs_humans_models}
\end{figure*}

\Cref{si:fig:probs_q1_freqs_humans_models} (top) on page \pageref{si:fig:probs_q1_freqs_humans_models} shows the frequencies of the human and model rating distributions. As depicted, human judgments are widely varied, with many statements receiving close to 50/50 ratings. For most models, on the other hand, these probabilities are predominantly driven to either 0 or 1. The same is observed for humans' and models' answers to question (b).

\begin{figure}[p!]
    \centering
    \includegraphics[width=1\linewidth]{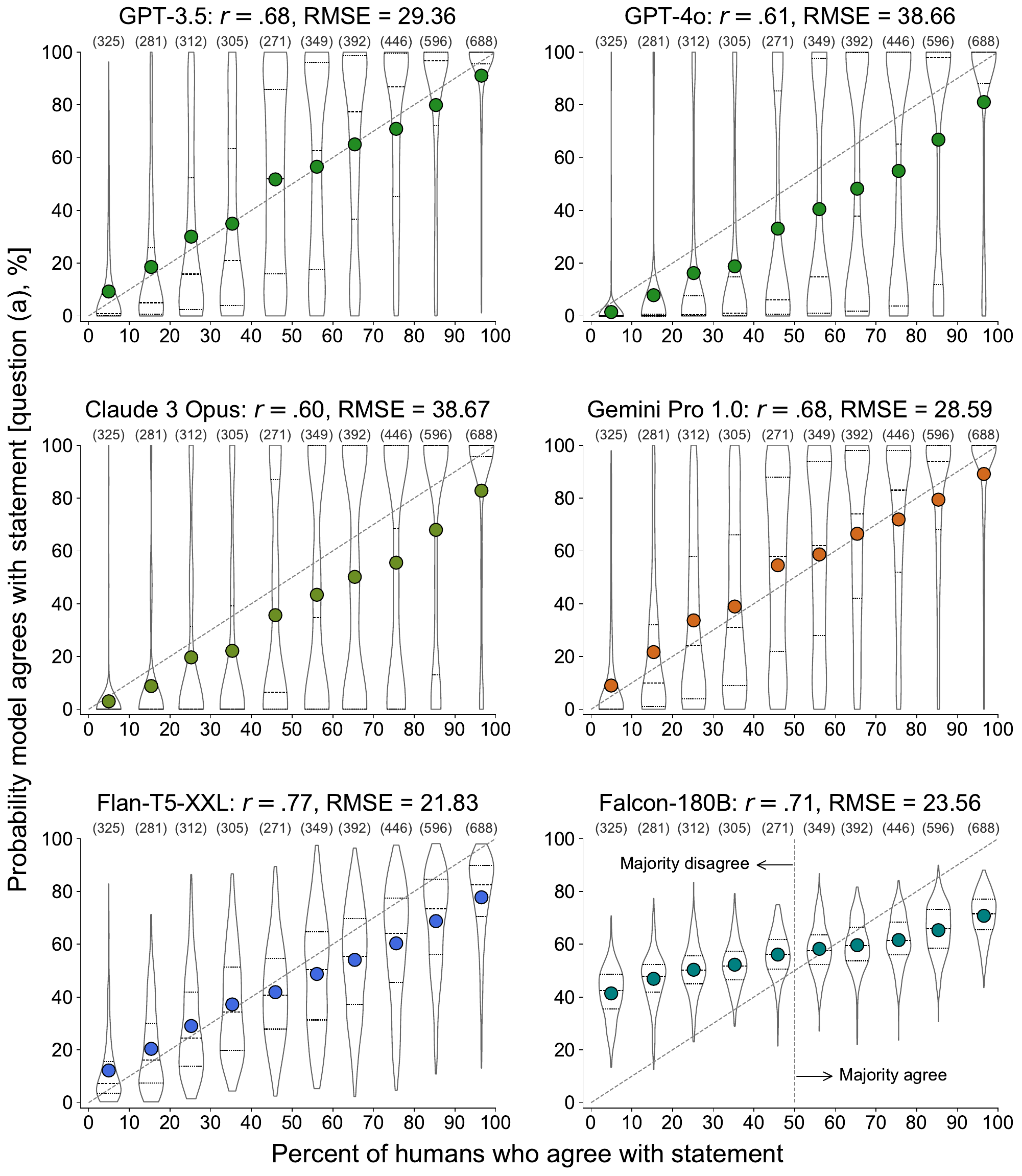}
    \caption{Relationship between human and model ratings of statements. The \textit{x}-axis is the percentage of humans who expressed that they agree with a statement, divided into ten intervals. A value of at least 50\% on the \textit{x}-axis means that the majority of humans agree with the statement. The \textit{y}-axis is the probability that an LLM agrees with a statement, in response to question (a): ``Do you agree with this statement?'' For each violin plot, the number on top and in parentheses is the number of statements in the corresponding interval; the three horizontal lines within the plot from top to bottom depict the 75th, 50th and 25th percentiles of the distribution; and the point within the plot shows the average y-value for statements within each bin. Also shown on top of each subplot are the Pearson correlation $r$ (all significant with $p < .001$) and the root mean squared error (RMSE).}
    \label{si:fig:calibration_q1}
\end{figure}

\begin{figure}[p!]
    \centering
    \includegraphics[width=1\linewidth]{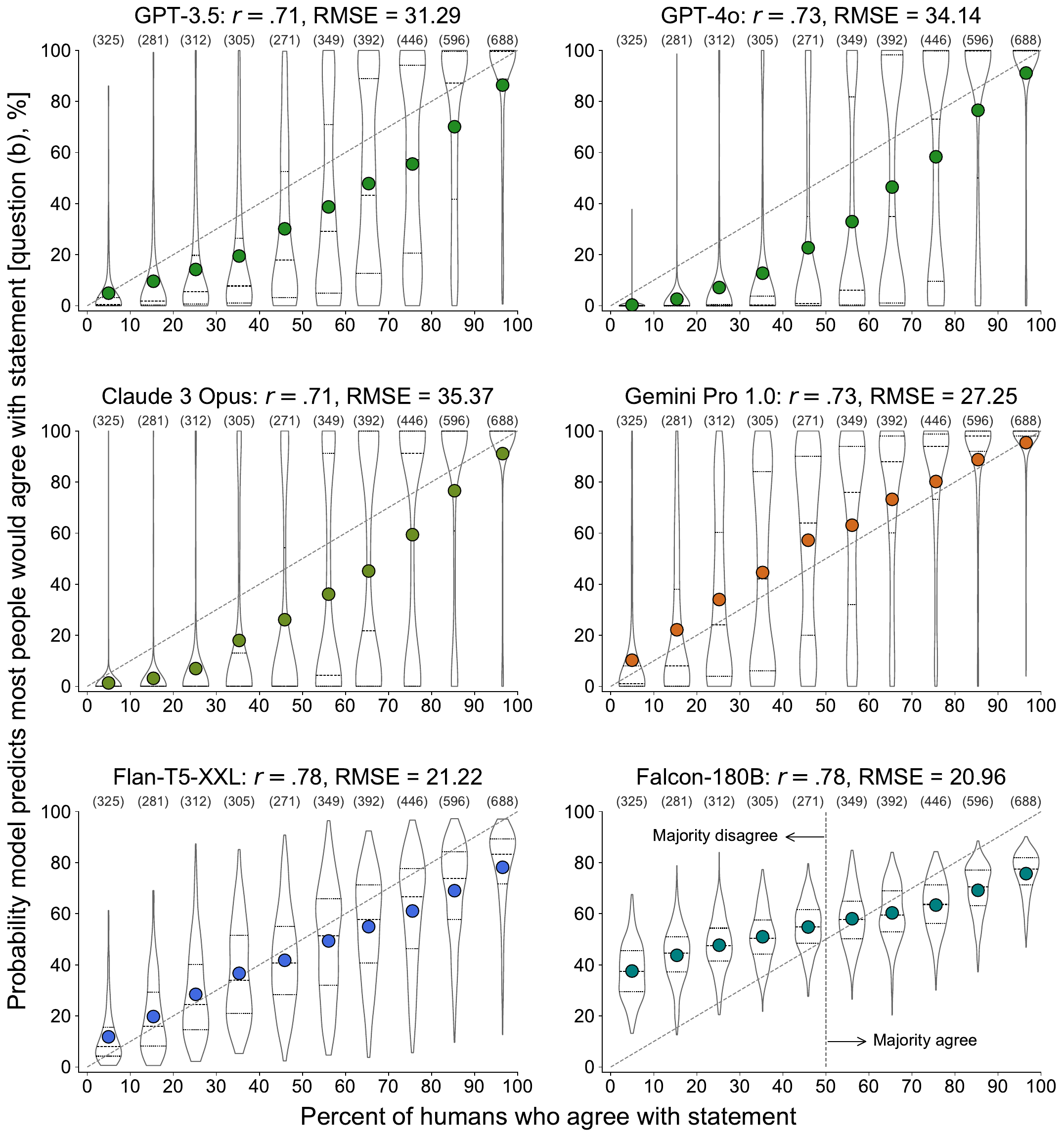}
    \caption{Relationship between human and model ratings of statements. The \textit{x}-axis is the percentage of humans who expressed that they agree with a statement, divided into ten intervals. A value of at least 50\% on the \textit{x}-axis means that the majority of humans agree with the statement. The \textit{y}-axis is an LLM's predicted probability that most people would agree with the statement, in response to question (b): ``Do you think most people would agree with this statement?'' For each violin plot, the number on top and in parentheses is the number of statements in the corresponding interval; the three horizontal lines within the plot from top to bottom depict the 75th, 50th and 25th percentiles of the distribution; and the point within the plot shows the average y-value for statements within each bin. Also shown on top of each subplot are the Pearson correlation $r$ (all significant with $p < .001$) and the root mean squared error (RMSE).}
    \label{si:fig:calibration_q2}
\end{figure}

\begin{table}[p!]

    \centering
    \small
    \renewcommand{\arraystretch}{1.4}
\begin{tabular}{p{0.1\linewidth}p{0.62\linewidth}c}
\toprule
\textbf{Source} & \textbf{Statement} & \textbf{Number (\%) of models} \\
\midrule
Aphorisms & Life is not separate from death it only seems that way. & 19 / 35 (54.3\%) \\
Aphorisms & One must discard the prejudice that the truth must be something tangible. & 18 / 35 (51.4\%) \\
News media & Public private partnerships should be adopted as a more important mechanism by congress. & 18 / 35 (51.4\%) \\
Aphorisms & We will bankrupt ourselves in the vain search for absolute security. & 17 / 35 (48.6\%) \\
Aphorisms & Philosophy is properly home sickness the wish to be everywhere at home. & 17 / 35 (48.6\%) \\
Aphorisms & The sage constantly has no set mind those who are good he regards as good and those who are not good he also regards as good in this way he attains goodness. & 17 / 35 (48.6\%) \\
Categorical response & The conflating of religion and god has been damning to humanity. & 16 / 35 (45.7\%) \\
Aphorisms & The miracles of earth are the laws of heaven. & 15 / 35 (42.9\%) \\
Aphorisms & Politeness has been defined to be artificial good nature but we may affirm with much greater propriety that good nature is natural politeness. & 15 / 35 (42.9\%) \\
Aphorisms & Live your life as though your every act were to become a universal law. & 15 / 35 (42.9\%) \\
Categorical response & When someone leaves a fast food store with a bag the bag will be full of food. & 15 / 35 (42.9\%) \\
Aphorisms & We act as though comfort and luxury were the chief requirements of life when all that we need to make us happy is something to be enthusiastic about. & 15 / 35 (42.9\%) \\
Campaign email & We should sign up for the mailing list. & 15 / 35 (42.9\%) \\
Aphorisms & Blessed is he who has found his work let him ask no other blessedness. & 14 / 35 (40.0\%) \\
Aphorisms & Chaos often breeds life when order breeds habit. & 14 / 35 (40.0\%) \\
Atomic & If max accidentally shot then max wants to call the ambulance. & 14 / 35 (40.0\%) \\
Aphorisms & Remember that life is neither pain nor pleasure it is serious business to be entered upon with courage and in a spirit of self sacrifice. & 14 / 35 (40.0\%) \\
News media & Governments should offer a path to citizenship for millions of illegal immigrants. & 14 / 35 (40.0\%) \\
Aphorisms & Doubt is a pain too lonely to know that faith is his twin brother. & 14 / 35 (40.0\%) \\
Campaign email & Lizzie fletcher will be a effective leader. & 14 / 35 (40.0\%) \\
Campaign email & We should support abortion and woman's right to choose. & 14 / 35 (40.0\%) \\
Campaign email & We should vote an underdog who supports a more civil government. & 14 / 35 (40.0\%) \\
Categorical response & No one religion is the right path. & 14 / 35 (40.0\%) \\
Aphorisms & Solitude produces originality bold and astonishing beauty poetry but solitude also produces perverseness the disproportionate the absurd and the forbidden. & 14 / 35 (40.0\%) \\
Aphorisms & The sage treasures everything so nothing is wasted. & 13 / 35 (37.1\%) \\
\bottomrule
\end{tabular}
\caption{Statements that receive diverging answers to questions (a) and (b) from LLM. For each statement, every LLM is asked two questions: (a) ``Do you agree with this statement?'' and (b) ``Do you think most people would agree with this statement?'' Most of the time, models give the same answer (either both ``yes'' or both ``no'') to these questions. In a minority of cases (about 10\%), we observe the opposite, where one question receives ``yes'' and the other ``no.'' This table shows the 25 statements that receive such divergence from the most models.}
\label{si:tab:nonconformity_statements}
\end{table}

\clearpage    

\newpage
\section{Measuring Individual-Level Commonsensicality in LLMs}
\label{si:sec:model_commonsensicality}

In this section, we consider each language model as a hypothetical human survey respondent, just like those described in \Cref{si:sec:commonsense_recap:human_commonsensicality}.

Start from $d_{i}^{m, a}$ (\Cref{si:eq:model_ratings_dist}), which is the probability that the model agrees with statement $i$, via its answer to question (a). We binarize this probabilistic output by taking its mode
\begin{align}
    \alpha_i^{m} = \mathbbm{1}[d_{i}^{m, a} \geq 0.5].
    \label{si:eq:model_binary_rating_q1}
\end{align}
In other words, we consider that the model agrees with statement $i$ (\ie, $\alpha_i^{m, a} = 1$) if and only if it agrees with this statement with a probability of at least 0.5.

Similarly, the model believes that most people would agree with this statement if and only if it answers ``yes'' to question (b) with a probability of at least 0.5:
\begin{align}
    \beta_i^{m} = \mathbbm{1}[d_{i}^{m, b} \geq 0.5].
    \label{si:eq:model_binary_rating_q2}
\end{align}

Based on these two binary answers, the model's \emph{consensus} and \emph{awareness} scores, respectively, are
\begin{align}
    C_m &= \frac{1}{n_s} \sum_{i=1}^{n_s} \mathbbm{1} \left[ \alpha_i^m = \texttt{majority}_i^h \right], \label{si:eq:model_consensus} \\
    A_m &= \frac{1}{n_s} \sum_{i=1}^{n_s} \mathbbm{1} \left[ \beta_i^m = \texttt{majority}_i^h \right],\label{si:eq:model_awareness}
\end{align}
where $\texttt{majority}_i^h$ is the human majority rating for statement $i$ (\Cref{si:eq:human_majority_rating}). Compare these calculations with the consensus and awareness scores for humans, in \Cref{si:eq:human_consensus,si:eq:human_awareness}.

Finally, similar to \Cref{si:eq:human_commonsensicality}, the model's \emph{commonsensicality} score is the geometric mean of its consensus and awareness scores: 
\begin{align}
    M_m = \sqrt{C_m \times A_m}.
    \label{si:eq:model_commonsensicality}
\end{align}
All three scores are between 0 and 1, and higher values indicate closer proximity to absolute common sense. The goal of this calculation is to compare $M_m$ with $M_j$, the commonsensicality of person $j$ defined in \Cref{si:eq:human_commonsensicality}.

\Cref{si:tab:model_individual_commonsensicality} shows every model's consensus, awareness and commonsensicality scores.

\subsection{Model Size and Commonsensicality}
\label{si:sec:model_commonsensicality:effect_of_model_size}
To investigate the relationship between a model's size and its commonsensicality score, we select from \Cref{si:tab:model_list} (on page \pageref{si:tab:model_list}) 6 model families: Falcon, Flan-T5, Gemma, LLaMA-2/3, Mistral and Qwen2. Our selection criterion is that the family must include at least 2 models for which we know the sizes. See \Cref{si:tab:model_individual_commonsensicality}. We also exclude Gemini 1.0 Pro in the Gemma/Gemini family and Mistral-Large in the Mistral family because these models are proprietary with unknown sizes. For GPT, although we know that GPT-3.5 has 175 billion parameters, we do not know this information for GPT-4 and GPT-5.

These 6 families give us 23 models in total. Since model commonsensicality varies substantially across families, we perform a mixed-effects regression analysis to analyze the relationship between model size and commonsensicality. In R's syntax, the model is described by the following formula:
\begin{align}
    \texttt{commonsensicality} ~\sim~ \log_{10}(\texttt{model size}) + (1 \mid \texttt{model family}).
    \label{si:eq:model_commonsensicality_vs_size_mixed_eff}
\end{align}
We also log-transform model size since these models can be orders of magnitude different from each other. In addition, we establish the 95\% confidence interval of the effect of (log) model size using the \texttt{confint} function in R.

\subsection{Model General Performance on LMArena and Commonsensicality}
\label{si:sec:model_commonsensicality:effect_of_model_elo}

There are numerous general benchmarks of LLMs available publicly. We choose the LMArena benchmark~\citepSI{chiangChatbotArenaOpen2024SI} to represent a model's general performance, relative to other models.

LMArena compares 230 models side by side, which include 24 models in our collection. When a user visits its website,\footnote{LMArena can be accessed at \url{https://lmarena.ai}. Previously, it was called LMSYS Arena and could be accessed at \url{https://chat.lmsys.org}.} two models are randomly chosen. The user then interacts with these models simultaneously, asking the same question and getting corresponding answers from both. Then the user decides whether they prefer the left or the right model. (They also can choose to tie them, or to indicate both models are bad.)

Models in this collection are then assigned an Elo score, with higher values indicating a higher chance of being preferred by humans. Essentially, the Elo scores are used to approximate the win-rate matrix, and these scores are estimated by Chiang \etal~\citepSI{chiangChatbotArenaOpen2024SI} using the Bradley-Terry maximum likelihood method.

The Elo score by LMArena can be interpreted as follows. Suppose two models, $a$ and $b$, receive their respective Elo scores of $R_a$ and $R_b$. Then the probability that model $a$ is chosen over model $b$ is
\begin{align}
    \frac{1}{1 + 10^{\frac{R_b - R_a}{400}}}.
    \label{si:eq:elo_interp}
\end{align}
As long as $R_a < R_b$, the denominator is greater than 2, the fraction is less than 0.5, and thus model $a$ is preferred less frequently than model $b$. For example, GPT-3.5 (model $a$) and GPT-4-0125 (model $b$) are assigned Elo scores of $R_a = 1,224$ and $R_b = 1,314$, respectively. The probability that GPT-3.5 ($a$) is preferred over GPT-4-0125 ($b$) is
\begin{align*}
    \frac{1}{1 + 10^{\frac{R_b - R_a}{400}}} = \frac{1}{1 + 10^{1,314 - 1,224}} \approx 0.37.
\end{align*}
In other words, GPT-4o is expected to win about $100\% - 37\% = 63\%$ of the time. Elo scores are invariant to translation, \ie, if all scores shift up or down by the same amount, the interpretation above does not change. In this paper, we keep the original scores that are reported on LMArena.

Since this calculation is updated over time---as more models are added to the benchmark and more users participate in evaluation---we report the scores that are the most updated at the time of writing, August 16, 2025. Models' Elo scores are calculated based on 3,866,847 pairwise comparisons.

Finally, for the 24 models in our collection that also exist on LMArena (see \Cref{si:tab:model_individual_commonsensicality}), we also extract their commonsensicality scores and evaluate the Pearson correlation between the two types of scores.

\begin{table}[p!]
    \small
    \centering
    \begin{tabular}{lcccccccc}
    \toprule
    \textbf{Model} & {\shortstack[l]{\textbf{Consensus}}} & \textbf{Awareness} & \textbf{\shortstack[l]{\textbf{Commonsensicality}}} & \textbf{Size} & \textbf{LMArena} & \textbf{OpenBookQA} \\
    \midrule
    Claude 3 Haiku & 58.8 & 64.1 & 61.4 & -- & 1,262 & 85.4 &  \\
    Claude 3 Sonnet & 60.9 & 62.2 & 61.5 & -- & 1,282 & 91.8 & \\
    Claude 3 Opus & 73.4 & 77.4 & 75.4 & -- & 1,322 & 95.6 & \\
    \midrule
    DBRX & 73.7 & 79.0 & 76.3 & 132 & 1,198 & 91.0 & \\
    \midrule
    Falcon-7B & 66.6 & 66.1 & 66.3 & 7 & -- & 26.0 & \\
    Falcon-40B & 73.0 & 77.2 & 75.1 & 40  & -- & 66.2 & \\
    Falcon-180B & 78.6 & 81.3 & 79.9 & 180 & 1,148 & -- \\
    \midrule
    Flan-T5-Small & 34.4 & 33.9 & 34.2 & .08 & -- & -- &  \\
    Flan-T5-Base & 56.8 & 59.5 & 58.1 & .25 & --  & -- &  \\
    Flan-T5-Large & 77.3 & 76.5 & 76.9 & .78 & --  & -- &  \\
    Flan-T5-XL & 73.3 & 72.7 & 73.0 & 3 & -- & -- &  \\
    Flan-T5-XXL & 79.9 & 80.9 & 80.4 & 11 & -- & -- &  \\
    \midrule
    Gemma-2B & 65.2 & 66.6 & 65.9 & 2 & 1,093 & -- &  \\
    Gemma-7B & 73.2 & 70.9 & 72.0 & 7 & 1,136 & 80.8 & \\
    Gemini Pro 1.0 & 78.4 & 81.1 & 79.7 & -- & 1,221 & 78.8 \\
    \midrule
    GPT-3.5 & 78.3 & 75.4 & 76.8 & 175 & 1,224 & 83.8 & \\
    GPT-4-0125 & 77.6 & 79.2 & 78.4 & -- & 1,314 & -- &  \\
    GPT-4-0409 & 78.0 & 83.3 & 80.6 & -- & 1,324 & 97.0 & \\
    GPT-4o & 72.5 & 77.9 & 75.2 & -- & 1,441 & 96.6 & \\
    GPT-5  & 71.9 & 79.6 & 75.7 & -- & 1,461 & -- & \\
    \midrule
    LLaMA-2-7B & 74.0 & 76.0 & 75.0 & 7 & 1,113 & 54.4 & \\
    LLaMA-2-13B & 48.5 & 44.5 & 46.5 & 13 & 1,147 & 63.4 & \\
    LLaMA-2-70B & 65.7 & 61.4 & 63.5 & 70 & 1,175 & 83.8 & \\
    LLaMA-3-8B & 57.2 & 66.5 & 61.7 & 8 & 1,225 & 76.6 & \\
    LLaMA-3-70B & 72.0 & 76.8 & 74.4 & 70 & 1,276 & 93.4 & \\
    \midrule
    Mistral-7B & 80.2 & 80.7 & 80.4 & 7 & 1,155 & 77.6 & \\
    Mixtral-8x7B & 77.8 & 75.0 & 76.4 & 47 & 1,201 & 86.8 & \\
    Mixtral-8x22B & 80.7 & 84.0 & 82.3 & 141 & 1,232 & 88.2 & \\
    Mistral-Large & 80.4 & 82.2 & 81.3 & -- & 1,244 & 89.4 & \\
    \midrule
    OLMo-7B & 74.3 & 71.0 & 72.7 & 7 & 1,080 & 22.2 & \\
    \midrule
    Qwen2-0.5B & 67.1 & 66.5 & 66.8 & .5 & -- & -- & \\
    Qwen2-1.5B & 75.4 & 73.8 & 74.6 & 1.5 & -- & -- & \\
    Qwen2-7B & 79.7 & 81.1 & 80.4 & 7 & -- & -- & \\
    Qwen2-57B & 80.4 & 81.4 & 80.9 & 57 & -- & -- & \\
    Qwen2-72B & 80.5 & 81.8 & 81.1 & 72 & 1,264 & 95.4 & \\
    \bottomrule
    \end{tabular}
    \caption[]{Individual-level consensus ($C_m$, \Cref{si:eq:model_consensus}), awareness ($A_m$, \Cref{si:eq:model_awareness}) and commonsensicality ($M_m$, \Cref{si:eq:model_commonsensicality}) scores for all models. This table also presents the sizes of these models (in billions of parameters, also shown in \Cref{si:tab:model_list}) as well as their Elo scores (updated on August 16, 2025) on the LMArena benchmark~\citepSI{chiangChatbotArenaOpen2024SI} and their accuracy on the OpenBookQA benchmark~\citepSI{mihaylovCanSuitArmor2018SI}.}
    \label{si:tab:model_individual_commonsensicality}
\end{table}

\subsection{Model Performance on Static Commonsense Benchmark and Commonsensicality}
\label{si:sec:model_commonsensicality:openbookqa}

In addition to the LMArena benchmark, we compare an LLM's commonsensicality with its performance on the OpenBookQA benchmark~\citepSI{mihaylovCanSuitArmor2018SI}. OpenBookQA is a set of multiple-choice questions generated based on 1,326 facts: Amazon Mechanical Turk workers were given a fact, such as ``poison causes harm to living things,'' and then asked to author a question using that fact. There are 5,957 questions in this dataset, which are split into three datasets: training (4,957 questions), development (500), and test (500). Below are some examples of these questions. 

\vspace{1em}

\noindent \texttt{Example question: Poison causes harm to which of the following?\\
A. A tree.\\
B. A robot.\\
C. A house.\\
D. A car.\\
(Fact: Poison causes harm to living things.)
}

\vspace{1em}

\noindent \texttt{Example question: A magnet would attract a\\
A. wooden hairbrush.\\
B. Oranges.\\
C. Glassware.\\
D. Zippers.\\
(Fact: A magnet attracts ferromagnetic metals through magnetism.)
}

\vspace{1em}

\noindent \texttt{Example question: As a car approaches you in the night,\\
A. the headlights remain at a constant.\\
B. the headlights turn off.\\
C. the headlights become more intense.\\
D. the headlights recede into the dark.\\
(Fact: As a source of light becomes closer, that source will appear brighter.)
}

\vspace{1em}

Here we only use the test set with 500 questions. The human performance on this subset is reported to be 91.7\%. For a standardized evaluation setting, we refer to the Holistic Evaluation of Language Models (HELM) framework \citepSI[version 1.13.0, released January 10, 2025]{liangHolisticEvaluationLanguage2023SI}, and find that 22 out of 35 models that we test in this paper are present on this framework. For each question in the test set, HELM prompts every LLM in a 5-shot fashion, \ie, the model is given 5 questions with their corresponding correct answers first, and then the focal test question at the end without the answer. The job of the model is to output a letter, \texttt{A}, \texttt{B}, \texttt{C}, or \texttt{D}, for the focal question. Below is an example.

\vspace{1em}

\noindent {\tt The following are multiple choice questions (with answers) about common sense.\\
\\
Question: Your body goes into starvation mode when you have insufficient amounts of\\
A. shoes\\
B. fun\\
C. pants\\
D. sustenance\\
Answer: D\\
\\
Question: What cells can perform photosynthesis?\\
A. animals\\
B. inorganic minerals\\
C. flora\\
D. critters\\
Answer: C\\
\\
Question: What does the digestive system break down into simple substances?\\
A. metals\\
B. stones\\
C. plastic food\\
D. nutriment\\
Answer: D\\
\\
Question: evaporation is the first stage in the what cycle\\
A. H2O\\
B. lunar\\
C. growth\\
D. menstrual\\
Answer: A\\
\\
Question: A fire started in a forest but it wasn't started by people. What could have been the cause?\\
A. a careless bird\\
B. a smoking bear\\
C. electricity\\
D. a campfire\\
Answer: C\\
\\
Question: A person wants to start saving money so that they can afford a nice vacation at the end of the year. After looking over their budget and expenses, they decide the best way to save money is to\\
A. make more phone calls\\
B. quit eating lunch out\\
C. buy less with monopoly money\\
D. have lunch with friends\\
Answer:\\
}

\begin{figure}[t!]
    \centering
    \includegraphics[width=0.5\textwidth]{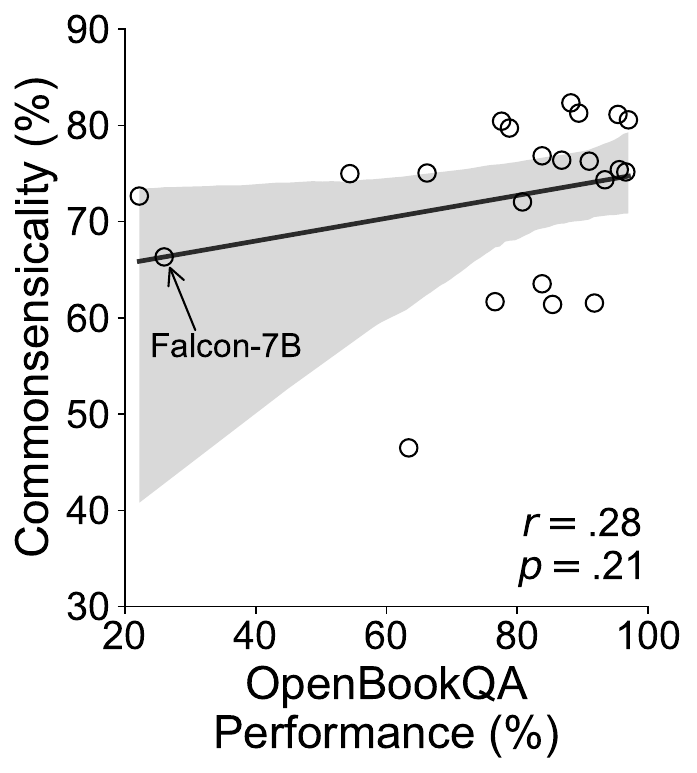}
    \caption{Correlation between an LLM's performance on the OpenBookQA benchmark and its commonsensicality. A total of 22 models are found on the HELM framework. Pearson correlation and the two-sided \textit{p}-value are reported. Also shown is the best-fit line and its 95\% confidence interval.}
    \label{si:fig:model_openbookqa_vs_commonsensicality}
\end{figure}

HELM then determines that the model is correct if and only if it outputs the letter \texttt{B} in its answer. A model's accuracy is the percentage of times it is correct across all 500 test questions. This performance is reported in \Cref{si:tab:model_individual_commonsensicality}.

Finally, \Cref{si:fig:model_openbookqa_vs_commonsensicality} plots the relationship between a model's accuracy on OpenBookQA and its commonsensicality for the 22 models found on HELM. The Pearson correlation and its \textit{p}-value are reported, as well as the best-fit line and its 95\% confidence interval.

\begin{table}[p!]
    
    \centering
    \small
    \begin{tabular}{l@{\hspace{4em}}ll@{\hspace{3em}}ll@{\hspace{3em}}ll}
    \toprule
    & \multicolumn{2}{l}{\hspace{1.5em}\textbf{Consensus}} & \multicolumn{2}{l}{\hspace{1.5em}\textbf{Awareness}} & \multicolumn{2}{l}{\hspace{-0.5em}\textbf{Commonsensicality}} \\
    \cmidrule(lr){2-7}
    & \multicolumn{1}{l}{\hspace{0.8em}(I)} & \multicolumn{1}{l}{\hspace{1.4em}(II)} & \multicolumn{1}{l}{\hspace{0.8em}(I)} & \multicolumn{1}{l}{\hspace{1.4em}(II)} & \multicolumn{1}{l}{\hspace{0.8em}(I)} & \multicolumn{1}{l}{\hspace{1.4em}(II)} \\
    \midrule
    Claude 3 Haiku & 58.8 {\footnotesize \textcolor{gray}{(31)}} & 60.5 {\footnotesize \textcolor{gray}{(31)}} & 64.1 {\footnotesize \textcolor{gray}{(30)}} & 65.5 {\footnotesize \textcolor{gray}{(30)}} & 61.4 {\footnotesize \textcolor{gray}{(32)}} & 63.0 {\footnotesize \textcolor{gray}{(32)}} \\
    Claude 3 Sonnet & 60.9 {\footnotesize \textcolor{gray}{(30)}} & 62.6 {\footnotesize \textcolor{gray}{(30)}} & 62.2 {\footnotesize \textcolor{gray}{(31)}} & 63.7 {\footnotesize \textcolor{gray}{(31)}} & 61.5 {\footnotesize \textcolor{gray}{(31)}} & 63.1 {\footnotesize \textcolor{gray}{(31)}} \\
    Claude 3 Opus & 73.4 {\footnotesize \textcolor{gray}{(19)}} & 75.1 {\footnotesize \textcolor{gray}{(20)}} & 77.4 {\footnotesize \textcolor{gray}{(15)}} & 78.2 {\footnotesize \textcolor{gray}{(15)}} & 75.4 {\footnotesize \textcolor{gray}{(17)}} & 76.6 {\footnotesize \textcolor{gray}{(17)}} \\ \midrule
    DBRX & 73.7 {\footnotesize \textcolor{gray}{(18)}} & 75.2 {\footnotesize \textcolor{gray}{(18)}} & 79.0 {\footnotesize \textcolor{gray}{(13)}} & 79.9 {\footnotesize \textcolor{gray}{(13)}} & 76.3 {\footnotesize \textcolor{gray}{(15)}} & 77.5 {\footnotesize \textcolor{gray}{(14)}} \\ \midrule
    Falcon-7B & 66.6 {\footnotesize \textcolor{gray}{(27)}} & 68.1 {\footnotesize \textcolor{gray}{(27)}} & 66.1 {\footnotesize \textcolor{gray}{(29)}} & 67.7 {\footnotesize \textcolor{gray}{(28)}} & 66.3 {\footnotesize \textcolor{gray}{(27)}} & 67.9 {\footnotesize \textcolor{gray}{(27)}} \\
    Falcon-40B & 73.0 {\footnotesize \textcolor{gray}{(22)}} & 74.8 {\footnotesize \textcolor{gray}{(22)}} & 77.2 {\footnotesize \textcolor{gray}{(16)}} & 77.9 {\footnotesize \textcolor{gray}{(17)}} & 75.1 {\footnotesize \textcolor{gray}{(19)}} & 76.3 {\footnotesize \textcolor{gray}{(20)}} \\
    Falcon-180B & 78.6 {\footnotesize \textcolor{gray}{(8)}} & 80.3 {\footnotesize \textcolor{gray}{(8)}} & 81.3 {\footnotesize \textcolor{gray}{(6)}} & 82.5 {\footnotesize \textcolor{gray}{(5)}} & 79.9 {\footnotesize \textcolor{gray}{(9)}} & 81.4 {\footnotesize \textcolor{gray}{(9)}} \\ \midrule
    Flan-T5-Small & 34.4 {\footnotesize \textcolor{gray}{(35)}} & 36.1 {\footnotesize \textcolor{gray}{(35)}} & 33.9 {\footnotesize \textcolor{gray}{(35)}} & 35.6 {\footnotesize \textcolor{gray}{(35)}} & 34.2 {\footnotesize \textcolor{gray}{(35)}} & 35.8 {\footnotesize \textcolor{gray}{(35)}} \\
    Flan-T5-Base & 56.8 {\footnotesize \textcolor{gray}{(33)}} & 58.6 {\footnotesize \textcolor{gray}{(33)}} & 59.5 {\footnotesize \textcolor{gray}{(33)}} & 60.6 {\footnotesize \textcolor{gray}{(33)}} & 58.1 {\footnotesize \textcolor{gray}{(33)}} & 59.6 {\footnotesize \textcolor{gray}{(33)}} \\
    Flan-T5-Large & 77.3 {\footnotesize \textcolor{gray}{(14)}} & 78.9 {\footnotesize \textcolor{gray}{(14)}} & 76.5 {\footnotesize \textcolor{gray}{(18)}} & 77.9 {\footnotesize \textcolor{gray}{(16)}} & 76.9 {\footnotesize \textcolor{gray}{(12)}} & 78.4 {\footnotesize \textcolor{gray}{(12)}} \\
    Flan-T5-XL & 73.3 {\footnotesize \textcolor{gray}{(20)}} & 75.0 {\footnotesize \textcolor{gray}{(21)}} & 72.7 {\footnotesize \textcolor{gray}{(23)}} & 74.1 {\footnotesize \textcolor{gray}{(23)}} & 73.0 {\footnotesize \textcolor{gray}{(23)}} & 74.6 {\footnotesize \textcolor{gray}{(23)}} \\
    Flan-T5-XXL & 79.9 {\footnotesize \textcolor{gray}{(6)}} & 81.6 {\footnotesize \textcolor{gray}{(6)}} & 80.9 {\footnotesize \textcolor{gray}{(9)}} & 82.1 {\footnotesize \textcolor{gray}{(8)}} & 80.4 {\footnotesize \textcolor{gray}{(7)}} & 81.8 {\footnotesize \textcolor{gray}{(6)}} \\ \midrule
    Gemma-2B & 65.2 {\footnotesize \textcolor{gray}{(29)}} & 66.8 {\footnotesize \textcolor{gray}{(29)}} & 66.6 {\footnotesize \textcolor{gray}{(26)}} & 67.9 {\footnotesize \textcolor{gray}{(27)}} & 65.9 {\footnotesize \textcolor{gray}{(28)}} & 67.3 {\footnotesize \textcolor{gray}{(28)}} \\
    Gemma-7B & 73.2 {\footnotesize \textcolor{gray}{(21)}} & 75.1 {\footnotesize \textcolor{gray}{(19)}} & 70.9 {\footnotesize \textcolor{gray}{(25)}} & 72.3 {\footnotesize \textcolor{gray}{(25)}} & 72.0 {\footnotesize \textcolor{gray}{(25)}} & 73.7 {\footnotesize \textcolor{gray}{(25)}} \\
    Gemini Pro 1.0 & 78.4 {\footnotesize \textcolor{gray}{(9)}} & 80.0 {\footnotesize \textcolor{gray}{(9)}} & 81.1 {\footnotesize \textcolor{gray}{(7)}} & 82.2 {\footnotesize \textcolor{gray}{(7)}} & 79.7 {\footnotesize \textcolor{gray}{(10)}} & 81.1 {\footnotesize \textcolor{gray}{(10)}} \\ \midrule
    GPT-3.5 & 78.3 {\footnotesize \textcolor{gray}{(10)}} & 80.0 {\footnotesize \textcolor{gray}{(10)}} & 75.4 {\footnotesize \textcolor{gray}{(20)}} & 76.3 {\footnotesize \textcolor{gray}{(20)}} & 76.8 {\footnotesize \textcolor{gray}{(13)}} & 78.1 {\footnotesize \textcolor{gray}{(13)}} \\
    GPT-4-0125 & 77.6 {\footnotesize \textcolor{gray}{(13)}} & 79.1 {\footnotesize \textcolor{gray}{(13)}} & 79.2 {\footnotesize \textcolor{gray}{(12)}} & 80.2 {\footnotesize \textcolor{gray}{(12)}} & 78.4 {\footnotesize \textcolor{gray}{(11)}} & 79.7 {\footnotesize \textcolor{gray}{(11)}} \\
    GPT-4-0409 & 78.0 {\footnotesize \textcolor{gray}{(11)}} & 79.4 {\footnotesize \textcolor{gray}{(11)}} & 83.3 {\footnotesize \textcolor{gray}{(2)}} & 84.2 {\footnotesize \textcolor{gray}{(2)}} & 80.6 {\footnotesize \textcolor{gray}{(5)}} & 81.8 {\footnotesize \textcolor{gray}{(7)}} \\
    GPT-4o & 72.5 {\footnotesize \textcolor{gray}{(23)}} & 74.2 {\footnotesize \textcolor{gray}{(23)}} & 77.9 {\footnotesize \textcolor{gray}{(14)}} & 78.8 {\footnotesize \textcolor{gray}{(14)}} & 75.2 {\footnotesize \textcolor{gray}{(18)}} & 76.4 {\footnotesize \textcolor{gray}{(18)}} \\
    GPT-5 & 71.9 {\footnotesize \textcolor{gray}{(25)}} & 73.5 {\footnotesize \textcolor{gray}{(25)}} & 79.6 {\footnotesize \textcolor{gray}{(11)}} & 80.6 {\footnotesize \textcolor{gray}{(11)}} & 75.7 {\footnotesize \textcolor{gray}{(16)}} & 77.0 {\footnotesize \textcolor{gray}{(16)}} \\ \midrule
    LLaMA-2-7B & 74.0 {\footnotesize \textcolor{gray}{(17)}} & 75.6 {\footnotesize \textcolor{gray}{(17)}} & 76.0 {\footnotesize \textcolor{gray}{(19)}} & 77.2 {\footnotesize \textcolor{gray}{(19)}} & 75.0 {\footnotesize \textcolor{gray}{(20)}} & 76.4 {\footnotesize \textcolor{gray}{(19)}} \\
    LLaMA-3-8B & 57.2 {\footnotesize \textcolor{gray}{(32)}} & 58.9 {\footnotesize \textcolor{gray}{(32)}} & 66.5 {\footnotesize \textcolor{gray}{(27)}} & 67.7 {\footnotesize \textcolor{gray}{(28)}} & 61.7 {\footnotesize \textcolor{gray}{(30)}} & 63.2 {\footnotesize \textcolor{gray}{(30)}} \\
    LLaMA-2-13B & 48.5 {\footnotesize \textcolor{gray}{(34)}} & 50.3 {\footnotesize \textcolor{gray}{(34)}} & 44.5 {\footnotesize \textcolor{gray}{(34)}} & 46.1 {\footnotesize \textcolor{gray}{(34)}} & 46.5 {\footnotesize \textcolor{gray}{(34)}} & 48.1 {\footnotesize \textcolor{gray}{(34)}} \\
    LLaMA-2-70B & 65.7 {\footnotesize \textcolor{gray}{(28)}} & 67.4 {\footnotesize \textcolor{gray}{(28)}} & 61.4 {\footnotesize \textcolor{gray}{(32)}} & 62.4 {\footnotesize \textcolor{gray}{(32)}} & 63.5 {\footnotesize \textcolor{gray}{(29)}} & 64.9 {\footnotesize \textcolor{gray}{(29)}} \\
    LLaMA-3-70B & 72.0 {\footnotesize \textcolor{gray}{(24)}} & 73.6 {\footnotesize \textcolor{gray}{(24)}} & 76.8 {\footnotesize \textcolor{gray}{(17)}} & 77.8 {\footnotesize \textcolor{gray}{(18)}} & 74.4 {\footnotesize \textcolor{gray}{(22)}} & 75.7 {\footnotesize \textcolor{gray}{(22)}} \\ \midrule
    Mistral-7B & 80.2 {\footnotesize \textcolor{gray}{(5)}} & 81.7 {\footnotesize \textcolor{gray}{(5)}} & 80.7 {\footnotesize \textcolor{gray}{(10)}} & 82.1 {\footnotesize \textcolor{gray}{(9)}} & 80.4 {\footnotesize \textcolor{gray}{(6)}} & 81.9 {\footnotesize \textcolor{gray}{(5)}} \\
    Mixtral-8x7B & 77.8 {\footnotesize \textcolor{gray}{(12)}} & 79.4 {\footnotesize \textcolor{gray}{(12)}} & 75.0 {\footnotesize \textcolor{gray}{(21)}} & 75.7 {\footnotesize \textcolor{gray}{(21)}} & 76.4 {\footnotesize \textcolor{gray}{(14)}} & 77.5 {\footnotesize \textcolor{gray}{(15)}} \\
    Mixtral-8x22B & 80.7 {\footnotesize \textcolor{gray}{(1)}} & 82.2 {\footnotesize \textcolor{gray}{(1)}} & 84.0 {\footnotesize \textcolor{gray}{(1)}} & 84.9 {\footnotesize \textcolor{gray}{(1)}} & 82.3 {\footnotesize \textcolor{gray}{(1)}} & 83.6 {\footnotesize \textcolor{gray}{(1)}} \\
    Mistral-Large & 80.4 {\footnotesize \textcolor{gray}{(4)}} & 82.0 {\footnotesize \textcolor{gray}{(4)}} & 82.2 {\footnotesize \textcolor{gray}{(3)}} & 83.4 {\footnotesize \textcolor{gray}{(3)}} & 81.3 {\footnotesize \textcolor{gray}{(2)}} & 82.7 {\footnotesize \textcolor{gray}{(2)}} \\ \midrule
    OLMo-7B & 74.3 {\footnotesize \textcolor{gray}{(16)}} & 75.9 {\footnotesize \textcolor{gray}{(16)}} & 71.0 {\footnotesize \textcolor{gray}{(24)}} & 72.3 {\footnotesize \textcolor{gray}{(24)}} & 72.7 {\footnotesize \textcolor{gray}{(24)}} & 74.1 {\footnotesize \textcolor{gray}{(24)}} \\ \midrule
    Qwen2-0.5B & 67.1 {\footnotesize \textcolor{gray}{(26)}} & 68.7 {\footnotesize \textcolor{gray}{(26)}} & 66.5 {\footnotesize \textcolor{gray}{(27)}} & 67.9 {\footnotesize \textcolor{gray}{(26)}} & 66.8 {\footnotesize \textcolor{gray}{(26)}} & 68.3 {\footnotesize \textcolor{gray}{(26)}} \\
    Qwen2-1.5B & 75.4 {\footnotesize \textcolor{gray}{(15)}} & 76.8 {\footnotesize \textcolor{gray}{(15)}} & 73.8 {\footnotesize \textcolor{gray}{(22)}} & 74.9 {\footnotesize \textcolor{gray}{(22)}} & 74.6 {\footnotesize \textcolor{gray}{(21)}} & 75.8 {\footnotesize \textcolor{gray}{(21)}} \\
    Qwen2-7B & 79.7 {\footnotesize \textcolor{gray}{(7)}} & 81.2 {\footnotesize \textcolor{gray}{(7)}} & 81.1 {\footnotesize \textcolor{gray}{(8)}} & 82.0 {\footnotesize \textcolor{gray}{(10)}} & 80.4 {\footnotesize \textcolor{gray}{(8)}} & 81.6 {\footnotesize \textcolor{gray}{(8)}} \\
    Qwen2-57B & 80.4 {\footnotesize \textcolor{gray}{(3)}} & 82.1 {\footnotesize \textcolor{gray}{(2)}} & 81.4 {\footnotesize \textcolor{gray}{(5)}} & 82.4 {\footnotesize \textcolor{gray}{(6)}} & 80.9 {\footnotesize \textcolor{gray}{(4)}} & 82.3 {\footnotesize \textcolor{gray}{(4)}} \\
    Qwen2-72B & 80.5 {\footnotesize \textcolor{gray}{(2)}} & 82.1 {\footnotesize \textcolor{gray}{(3)}} & 81.8 {\footnotesize \textcolor{gray}{(4)}} & 82.8 {\footnotesize \textcolor{gray}{(4)}} & 81.1 {\footnotesize \textcolor{gray}{(3)}} & 82.5 {\footnotesize \textcolor{gray}{(3)}} \\ \midrule
    Spearman's $r$ & \multicolumn{2}{l}{\hspace{2.2em}$.9989$} & \multicolumn{2}{l}{\hspace{2.2em}$.9974$} & \multicolumn{2}{l}{\hspace{2.2em}$.9986$} \\
    \bottomrule
    \end{tabular}
    \caption{Individual-level consensus, awareness and commosensicality for LLMs under two calculation variants. In variant (I), the majority rating for each statement $i$ follows \Cref{si:eq:human_majority_rating}, based only on the \emph{human} ratings. In variant (II), the majority rating follows \Cref{si:eq:human_majority_rating_with_model}, where the determination of the majority also takes into account the model's rating. The numbers in parentheses indicate the ranking of these scores within their respective column. Spearman's ranking correlation \textit{r} is calculated by comparing variants (I) and (II) for each type of score. All correlations are significant with \textit{p} < .00001.}
    \label{si:tab:model_commosensicality_two_variants}
\end{table}

\subsection{Comparing Models and Humans Side by Side on Commonsensicality}
\label{si:sec:model_commonsensicality:pair_comparison}

When comparing a model and a human directly on their commonsensicality scores, note that each person was only tasked with labeling 50 randomly chosen statements, whereas a model labels all 4,407 of them. For a fair comparison, we restrict the calculation of consensus, awareness and commonsensicality of the model only to the 50 statements that the person labeled.

Specifically, for person $j$ the calculations in \Cref{si:eq:human_consensus,si:eq:human_awareness,si:eq:human_commonsensicality} (\pageref{si:eq:human_consensus}) do not change because they are already with respect to $\Phi_j$, the subset of statements that were labeled by $j$. To compare model $m$ with person $j$, we modify \Cref{si:eq:model_consensus,si:eq:model_awareness} above so that the average is with respect to $\Phi_j$:
\begin{align*}
    C_m &= \frac{1}{|\Phi_j|} \sum_{i \in \Phi_j} \mathbbm{1} \left[ \alpha_i^m = \texttt{majority}_i^h \right], \\
    A_m &= \frac{1}{|\Phi_j|} \sum_{i \in \Phi_j} \mathbbm{1} \left[ \beta_i^m = \texttt{majority}_i^h \right].
\end{align*}

\subsection{Alternative Calculation of Model Commonsensicality with Model-Contributed Majority Vote}
\label{si:sec:model_commonsensicality:alternative_calculation}

In calculating the model commonsensicality, we rely on the majority rating of each statement; see \Cref{si:eq:human_majority_rating} (page \pageref{si:eq:human_majority_rating}). When each LLM is treated as a survey participant, its rating can also contribute to this majority vote. Here, we explore this alternative calculation to see whether model commonsensicality can be sensitive to it.

Formally, for each model $m$, define the human-plus-model majority rating for statement $i$ as
\begin{align}
    \texttt{majority}_i^h = \mathbbm{1} \left[ \frac{1}{|\Omega_i| + 1} \left( \alpha_i^m + \sum_{j \in \Omega_i} A_{i,j} \right) \geq 0.5 \right],
    \label{si:eq:human_majority_rating_with_model}
\end{align}
where $\Omega_i$ is the set of humans that were assigned to rate statement $i$; $\alpha_i^m$ is the model's binary rating for the statement in response to question (a) from \Cref{si:eq:model_binary_rating_q1} (page \pageref{si:eq:model_binary_rating_q1}); and $A$ is the matrix of human ratings for all statements from \Cref{si:eq:human_ratings_q1} (page \pageref{si:eq:human_ratings_q1}).

\Cref{si:tab:model_commosensicality_two_variants} on page \pageref{si:tab:model_commosensicality_two_variants} compares the consensus, awareness and commonsensicality scores for each model with respect to the two calculation variants. While their raw scores change, the models' relative rankings do not significantly change, as evidenced by the Spearman correlation coefficients at the bottom of the table.

\clearpage
\newpage
\section{Measuring Commonsensicality when LLMs are Simulators of Silicon Samples}
\label{si:sec:model_collective}

In this section, we use each language model not as an individual survey participant (\cf~\Cref{si:sec:model_commonsensicality}), but rather as a generator of ``silicon samples'' which, together, form a hypothetical population of raters. Common sense in this case is measured with respect to this entire population, rather than for each silicon sample within.

To make this clearer, imagine a population of silicon samples, each represented by the same language model. When we ask each respondent in this population questions (a) and (b), we record their binary answers (``yes'' or ``no''). Each respondent is treated as independent from the others, and so collecting answers for this entire population amounts to resampling the answers to the same questions. In the limit---\ie, if the size of the population goes to infinity---the proportion of respondents that agree with statement $i$ is exactly the probability with which the language model agrees with this statement, $d_i^{m, a}$ (defined in \Cref{si:eq:model_ratings_dist} on page \pageref{si:eq:model_ratings_dist}). Similarly, the proportion of silicon respondents that believe most people would agree with this statement is exactly $d_i^{m, b}$ (\Cref{si:eq:model_ratings_dist_q2} on page \pageref{si:eq:model_ratings_dist_q2}).

Mimicking the calculations in \Cref{si:sec:commonsense_recap:statement_commonsensicality}, we first define the majority rating within this population of silicon samples for statement $i$ as
\begin{align}
    \texttt{majority}_i^m = \mathbbm{1}[d_i^{m, a} \geq 0.5].
    \label{si:eq:silicon_majority_rating}
\end{align}

The consensus score of statement $i$ with respect to this population is
\begin{align}
    c_i^m = 2 \times |d_i^{m, a} - 0.5|.
    \label{si:eq:silicon_statement_consensus}
\end{align}
The statement's awareness score is the proportion of silicon samples that accurately predict the majority opinion via their answer to question (b):
\begin{align}
    a_i^m = \begin{cases}
      d_i^{m, b} & \text{if\quad} \texttt{majority}_i^m = 1 \\
      1 - d_i^{m, b} & \text{if\quad} \texttt{majority}_i^m = 0.
    \end{cases}
    \label{si:eq:silicon_statement_awareness}
\end{align}
Which leads to its commonsensicality score of
\begin{align}
    m_i^m = \sqrt{c_i^m \times a_i^m}.
    \label{si:eq:silicon_statement_commonsensicality}. 
\end{align}

The objective is to compare $m_i^m$ with $m_i^h$ (\Cref{si:eq:statement_commonsensicality} on page \pageref{si:eq:statement_commonsensicality}). First, we look at the correlation between these two scores, taken over $n_s = 4,407$ statements. Specifically, we calculate the Pearson correlation between $m_i^m$ with $m_i^h$ as well as its two-sided \textit{p}-values (after Bonferroni correction).

We also make this comparison through a different lens: how well does a statement's commonsensicality score in the population of silicon samples predict its same score in the population of humans? To do so, we fit a linear regression model, predicting $m_i^h$ using $m_i^m$. We report the regression coefficient as well as its two-sided \textit{p}-values.

\subsection{Comparing Statement Commonsensicality Scores within Humans and Silicon Samples by Statement Type}
\label{si:sec:model_collective:statement_comm_by_type} 

Each statement in our corpus was labeled by MTurk workers according to six epistemological features: 
\begin{enumerate}
    \item \textit{Behavior}
    \begin{itemize}
        \item \textit{Social} (\textit{N} = 2,939): it refers to beliefs, perceptions, preferences, and socially constructed rules that govern human experience; it can be ``real'' or opinion, but is intrinsically of human origins. \eg, ``I exist and am the same person I was yesterday,'' ``He yelled at me because he was angry,'' ``There are seven days in the week.''
        \item \textit{Physical} (\textit{N} = 1,468): it refers to objective features of the world as described by, say, physics, biology, engineering, mathematics or other natural rules; it can be measured empirically, or derived logically. \eg, ``Men on average are taller than women,'' ``The Earth is the third planet from the Sun,'' ``Ants are smaller than Elephants.''
    \end{itemize}
    \item \textit{Everyday}
    \begin{itemize}
        \item \textit{Everyday} (\textit{N} = 2,779): people encounter, or could encounter, situations like this in the course of their ordinary, everyday experiences, \eg, ``Touching a hot stove will burn you,'' ``Commuting at rush hour takes longer,'' ``It is rude to jump the line.''
        \item \textit{Abstract} (\textit{N} = 1,628): this claim refers to regularities or conclusions that cannot be observed or arrived at solely through individual experience, \eg, ``Capitalism is a better economic system than Communism,'' ``Strict gun laws save lives,'' ``God exists.''
    \end{itemize}
    \item \textit{Figure of speech}
    \begin{itemize}
        \item \textit{Figure of speech} (\textit{N} = 854): it contains an aphorism, metaphor, hyperbole, \eg, ``Birds of a feather flock together,'' ``A friend to all is a friend to none.''
        \item \textit{Literal language} (\textit{N} = 3,553): it is plain and ordinary language that means exactly what it says. e.g. ``The sky is blue,'' ``Elephants are larger than dogs,'' ``Abraham Lincoln was a great president.''
    \end{itemize}
    \item \textit{Judgment}
    \begin{itemize}
        \item \textit{Normative} (\textit{N} = 3,191): it refers to a judgment, belief, value, social norm or convention. \eg, ``If you are going to the office, you should wear business attire, not a bathing suit,'' ``Treat others how you want them to treat you,'' ``Freedom is a fundamental human right.''
        \item \textit{Positive} (\textit{N} = 1,216): it refers to something in the world such as an empirical regularity or scientific law, \eg, ``Hot things will burn you,'' ``The sun rises in the east and sets in the west.''s
    \end{itemize}
    \item \textit{Opinion}
    \begin{itemize}
        \item \textit{Opinion} (\textit{N} = 2,805): it is something that someone might think is true, or wants others to think is true, but can’t be demonstrated to be objectively correct or incorrect; it is inherently subjective. \eg, ``FDR was the greatest US president of the 20th Century,'' ``The Brooklyn Bridge is prettier than the Golden Gate,'' ``Vaccine mandates are a tolerable imposition on individual freedom.''
        \item \textit{Fact} (\textit{N} = 1,602): it is something that can be demonstrated to be correct or incorrect, independently of anyone’s opinion, \eg, ``the earth is the third planet from the sun,'' (this is correct and we know it is correct), ``Obama was the 24th president of the United States'' (this is incorrect, but we know it’s incorrect) ``It will be sunny next Tuesday'' (we don’t yet know if this is correct, but we will be able to check in the future).
    \end{itemize}
    \item \textit{Reasoning}
    \begin{itemize}
        \item \textit{Knowledge} (\textit{N} = 1,672): the claim refers to some observation about the world; it may be true or false, opinion or fact, subjective or objective \eg, ``The sun rises in the east and sets in the west,'' ``Dogs are nicer than cats,'' ``Glasses break when they are dropped.''
        \item \textit{Reasoning} (\textit{N} = 2,735): the claim presents a conclusion that is arrived at by combining knowledge and logic, \eg, ``The sun is in the east, therefore it is morning,'' ``My dog is wagging its tail, therefore it is happy,'' ``The glass fell off the table, therefore it will break and the floor will become wet.''
    \end{itemize}
\end{enumerate}

\begin{figure}[p!]
    \centering
    \includegraphics[width=1\linewidth]{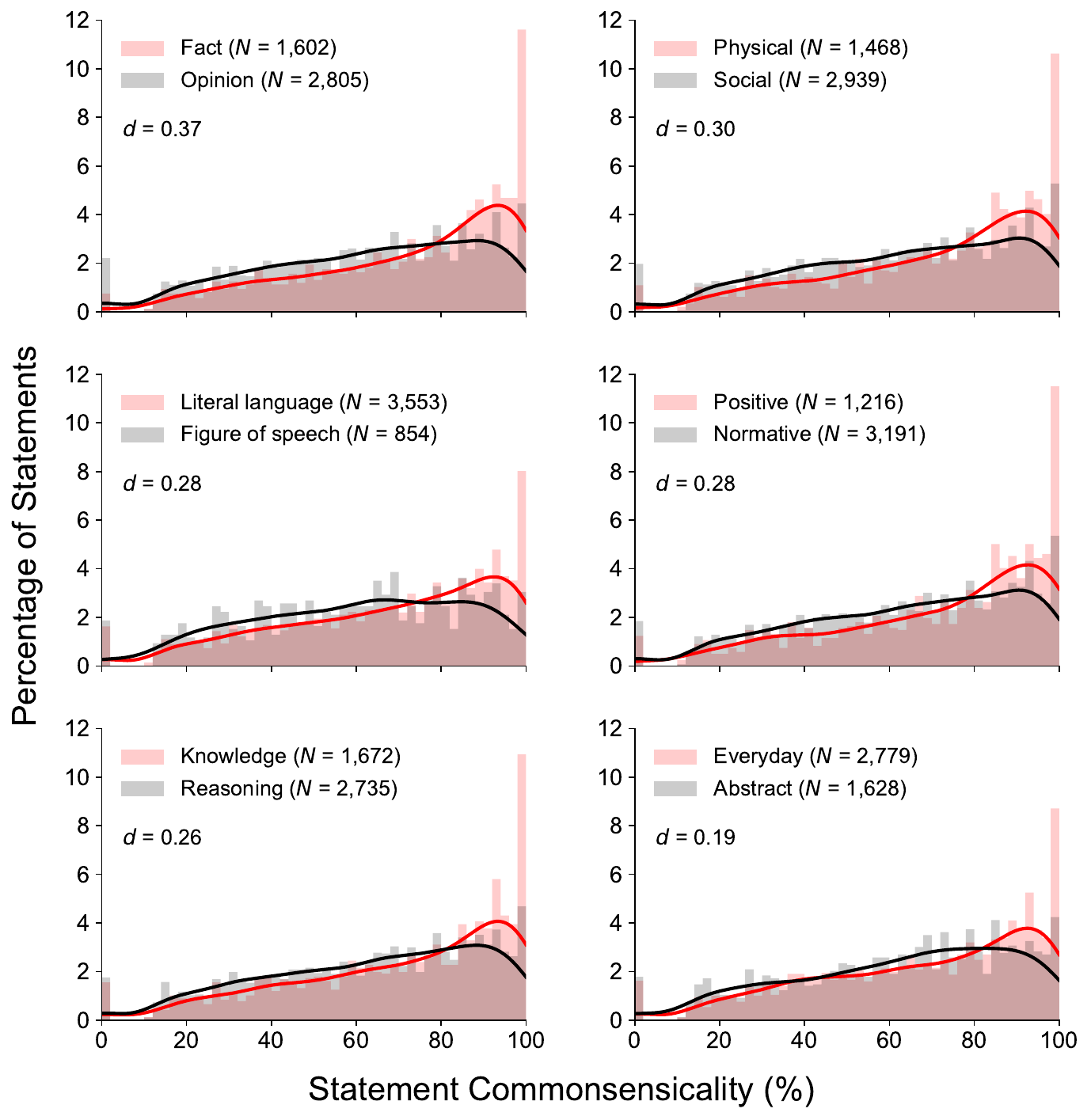}
    \caption{Comparison of statement commonsensicality with respect to each epistemological dichotomy. In each subplot, the corpus of 4,407 statements is split into two complementary subsets, such as facts and opinions. Within each subset, the distribution of statement commonsensicality is shown with a histogram and a kernel density estimation (KDE) line of the corresponding color. Also noted in each subplot is the standardized difference in means between the two (red and black) distributions.}
    \label{si:fig:statement_comm_by_feature_all}
\end{figure}

For every dichotomy, the corpus of $n_s = 4,407$ statements is divided into two subsets corresponding to the two opposing dimensions in that dichotomy. \Cref{si:fig:statement_comm_by_feature_all} compares the distributions of statement commonsensicality, computed using human ratings, with respect to each subset in every dichotomy.

Our goal is to compare the commonsensicality scores (either by humans or by models) between these two dimensions. For instance, a statement is either about a fact or an opinion, corresponding to dichotomy 5 above. Are statements depicting facts more commonsensical than those depicting opinions?

To do so, we compute the difference in the mean commonsensicality score between the two subsets of statements. This is done for every population: of humans and of silicon samples created by 35 models. To establish a confidence interval, we construct 1,000 bootstrap samples and calculate this difference for each sample.

\begin{figure}[p!]
    \centering
    \includegraphics[width=0.9\linewidth]{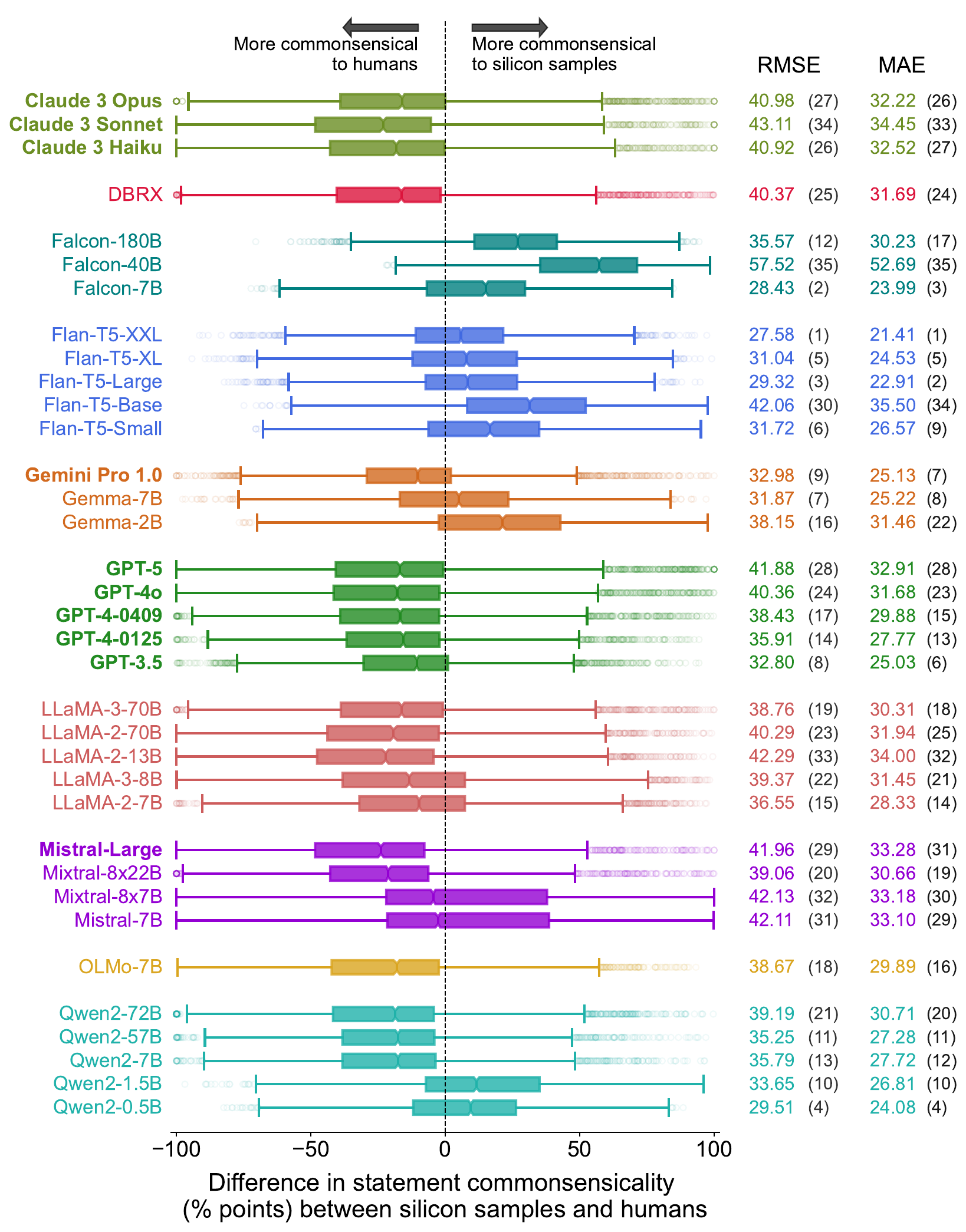}
    \caption{ Comparison of statement commonsensicality scores between humans and silicon samples generated by LLMs. The \textit{x}-axis represents the difference in a statement's scores between these two populations, where a positive difference indicates that the statement is more commonsensical to silicon samples than it is to humans (\ie, $m_i^m > m_i^h$). Each box indicates the inter-quartile range, and the median notch is contained within it. The root mean-squared error (RMSE) and mean absolute error (MAE) are calculated based on the difference between $m_i^m$ and $m_i^h$, averaged over 4,407 statements. The numbers in parentheses are the rankings of the RMSE and MAE (in ascending order); for example, Claude 3 Opus has an RMSE ranking of 27 out of 35 models, whereas its MAE ranking is 26 out of 35.}
    \label{si:fig:statement_comm_errors_with_humans} 
\end{figure}

\newpage

\section{Other Variants of Model Evaluation Settings}
\label{si:sec:other_prompts}

In this section we present additional results of model ratings in response to several variants of the default prompt we use in the main paper.

\subsection{Model's Reasoning Following Binary Answer}
\label{si:sec:other_prompts:posthoc_reasoning}

In \Cref{si:tab:model_posthoc_reasoning_q1,si:tab:model_posthoc_reasoning_q2}, we keep our original prompts and let GPT-4o continue sampling to get its full answer. The text following the first tokens---``yes'' or ''no''---typically represents the model's elaboration of its answer.

\subsection{Model's Reasoning Before Binary Answer}
\label{si:sec:other_prompts:prehoc_reasoning}

In \Cref{si:tab:model_prehoc_reasoning_q1,si:tab:model_prehoc_reasoning_q2}, we slightly modify our prompts as follows. Instead of asking GPT-4o to start its answer with a ``yes'' or ''no,'' we instruct it to explain what it thinks in a few sentences first.

\subsection{Clarifying the Role of LLMs in the Survey}
\label{si:sec:other_prompts:role_clarification}

In designing our default prompts, we intend to blind the LLMs to the purpose of data collection. Specifically, we do not disclose who we are and why we would like to solicit their ratings of the given statements.

It is possible that in responding to our prompts, the models may not do so as an independent survey participant, but rather as a useful assistant that is trying to give an answer that satisfies its user. Here, we design a new prompt setting, which we call ``role clarification,'' that emphasizes this aspect. Specifically, we use the following instruction in the system message: ``You are an independent participant in a survey administered by academic researchers, who study commonsense beliefs. You will be presented with a statement, and asked a question about that statement. Answer the question independently, and please do not take into account what you think the researchers might want you to say.''

\begin{figure}[t] 
    \centering
    \includegraphics[width=1\linewidth]{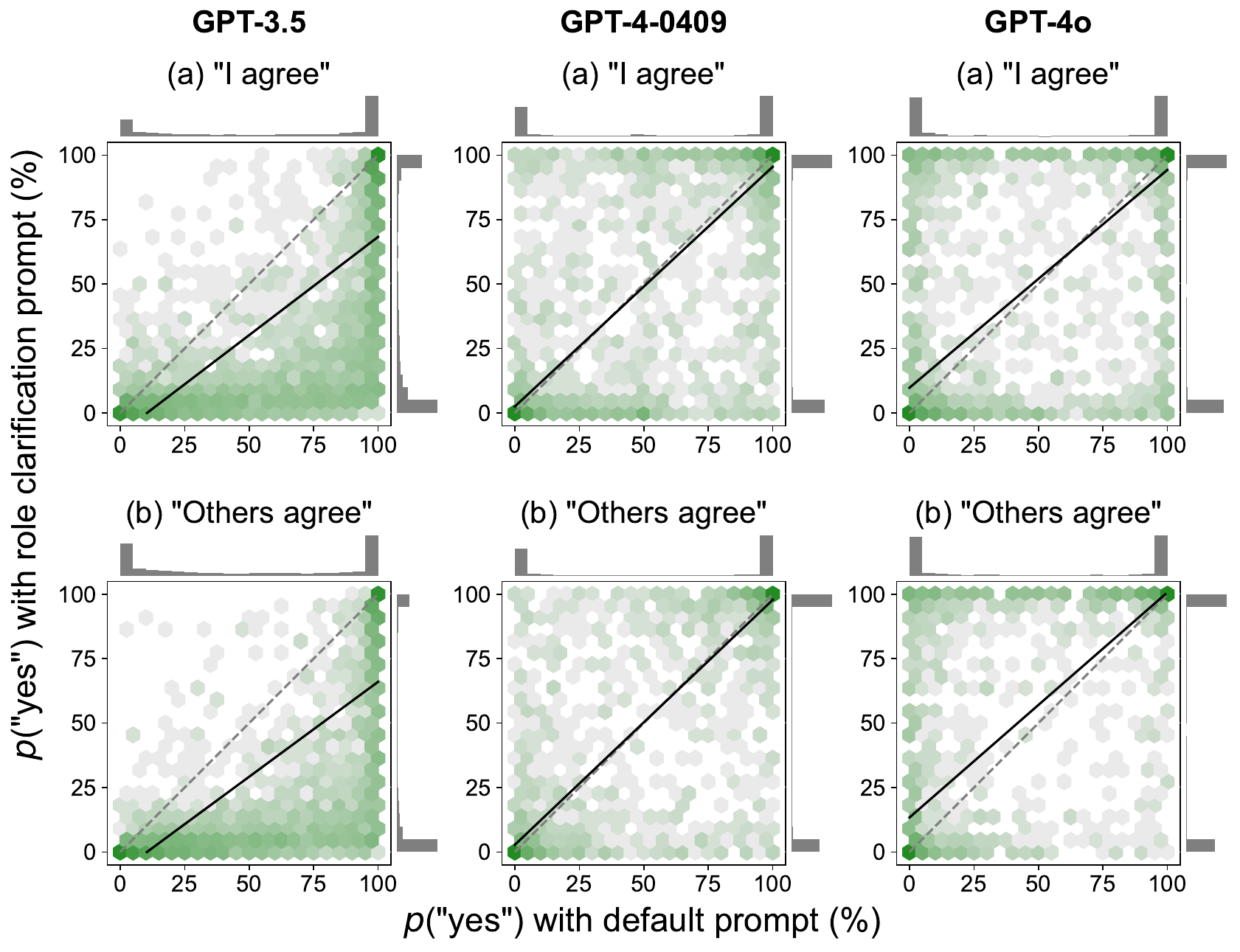}
    \caption{Calibration between a model's probabilities in two settings: without system prompt and with system prompt. The model is given a statement, and asked two questions: (a) ``Do you agree with this statement?'' and (b) ``Do you think most people would agree with this statement?'' The \textit{x}-axis is the probability of the model's answer to the default prompts. The \textit{y}-axis is the same probability but to the role clarification prompts. This is to explicitly discourage the model from answering \emph{only} to satisfy the user. Also illustrated are the marginal distributions, the perfect calibration line (dashed black) and the best-fit line (solid black).}
    \label{si:tab:role_clarification_calibration}
\end{figure}

We collect ratings for three models: GPT-3.5, GPT-4-0409 and GPT-4o. First, \Cref{si:tab:role_clarification_calibration} on page \pageref{si:tab:role_clarification_calibration} presents the calibration of the models' output probabilities for both questions (a) and (b). Second, \Cref{si:fig:role_clarification_eval_results} on page \pageref{si:fig:role_clarification_eval_results} compares the evaluation results of these models under two settings: default and role clarification. In particular, we replicate our calculation of two metrics: individual common sense (measured by the commonsensicality score) and group common sense (measured by the Pearson correlation between the statement scores calculated from silicon samples generated by a model and the same statement scores arising from human ratings). Finally, \Cref{si:tab:role_clarification_posthoc_reasoning} on page \pageref{si:tab:role_clarification_posthoc_reasoning} shows some examples of GPT-4o's elaboration after its binary ratings.

\begin{table}[p!]

\centering
\small
\begin{tabular}{p{0.47\linewidth}@{\hspace{2em}}p{0.47\linewidth}}
\toprule
\multicolumn{2}{l}{\textbf{(a) Do you agree with this statement? Start your answer with a ``yes'' or ``no''.}}                                          \\
\midrule
\multicolumn{2}{l}{\textbf{Statement: ``Starting a new job people do what is expected of them for that job.'' \vspace{0.5em}}} \\
\begin{tabular}[t]{@{}p{1\linewidth}} 
        Yes. Starting a new job typically involves meeting the expectations set by the employer. New employees often try to understand and fulfill their roles and responsibilities to make a positive impression and secure their position. They might be motivated to demonstrate their capabilities, align with the company's goals, and adhere to the prescribed norms and procedures. However, this doesn't mean that every individual will perfectly meet these expectations from the start. There can be a learning curve, and the extent to which a new employee successfully meets job expectations can vary
\end{tabular}                                                 &  
\begin{tabular}[t]{@{}p{1\linewidth}} 
        No. While many people strive to meet or exceed expectations when starting a new job, this is not universally true for everyone. Various factors can influence an individual's performance, such as their understanding of the job requirements, the level of training and support they receive, their personal motivation and work ethic, and the overall work environment. Some individuals may exceed expectations due to high motivation and enthusiasm, while others might struggle to meet basic requirements due to inadequate onboarding, lack of clarity in their role, or personal challenges.
\end{tabular}\\
\midrule
\multicolumn{2}{l}{\textbf{Statement: ``One must discard the prejudice that the truth must be something tangible.'' \vspace{0.5em}}} \\
\begin{tabular}[t]{@{}p{1\linewidth}} 
        Yes. The statement encourages us to broaden our understanding of truth beyond the confines of physical tangibility. While tangible evidence and sensory experiences are crucial in many contexts, especially in science and daily life, they do not encompass the entirety of what can be considered true. For example, mathematical truths, logical principles, and abstract concepts such as justice or love are not tangible, yet they hold significant validity and impact. Additionally, philosophical and metaphysical inquiries often deal with truths that are not accessible through sensory
\end{tabular}                                                 &  
\begin{tabular}[t]{@{}p{1\linewidth}} 
        No. The statement suggests that truth is not necessarily tied to tangible or physical entities. I tend to agree with this view. Truth can encompass abstract concepts such as love, justice, and mathematical principles, which are not physical but can be understood intellectually and emotionally. Moreover, many scientific truths describe principles that are not directly observable but are inferred through indirect evidence and logical reasoning, like the existence of subatomic particles or the concept of gravitational forces. Discarding the notion that truth must be tangible allows for
\end{tabular}\\
\midrule
\multicolumn{2}{l}{\textbf{Statement: ``People open a window when they feel hot.'' \vspace{0.5em}}} \\
\begin{tabular}[t]{@{}p{1\linewidth}} 
        Yes, I agree with the statement to an extent. Many people do open a window when they feel hot as a way to allow fresh air to circulate and help cool down the indoor environment. However, this behavior can depend on various factors, such as the availability of air conditioning, the outside temperature, air quality, and individual preferences. For example, if it is very hot or humid outside, or if the air quality is poor, people might opt to use fans or air conditioning instead of opening a
\end{tabular}                                                 &  
\begin{tabular}[t]{@{}p{1\linewidth}} 
        No, I do not entirely agree with this statement. While it is true that some people may choose to open a window when they feel hot as a way to cool down, others might opt for different methods. These alternate methods can include using air conditioning, electric fans, or cool towels. The choice depends largely on individual preferences, availability of resources, and specific circumstances, such as outdoor temperatures and air quality. Therefore, while opening a window can be a common response to feeling hot, it is not
\end{tabular}\\
\midrule
\multicolumn{2}{l}{\textbf{Statement: ``There is nothing more tragic in life than the utter impossibility of changing what you have done.'' \vspace{0.5em}}} \\
\begin{tabular}[t]{@{}p{1\linewidth}} 
        Yes. The statement highlights a poignant aspect of human existence: the permanence of past actions. This notion can indeed evoke a deep sense of tragedy, as it implies an irreversible impact of decisions that, in hindsight, one might wish to alter. The inability to change past mistakes can lead to feelings of regret and remorse, which are powerful emotions that can shape a person's future and mentality. However, it's important to consider the complexity of human experience. While the inability to change the past is unchangeable,
\end{tabular}                                                 &  
\begin{tabular}[t]{@{}p{1\linewidth}} 
        No. While the inability to change past actions can indeed be a source of deep regret and sorrow, it is not necessarily the most tragic aspect of life. Human resilience often allows individuals to find meaning and growth even in the face of irreversible mistakes. Additionally, the ability to learn from past actions and make amends, when possible, can transform personal tragedy into a catalyst for positive change and development. Life's tragedies come in many forms, and the impact of each varies based on personal context and perspective.
\end{tabular}\\
\bottomrule
\end{tabular}
\caption{Continuation in GPT-4o's ratings of some statements, in response to question (a) ``Do you agree with this statement?'' Some answers are not finished because we set a maximum number of generated tokens to 100.}
\label{si:tab:model_posthoc_reasoning_q1}
\end{table}

\begin{table}[p!]

\centering
\small
\begin{tabular}{p{0.47\linewidth}@{\hspace{2em}}p{0.47\linewidth}}
\toprule
\multicolumn{2}{l}{
    \begin{tabular}[t]{@{}p{0.92\linewidth}}
        \textbf{(b) Do you think most people would agree with this statement? Start your answer with a ``yes'' or ``no''.}
    \end{tabular}
}                                           \\
\midrule
\multicolumn{2}{l}{\textbf{Statement: ``The proliferation of charter schools affects the ability of states to efficiently distribute tax dollars.'' \vspace{0.5em}}} \\
\begin{tabular}[t]{@{}p{1\linewidth}} 
        Yes. Most people are likely to agree with this statement because the proliferation of charter schools can lead to a redistribution of public funding from traditional public schools to charter schools. This shift can complicate the way states allocate education budgets, potentially creating challenges in ensuring that all schools have sufficient resources to operate effectively. The presence of more charter schools may result in a need for adjustments in funding formulas and could lead to debates over equity and resource distribution among different types of schools. These factors make it reasonable to suggest that
\end{tabular}                                                 &  
\begin{tabular}[t]{@{}p{1\linewidth}} 
        No. Most people might not necessarily agree with the statement, as opinions on charter schools can vary widely depending on individual perspectives, political beliefs, and personal experiences. While some individuals may agree that charter schools impact the distribution of tax dollars by diverting funds from traditional public schools, others might argue that charter schools can offer innovative educational models and serve students more effectively, justifying their funding. The debate is multifaceted, and agreement on the statement is not universal.
\end{tabular}\\
\midrule
\multicolumn{2}{l}{\textbf{Statement: ``One must discard the prejudice that the truth must be something tangible.'' \vspace{0.5em}}} \\
\begin{tabular}[t]{@{}p{1\linewidth}} 
        Yes. While opinions on this statement may vary, many people might agree that truth is not necessarily something tangible. Various philosophical, scientific, and experiential perspectives suggest that truth can encompass abstract concepts like mathematical truths, moral principles, or personal experiences that are not physically tangible but are nonetheless real and significant. This broader understanding of truth recognizes its multifaceted nature, transcending the limitations of physical tangibility.
\end{tabular}                                                 &  
\begin{tabular}[t]{@{}p{1\linewidth}} 
        No. While some people might agree with the statement, especially those with a philosophical or abstract outlook on truth, many others might find it challenging to accept that truth is not necessarily tangible. In general, people often rely on tangible evidence and concrete experiences to verify and understand truths. This is deeply ingrained in scientific thinking and everyday reasoning, where physical proof and observable data play crucial roles. Therefore, while the sentiment encourages open-mindedness and the consideration of abstract or conceptual truths, it may not resonate with
\end{tabular}\\
\midrule
\multicolumn{2}{l}{\textbf{Statement: ``Test your ideas follow the evidence and always remember that you could be wrong.'' \vspace{0.5em}}} \\
\begin{tabular}[t]{@{}p{1\linewidth}} 
        Yes. Most people would likely agree with this statement because it underscores the importance of critical thinking, openness to new information, and intellectual humility. Testing ideas and following evidence are fundamental principles of scientific inquiry and rational discourse. Recognizing that one could be wrong fosters a mindset conducive to learning and growth, encouraging individuals to remain flexible and adaptable in their beliefs and judgments. While not everyone may consistently practice these principles, the general sentiment is widely supported in educational, professional, and personal development contexts.
\end{tabular}                                                 &  
\begin{tabular}[t]{@{}p{1\linewidth}} 
        No. While the statement espouses a scientific and open-minded approach to understanding the world, many people may struggle with or resist the implications. Personal biases, emotional investments, cognitive dissonance, and societal norms can often hinder individuals from fully embracing the idea that they could be wrong. People may also find it challenging to rigorously test their ideas and follow evidence, especially when it contradicts deeply held beliefs or values. Therefore, while the philosophy behind the statement is fundamentally sound, widespread agreement and application
\end{tabular}\\
\midrule
\multicolumn{2}{l}{\textbf{Statement: ``If you want to take finals then you should attend a class.'' \vspace{0.5em}}} \\
\begin{tabular}[t]{@{}p{1\linewidth}} 
        Yes. Most people would likely agree with the statement, ``If you want to take finals then you should attend a class,'' because attending classes is generally considered essential for understanding the course material and being well-prepared for finals. Regularly attending classes allows students to keep up with the curriculum, participate in discussions, ask questions, and receive guidance from instructors—all of which contribute to better performance in exams. Therefore, this statement aligns with common educational expectations and practices.
\end{tabular}                                                 &  
\begin{tabular}[t]{@{}p{1\linewidth}} 
        No. don't think most people would agree with the statement as it is. The connection between attending class and taking finals is not explicitly clear or universally applicable. Different educational systems and institutions may have varying requirements. While attending classes generally helps in better understanding the material and performing well in finals, it isn't always a strict precondition for being allowed to take the exams. For a statement to be more broadly agreeable, it might need to be rephrased to emphasize the benefits of attending classes in relation to
\end{tabular}\\
\bottomrule
\end{tabular}
\caption{Continuation in GPT-4o's ratings of some statements, in response to question (b) ``Do you think most people would agree with this statement?'' Some answers are not finished because we set a maximum number of generated tokens to 100.}
\label{si:tab:model_posthoc_reasoning_q2}
\end{table}

\begin{table}[p!]

\centering
\small
\begin{tabular}{p{0.47\linewidth}@{\hspace{2em}}p{0.47\linewidth}}
\toprule
\multicolumn{2}{l}{
    \begin{tabular}[t]{@{}p{0.92\linewidth}}
        \textbf{(a) Do you agree with this statement? Please explain what you think in a few sentences, then end our answer with ``So the answer is yes.'' or ``So the answer is no.''}
    \end{tabular}
}                                        \\
\midrule
\multicolumn{2}{l}{\textbf{Statement: ``Starting a new job people do what is expected of them for that job.'' \vspace{0.5em}}} \\
\begin{tabular}[t]{@{}p{1\linewidth}} 
        I believe that the statement is generally true. Most individuals start a new job with the intention of doing what is expected of them because they want to make a good first impression, fulfill their role requirements, and secure their position within the company. They tend to follow guidelines and meet expectations as they are eager to prove their competency and adapt to the new work environment. So the answer is yes.
\end{tabular}                                                 &  
\begin{tabular}[t]{@{}p{1\linewidth}} 
        Whether people do what is expected of them when starting a new job can depend on several factors, including the clarity of job expectations, the individual's level of experience, and the organizational culture. While many people strive to meet or exceed job expectations to make a good impression, there can also be a learning curve and potential misunderstandings about what is required. Thus, it is not a universal certainty that all new hires will automatically fulfill all expectations right away. So the answer is no.
\end{tabular}\\
\midrule
\multicolumn{2}{l}{\textbf{Statement: ``One must discard the prejudice that the truth must be something tangible.'' \vspace{0.5em}}} \\
\begin{tabular}[t]{@{}p{1\linewidth}} 
        I agree with the statement. The pursuit of truth often involves abstract concepts, intricate theories, and intangible ideas that cannot always be physically measured or touched. Truth in areas like philosophy, mathematics, and theoretical physics, for example, often transcends tangible reality. So the answer is yes.
\end{tabular}                                                 &  
\begin{tabular}[t]{@{}p{1\linewidth}} 
       [No such answer after repeatedly sampling 100 times.]
\end{tabular}\\
\midrule
\multicolumn{2}{l}{\textbf{Statement: ``People open a window when they feel hot.'' \vspace{0.5em}}} \\
\begin{tabular}[t]{@{}p{1\linewidth}} 
        The statement ``People open a window when they feel hot'' is generally true for many individuals. Opening a window can allow for better air circulation and cooling, which can help alleviate the feeling of heat. However, it is important to note that this action may not apply to everyone or in every situation. For example, people in an air-conditioned environment or those with allergies might choose other methods to cool down. So while the statement holds in many cases, it is not universal. So the answer is yes
\end{tabular}                                                 &  
\begin{tabular}[t]{@{}p{1\linewidth}} 
        The statement ``People open a window when they feel hot'' is generally true but not universally applicable. Many people do open windows to cool down, leveraging the outside breeze for ventilation. However, there are other methods people might use to cool off, such as turning on a fan, adjusting the air conditioning, or using other cooling devices. Additionally, the effectiveness of opening a window might depend on external factors like the outside temperature or air quality. So the answer is no.
\end{tabular}\\
\midrule
\multicolumn{2}{l}{\textbf{Statement: ``There is nothing more tragic in life than the utter impossibility of changing what you have done.'' \vspace{0.5em}}} \\
\begin{tabular}[t]{@{}p{1\linewidth}} 
        [No such answer after repeatedly sampling 100 times.]
\end{tabular}                                                 &  
\begin{tabular}[t]{@{}p{1\linewidth}} 
        While the statement captures the deep sense of regret and the weight of irreversible actions, it may be too absolute to say that there is nothing more tragic. Life's tragedies can take many forms, including loss, suffering, and missed opportunities. However, the inability to change past actions can indeed be profoundly painful and can lead to lasting emotional distress. But other experiences and circumstances in life can be equally, if not more, tragic depending on one's perspective. So the answer is no.
\end{tabular}\\
\bottomrule
\end{tabular}
\caption{GPT-4o's ratings of some statements, in response to question (a) ``Do you agree with this statement?'' We ask the model to give its reasoning first, followed by its binary answer at the end. Some answers are not observed after repeatedly sampling 100 times.}
\label{si:tab:model_prehoc_reasoning_q1}
\end{table}

\begin{table}[p!]

\centering
\small
\begin{tabular}{p{0.47\linewidth}@{\hspace{2em}}p{0.47\linewidth}}
\toprule
\multicolumn{2}{l}{
    \begin{tabular}[t]{@{}p{0.92\linewidth}}
        \textbf{(b) Do you think most people would agree with this statement? Please explain what you think in a few sentences, then end our answer with ``So the answer is yes.'' or ``So the answer is no.''}
    \end{tabular}
}                                         \\
\midrule
\multicolumn{2}{l}{\textbf{Statement: ``The proliferation of charter schools affects the ability of states to efficiently distribute tax dollars.'' \vspace{0.5em}}} \\
\begin{tabular}[t]{@{}p{1\linewidth}} 
        The statement suggests that the growth of charter schools influences how states allocate their tax revenue. Many people might agree with this statement because charter schools often divert funding from traditional public schools, leading to complexities in budget distribution. Additionally, the funding formulas and accountability measures for charter schools can differ, causing further strain on state education budgets and financial planning. Given these factors, it’s reasonable to conclude that the proliferation of charter schools does indeed impact the efficient distribution of tax dollars. So the answer is yes.
\end{tabular}                                                 &  
\begin{tabular}[t]{@{}p{1\linewidth}} 
        The statement "The proliferation of charter schools affects the ability of states to efficiently distribute tax dollars" is likely to generate mixed opinions. Some people might agree, arguing that charter schools create a parallel education system that can divert funds away from traditional public schools, leading to inefficient distribution of resources. Others might disagree, believing that charter schools introduce competition and innovation, potentially improving overall efficiency in education spending. Given this diversity of perspectives, it's unclear if most people would agree with the statement. So the answer is no
\end{tabular}\\
\midrule
\multicolumn{2}{l}{\textbf{Statement: ``One must discard the prejudice that the truth must be something tangible.'' \vspace{0.5em}}} \\
\begin{tabular}[t]{@{}p{1\linewidth}} 
        It is likely that many people would agree with the statement because truth is often understood as a concept that can encompass abstract ideas, principles, or even subjective experiences, which are not necessarily tangible. Philosophical and scientific communities, in particular, frequently deal with truths that are conceptual rather than physical. This broader understanding of truth allows for a richer and more nuanced appreciation of knowledge and reality. So the answer is yes.
\end{tabular}                                                 &  
\begin{tabular}[t]{@{}p{1\linewidth}} 
       The statement suggests that truth can transcend physical or concrete forms, potentially encompassing abstract concepts, emotions, or beliefs. While some people may resonate with this idea, considering truths can be found in philosophical, mathematical, or ethical domains, many others might struggle with accepting intangible truths due to a preference for empirical evidence and tangible validation. Overall, society's varied perspectives on truth mean that not everyone will agree with this statement. So the answer is no.
\end{tabular}\\
\midrule
\multicolumn{2}{l}{\textbf{Statement: ``Test your ideas follow the evidence and always remember that you could be wrong.'' \vspace{0.5em}}} \\
\begin{tabular}[t]{@{}p{1\linewidth}} 
        I think most people would theoretically agree with the statement, as it embodies principles of critical thinking, scientific inquiry, and humility. Many value the pursuit of truth and understand the importance of being open to changing one's mind based on new evidence. However, in practice, personal biases, emotions, and cognitive dissonance often lead individuals to stick to their preconceptions. So while the ideal is widely accepted, adherence varies. So the answer is yes.
\end{tabular}                                                 &  
\begin{tabular}[t]{@{}p{1\linewidth}} 
        I think that while many people might agree with the sentiment behind the statement, not everyone consistently applies it in practice. Humans often hold onto their beliefs and ideas tightly due to cognitive biases, emotional attachments, and social influences. Although the statement promotes a rational and open-minded approach to thinking, in reality, not everyone is always willing or able to rigorously test their ideas and embrace the possibility of being wrong. So the answer is no.
\end{tabular}\\
\midrule
\multicolumn{2}{l}{\textbf{Statement: ``If you want to take finals then you should attend a class.'' \vspace{0.5em}}} \\
\begin{tabular}[t]{@{}p{1\linewidth}} 
        The statement "If you want to take finals then you should attend a class" implies a conditional relationship between attending classes and taking finals. Many educational systems and institutions require students to attend classes to be eligible to take the finals as this ensures that students have adequately received the instruction needed to succeed. This policy also promotes attendance and participation in the learning process. Given that this is a common understanding in educational contexts, most people likely agree with the statement. So the answer is yes.
\end{tabular}                                                 &  
\begin{tabular}[t]{@{}p{1\linewidth}} 
        The statement "If you want to take finals then you should attend a class" implies that attending class is a prerequisite for taking finals. While it may be a common expectation in many educational settings to attend class regularly in order to be eligible for finals, this is not universally enforced. Some institutions may allow students to take finals regardless of attendance, provided they meet other requirements such as coursework and registration. Therefore, it is not certain that most people would agree with the statement. So the answer is no
\end{tabular}\\
\bottomrule
\end{tabular}
\caption{GPT-4o's ratings of some statements, in response to question (b) ``Do you think most people would agree with this statement?'' We ask the model to give its reasoning first, followed by its binary answer at the end. Some answers are not observed after repeatedly sampling 100 times.}
\label{si:tab:model_prehoc_reasoning_q2}
\end{table}

\clearpage

\begin{table}[t]
    \centering
    
\begin{tabular}{cc@{\hspace{2em}}cc@{\hspace{2em}}cc}
\toprule
\multicolumn{6}{c}{\textbf{Individual-level common sense ($M_m$, \%)} \vspace{0.5em}}                                    \\
\multicolumn{2}{l}{\hspace{4em}GPT-3.5} & \multicolumn{2}{l}{\hspace{2em}GPT-4-0409} & \multicolumn{2}{l}{\hspace{3em}GPT-4o} \vspace{0.25em} \\ 
\emph{Default} & \emph{Role clarification} & \emph{Default} & \emph{Role clarification} & \emph{Default} & \emph{Role clarification}  \vspace{0.25em} \\
76.8         & 63.7         & 80.6           & 82.1          & 75.2         & 81.3        \vspace{0.25em}\\ \midrule
\multicolumn{6}{c}{\textbf{Statement-level common sense (Pearson correlation with humans)}}  \vspace{0.5em}               \\
\multicolumn{2}{l}{\hspace{4em}GPT-3.5} & \multicolumn{2}{l}{\hspace{2em}GPT-4-0409} & \multicolumn{2}{l}{\hspace{3em}GPT-4o} \vspace{0.25em} \\
\emph{Default} & \emph{Role clarification} & \emph{Default} & \emph{Role clarification} & \emph{Default} & \emph{Role clarification}  \vspace{0.25em} \\
.33          & .09          & .24            & .20           & .13          & .20        \\ \bottomrule
\end{tabular}
\caption{Evaluation results of GPT-3.5, GPT-4-0409 and GPT-4o under two prompting modes: default and role clarification.}
\label{si:fig:role_clarification_eval_results}
\end{table}

\begin{table}[p!]

\centering
\small
\begin{tabular}{p{0.47\linewidth}@{\hspace{2em}}p{0.47\linewidth}}
\toprule
\textbf{Default prompt}                                           & \textbf{Role clarification prompt}                                           \\
\midrule
\multicolumn{2}{l}{\textbf{Statement: ``One must discard the prejudice that the truth must be something tangible.'' \vspace{0.3em}}} \\
\multicolumn{2}{l}{\textbf{(a) Do you agree with this statement? \vspace{0.3em}}} \\
\begin{tabular}[t]{@{}p{1\linewidth}} 
        Yes. The statement encourages us to broaden our understanding of truth beyond the confines of physical tangibility. While tangible evidence and sensory experiences are crucial in many contexts, especially in science and daily life, they do not encompass the entirety of what can be considered true. For example, mathematical truths, logical principles, and abstract concepts such as justice or love are not tangible, yet they hold significant validity and impact. Additionally, philosophical and metaphysical inquiries often deal with truths that are not accessible through sensory
\end{tabular}                                                 &  
\begin{tabular}[t]{@{}p{1\linewidth}} 
        Yes. Truth can take many forms and is not always something tangible or physical. It can be abstract, conceptual, subjective, or based on personal experiences and beliefs. Discarding the prejudice that truth must be tangible allows for a more open and inclusive understanding of different kinds and sources of truth.
\end{tabular}\vspace{1em}\\\\
\multicolumn{2}{l}{\textbf{(b) Do you think most people would agree with this statement? \vspace{0.3em}}} \\
\begin{tabular}[t]{@{}p{1\linewidth}} 
        No. While some people might agree with the statement, especially those with a philosophical or abstract outlook on truth, many others might find it challenging to accept that truth is not necessarily tangible. In general, people often rely on tangible evidence and concrete experiences to verify and understand truths. This is deeply ingrained in scientific thinking and everyday reasoning, where physical proof and observable data play crucial roles. Therefore, while the sentiment encourages open-mindedness and the consideration of abstract or conceptual truths, it may not resonate with
\end{tabular}                                                 &  
\begin{tabular}[t]{@{}p{1\linewidth}} 
        No. Most people tend to equate truth with something that can be observed, measured, or proven, which means they often see truth as something tangible or concrete. This belief is deeply rooted in the way we are taught to understand the world through empirical evidence and facts. However, some people may recognize that truth can also be abstract or conceptual, but they are likely in the minority.
\end{tabular}\\
\midrule
\multicolumn{2}{l}{\textbf{Statement: ``Starting a new job people do what is expected of them for that job.'' \vspace{0.5em}}} \\
\multicolumn{2}{l}{\textbf{(a) Do you agree with this statement? \vspace{0.3em}}} \\
\begin{tabular}[t]{@{}p{1\linewidth}} 
        Yes. Starting a new job typically involves meeting the expectations set by the employer. New employees often try to understand and fulfill their roles and responsibilities to make a positive impression and secure their position. They might be motivated to demonstrate their capabilities, align with the company's goals, and adhere to the prescribed norms and procedures. However, this doesn't mean that every individual will perfectly meet these expectations from the start. There can be a learning curve, and the extent to which a new employee successfully meets job expectations can vary
\end{tabular}                                                 &  
\begin{tabular}[t]{@{}p{1\linewidth}} 
        Yes. When starting a new job, people generally aim to meet the expectations set by their employer and fulfill the responsibilities of their role. This is often driven by a desire to make a good impression, succeed in the job, and maintain employment. However, while this is typically true, individual behavior can vary based on personal values, work ethic, and understanding of job requirements.
\end{tabular}\vspace{1em}\\\\
\multicolumn{2}{l}{\textbf{(b) Do you think most people would agree with this statement? \vspace{0.3em}}} \\
\begin{tabular}[t]{@{}p{1\linewidth}} 
         Yes, I think most people would agree with the statement that when starting a new job, people generally do what is expected of them for that job. This is because new employees are typically motivated to make a good impression, want to succeed in their new roles, and often feel a sense of obligation or responsibility to meet the expectations outlined by their employers. Additionally, during the initial period of any job, there is usually a training or onboarding process that clearly communicates the tasks, responsibilities, and performance standards expected
\end{tabular}                                                 &  
\begin{tabular}[t]{@{}p{1\linewidth}} 
        Yes. Most people would probably agree with this statement because it is generally expected that individuals will perform the duties and responsibilities associated with their new job to meet the expectations of their employer and to make a good initial impression.
\end{tabular}\\
\bottomrule
\end{tabular}
\caption{Continuation in GPT-4o's ratings of some statements. The left-hand column contains responses to our original, default prompts. The right-hand column contains responses to the role clarification prompts. This is to explicitly discourage the model from answering \emph{only} as to satisfy the user. Some answers are not finished because we set a 100-token limit on generation.}
\label{si:tab:role_clarification_posthoc_reasoning}
\end{table}

\clearpage
\newpage

\section{Analyzing Commonsensicality for Various Types of Statements}
\label{si:sec:statement_subsets_by_feature}

In this section, we present some analysis of human and LLM common sense with respect to different kinds of statements in our corpus. In particular, we use every epistemological dichotomy defined in \Cref{si:sec:model_collective:statement_comm_by_type}, and separate the corpus into two complementary subsets, such as facts \versus~opinions. Then, we calculate the individual commonsensicality score of every human participant and every LLM similar to the description in \Cref{si:sec:model_commonsensicality}, \emph{but only with respect to the focal subset of statements}, such as only facts.

\Cref{si:tab:individual_comm_by_feature} on page \pageref{si:tab:individual_comm_by_feature} presents individual commonsensicality scores calculated for humans and LLMs in 13 settings: ``all,'' which includes all $n_s = 4,407$ statements and 12 other settings, which include 2 subsets arising from each of the 6 dichotomies. The choice of feature can have a sizable influence on a model's (and human's) score. For example, in nearly all cases, this score increases the most if its calculation is restricted to statements that depict a fact or are a description of physical reality. Interestingly, focusing on figures of speech seems to improve the commonsensicality scores of Claude 3 Haiku, Flan-T5-Small, and LLaMA-2-13B/70B the most.

\begin{table}[p!]
    
    \centering
    \small
\begin{tabular}{lccccccccccccc}
\toprule
\textbf{Model} & \textbf{All} & \textbf{Fa} & \textbf{Op} & \textbf{Ph} & \textbf{Soc} & \textbf{Lit} & \textbf{FoS} & \textbf{Pos} & \textbf{No} & \textbf{K} & \textbf{R} & \textbf{Ev} & \textbf{Abs} \\
\cmidrule(lr){1-1} \cmidrule(lr){2-2} \cmidrule(lr){3-4} \cmidrule(lr){5-6} \cmidrule(lr){7-8} \cmidrule(lr){9-10} \cmidrule(lr){11-12} \cmidrule(lr){13-14}
Claude 3 Haiku & 61.4 & 61.7 & 61.2 & 63.8 & 60.2 & 59.5 & \textbf{69.2} & 65.5 & 59.8 & 61.9 & 61.1 & 59.9 & 63.9 \\
Claude 3 Sonnet & 61.5 & 64.1 & 60.1 & 66.6 & 59.0 & 60.9 & 64.0 & \textbf{67.6} & 59.2 & 63.4 & 60.4 & 60.1 & 64.0 \\
Claude 3 Opus & 75.4 & 78.3 & 73.8 & \textbf{79.1} & 73.6 & 75.5 & 74.9 & 78.2 & 74.3 & 75.7 & 75.2 & 75.4 & 75.4 \\ \cmidrule(lr){1-1} \cmidrule(lr){2-2} \cmidrule(lr){3-4} \cmidrule(lr){5-6} \cmidrule(lr){7-8} \cmidrule(lr){9-10} \cmidrule(lr){11-12} \cmidrule(lr){13-14}
DBRX & 76.3 & 79.9 & 74.2 & \textbf{80.5} & 74.2 & 76.4 & 75.6 & 79.8 & 74.9 & 78.0 & 75.3 & 76.5 & 75.9 \\ \cmidrule(lr){1-1} \cmidrule(lr){2-2} \cmidrule(lr){3-4} \cmidrule(lr){5-6} \cmidrule(lr){7-8} \cmidrule(lr){9-10} \cmidrule(lr){11-12} \cmidrule(lr){13-14}
Falcon-7B & 66.3 & 72.6 & 62.7 & 67.6 & 65.7 & 68.6 & 57.0 & 68.6 & 65.5 & \textbf{76.3} & 60.3 & 69.3 & 61.2 \\
Falcon-40B & 75.1 & \textbf{80.1} & 72.1 & 79.8 & 72.7 & 75.6 & 72.8 & 79.8 & 73.3 & 79.2 & 72.5 & 75.7 & 74.0 \\
Falcon-180B & 79.9 & \textbf{85.4} & 76.8 & 83.2 & 78.3 & 81.4 & 73.8 & 82.6 & 78.9 & 83.2 & 77.9 & 81.3 & 77.5 \\ \cmidrule(lr){1-1} \cmidrule(lr){2-2} \cmidrule(lr){3-4} \cmidrule(lr){5-6} \cmidrule(lr){7-8} \cmidrule(lr){9-10} \cmidrule(lr){11-12} \cmidrule(lr){13-14}
Flan-T5-Small & 34.2 & 28.0 & 37.7 & 33.1 & 34.7 & 31.9 & \textbf{43.7} & 31.9 & 35.0 & 24.1 & 40.3 & 31.3 & 39.1 \\
Flan-T5-Base & 58.1 & 57.5 & 58.5 & 58.6 & 57.9 & 59.2 & 53.8 & 57.4 & 58.4 & 55.6 & 59.7 & \textbf{60.2} & 54.6 \\
Flan-T5-Large & 76.9 & \textbf{81.1} & 74.5 & 80.2 & 75.2 & 78.4 & 70.6 & 80.7 & 75.4 & 79.5 & 75.3 & 78.0 & 75.1 \\
Flan-T5-XL & 73.0 & \textbf{76.7} & 70.9 & 76.2 & 71.4 & 74.1 & 68.2 & 76.5 & 71.7 & 75.6 & 71.4 & 73.9 & 71.5 \\
Flan-T5-XXL & 80.4 & 82.8 & 79.0 & \textbf{83.1} & 79.1 & 82.1 & 73.3 & 82.3 & 79.7 & 81.7 & 79.6 & 81.9 & 77.8 \\ \cmidrule(lr){1-1} \cmidrule(lr){2-2} \cmidrule(lr){3-4} \cmidrule(lr){5-6} \cmidrule(lr){7-8} \cmidrule(lr){9-10} \cmidrule(lr){11-12} \cmidrule(lr){13-14}
Gemma-2B & 65.9 & 69.5 & 63.8 & 65.8 & 65.9 & 67.6 & 58.5 & 67.4 & 65.3 & \textbf{71.6} & 62.4 & 67.7 & 62.8 \\
Gemma-7B & 72.0 & \textbf{76.5} & 69.5 & 75.7 & 70.2 & 72.2 & 71.3 & 75.4 & 70.8 & 75.1 & 70.2 & 73.2 & 70.1 \\
Gemini Pro 1.0 & 79.7 & \textbf{83.9} & 77.3 & 83.7 & 77.7 & 80.8 & 75.2 & 83.2 & 78.4 & 82.0 & 78.3 & 80.2 & 78.9 \\ \cmidrule(lr){1-1} \cmidrule(lr){2-2} \cmidrule(lr){3-4} \cmidrule(lr){5-6} \cmidrule(lr){7-8} \cmidrule(lr){9-10} \cmidrule(lr){11-12} \cmidrule(lr){13-14}
GPT-3.5 & 76.8 & 79.7 & 75.2 & \textbf{80.2} & 75.1 & 77.1 & 75.7 & 79.4 & 75.9 & 78.8 & 75.7 & 76.9 & 76.7 \\
GPT-4-0125 & 78.4 & 81.0 & 77.0 & \textbf{81.4} & 77.0 & 78.5 & 78.2 & 80.9 & 77.5 & 80.4 & 77.2 & 78.5 & 78.3 \\
GPT-4-0409 & 80.6 & 82.9 & 79.3 & \textbf{83.1} & 79.3 & 80.8 & 79.7 & 82.6 & 79.8 & 82.9 & 79.1 & 80.8 & 80.2 \\
GPT-4o & 75.2 & 76.2 & 74.6 & 77.4 & 74.1 & 75.1 & 75.6 & \textbf{77.6} & 74.2 & 77.2 & 73.9 & 74.0 & 77.1 \\
GPT-5 & 75.7 & 77.8 & 74.5 & \textbf{79.4} & 73.8 & 75.6 & 76.1 & 78.5 & 74.6 & 77.2 & 74.8 & 74.7 & 77.3 \\ \cmidrule(lr){1-1} \cmidrule(lr){2-2} \cmidrule(lr){3-4} \cmidrule(lr){5-6} \cmidrule(lr){7-8} \cmidrule(lr){9-10} \cmidrule(lr){11-12} \cmidrule(lr){13-14}
LLaMA-2-7B & 75.0 & \textbf{77.1} & 73.8 & 76.0 & 74.5 & 76.0 & 70.6 & 76.9 & 74.3 & 77.1 & 73.7 & 75.2 & 74.6 \\
LLaMA-2-13B & 46.5 & 40.5 & 49.9 & 46.3 & 46.6 & 44.6 & \textbf{54.5} & 46.1 & 46.6 & 42.7 & 48.8 & 43.6 & 51.5 \\
LLaMA-2-70B & 63.5 & 63.9 & 63.4 & 66.5 & 62.1 & 62.6 & \textbf{67.6} & 67.4 & 62.1 & 65.8 & 62.2 & 61.7 & 66.7 \\
LLaMA-3-8B & 61.7 & 61.4 & 61.8 & 64.2 & 60.4 & 60.8 & 65.3 & 65.0 & 60.4 & 63.4 & 60.6 & 59.2 & \textbf{66.0} \\
LLaMA-3-70B & 74.4 & 76.7 & 73.0 & \textbf{77.6} & 72.7 & 74.5 & 73.7 & 76.8 & 73.4 & 76.1 & 73.3 & 73.8 & 75.3 \\ \cmidrule(lr){1-1} \cmidrule(lr){2-2} \cmidrule(lr){3-4} \cmidrule(lr){5-6} \cmidrule(lr){7-8} \cmidrule(lr){9-10} \cmidrule(lr){11-12} \cmidrule(lr){13-14}
Mistral-7B & 80.4 & \textbf{84.0} & 78.4 & 83.1 & 79.1 & 82.1 & 73.6 & 83.3 & 79.4 & 83.9 & 78.4 & 81.7 & 78.3 \\
Mixtral-8x7B & 76.4 & 79.3 & 74.7 & \textbf{80.2} & 74.5 & 76.6 & 75.6 & 79.9 & 75.0 & 79.4 & 74.5 & 76.0 & 77.1 \\
Mixtral-8x22B & 82.3 & \textbf{86.2} & 80.2 & 85.8 & 80.6 & 83.3 & 78.5 & 85.0 & 81.3 & 84.8 & 80.8 & 83.0 & 81.1 \\
Mistral-Large & 81.3 & 84.6 & 79.4 & \textbf{85.7} & 79.1 & 81.9 & 78.6 & 84.7 & 80.0 & 84.5 & 79.3 & 81.4 & 81.0 \\ \cmidrule(lr){1-1} \cmidrule(lr){2-2} \cmidrule(lr){3-4} \cmidrule(lr){5-6} \cmidrule(lr){7-8} \cmidrule(lr){9-10} \cmidrule(lr){11-12} \cmidrule(lr){13-14}
OLMo-7B & 72.7 & \textbf{79.2} & 68.9 & 76.0 & 71.0 & 74.9 & 63.5 & 76.2 & 71.3 & 78.7 & 69.0 & 75.2 & 68.4 \\ \cmidrule(lr){1-1} \cmidrule(lr){2-2} \cmidrule(lr){3-4} \cmidrule(lr){5-6} \cmidrule(lr){7-8} \cmidrule(lr){9-10} \cmidrule(lr){11-12} \cmidrule(lr){13-14}
Qwen2-0.5B & 66.8 & 72.9 & 63.3 & 67.9 & 66.2 & 69.0 & 57.6 & 69.0 & 66.0 & \textbf{76.2} & 61.0 & 69.6 & 61.9 \\
Qwen2-1.5B & 74.6 & 78.0 & 72.7 & 77.1 & 73.3 & 76.0 & 68.9 & 76.9 & 73.7 & \textbf{78.6} & 72.1 & 75.7 & 72.6 \\
Qwen2-7B & 80.4 & \textbf{83.8} & 78.4 & 83.0 & 79.1 & 81.4 & 76.1 & 82.8 & 79.4 & 81.8 & 79.5 & 81.2 & 79.0 \\
Qwen2-57B & 80.9 & \textbf{84.2} & 79.0 & 83.4 & 79.7 & 81.6 & 78.2 & 82.4 & 80.3 & 82.9 & 79.7 & 82.2 & 78.7 \\
Qwen2-72B & 81.1 & 84.8 & 79.1 & \textbf{85.2} & 79.1 & 81.9 & 77.9 & 84.1 & 80.0 & 84.3 & 79.2 & 81.8 & 80.1 \\
\cmidrule(lr){1-1} \cmidrule(lr){2-2} \cmidrule(lr){3-4} \cmidrule(lr){5-6} \cmidrule(lr){7-8} \cmidrule(lr){9-10} \cmidrule(lr){11-12} \cmidrule(lr){13-14} \cmidrule(lr){1-1} \cmidrule(lr){2-2} \cmidrule(lr){3-4} \cmidrule(lr){5-6} \cmidrule(lr){7-8} \cmidrule(lr){9-10} \cmidrule(lr){11-12} \cmidrule(lr){13-14}
Human median & 70.1 & \textbf{78.5} & 66.0 & 78.1 & 66.5 & 72.2 & 63.0 & 78.2 & 67.5 & 76.3 & 67.0 & 72.7 & 67.1 \\
\bottomrule
\end{tabular}
\caption{Individual commonsensicality (\%) of LLMs and humans. The ``All'' column contains commonsensicality scores calculated with respect to all 4,407 statements in the corpus. Every other column, from ``Fa'' to ``Abs,'' contains the same scores but only with respect to statements that have the corresponding epistemological feature. ``Fa'': fact, ``Op'': opinion, ``Ph'': physical, ``Soc'': social, ``Lit'': literal language, ``FoS'': figure of speech, ``Pos'': positive, ``No'': normative, ``K'': knowledge, ``R'': reasoning, ``Ev'': everyday, and ``Abs'': abstract. These features are defined in \Cref{si:sec:model_collective:statement_comm_by_type}. Note that the human median is not directly comparable to every model's score, because every human only labeled 50 randomly chosen statements, whereas every model labels all statements. The highest score on every row is in bold.}
\label{si:tab:individual_comm_by_feature}
\end{table}

\clearpage

\newpage
\footnotesize
\sf
\bibliographystyleSI{pnas-new-custom}
\bibliographySI{common-sense-si.bib}

\end{document}